\pdfoutput=1
\documentclass[11pt]{article}

\usepackage[]{acl}

\usepackage{times}
\usepackage{latexsym}
\usepackage{amsmath, amssymb}
\newtheorem{theorem}{Theorem}
\usepackage[T1]{fontenc}
\usepackage{hyperref}
\usepackage{booktabs}
\usepackage{multirow}
\usepackage{mathtools}
\usepackage[utf8]{inputenc}
\usepackage{xspace}
\usepackage{graphicx}
\usepackage{subfigure}
\usepackage{tabularx}
\usepackage{stfloats}
\usepackage{microtype}
\newcommand{\modelname}{\textsc{Della}\xspace}
\newcommand*{\dif}{\mathop{}\!\mathrm{d}}
\title{Fuse It More Deeply! A Variational Transformer with Layer-Wise \\Latent Variable Inference for Text Generation}

\author{
    Jinyi Hu$^{1,2,3}$,\,  Xiaoyuan Yi$^{5}$,\,   Wenhao Li$^{1,2,3}$,\,   Maosong Sun$^{1,2,3,4}$\thanks{\ \ Corresponding author. Email: sms@tsinghua.edu.cn}, \, Xing Xie$^{5}$ \\
    $^1$ Department of Computer Science and Technology, Tsinghua University, Beijing, China \\
    $^2$ Beijing National Research Center for Information Science and Technology \\
    $^3$ Institute for Artificial Intelligence, Tsinghua University, Beijing, China \\
    $^4$ International Innovation Center of Tsinghua University, Shanghai, China \\
    $^5$ Microsoft Research Asia \\
    \texttt{hu-jy21@mails.tsinghua.edu.cn,}\,\texttt{xiaoyuanyi@microsoft.com}
}

\begin{document}
\maketitle
\begin{abstract}
%Variational Auto-Encoder(VAE) has been widely used in text modeling for its practical representation learning framework. However, trivial training methods often cause the auto-regressive decoder to ignore the latent variables and reduce to a language model, known as \textit{posterior collapse} problem. The recently emerged Transformer-based pre-trained decoder will exacerbate this problem due to the power itself, which restricts the combination of VAE and Transformer-based models. To ameliorate this problem, we proposed a variational Transformer framework to explicitly fuse the latent variables with hidden states in the decoder. Deep fusion consists of two aspects. The first is to learn a series of layer-wise latent variables in each encoder layer. The second is to feed the corresponding latent variables to the decoder with low-rank fusion. We give a principal proof of its validity of amelioration on posterior collapse. Empirically, we evaluate our methods on both unconditional generation and conditional generation tasks. Both quality and diversity can be improved compared with solid baselines.
The past several years have witnessed Variational Auto-Encoder's superiority in various text generation tasks. However, due to the sequential nature of the text, auto-regressive decoders tend to ignore latent variables and then reduce to simple language models, known as the \textit{KL vanishing} problem, which would further deteriorate when VAE is combined with Transformer-based structures. To ameliorate this problem, we propose \modelname, a novel variational Transformer framework. \modelname learns a series of layer-wise latent variables with each inferred from those of lower layers and tightly coupled with the hidden states by low-rank tensor product. In this way, \modelname forces these posterior latent variables to be fused deeply with the whole computation path and hence incorporate more information. We theoretically demonstrate that our method can be regarded as entangling latent variables to avoid posterior information decrease through layers, enabling \modelname to get higher non-zero KL values even without any annealing or thresholding tricks. Experiments on four unconditional and three conditional generation tasks show that \modelname could better alleviate KL vanishing and improve both quality and diversity compared to several strong baselines.

\end{abstract}

\section{Introduction}
\label{sec:introduction}
Variational Autoencoder (VAE) \citep{DiederikPKingma2014AutoEncodingVB, rezende2014stochastic} has proven to be successful in generating various kinds of text, such as stylistic text \citep{hu2017toward,john2019disentangled}, dialogue \citep{zhao-etal-2017-learning}, story \citep{MengHsuanYu2020DraftAE} and poetry \citep{Mixpoet:20}.
\begin{figure}[t]
\centering
\includegraphics[scale=0.55]{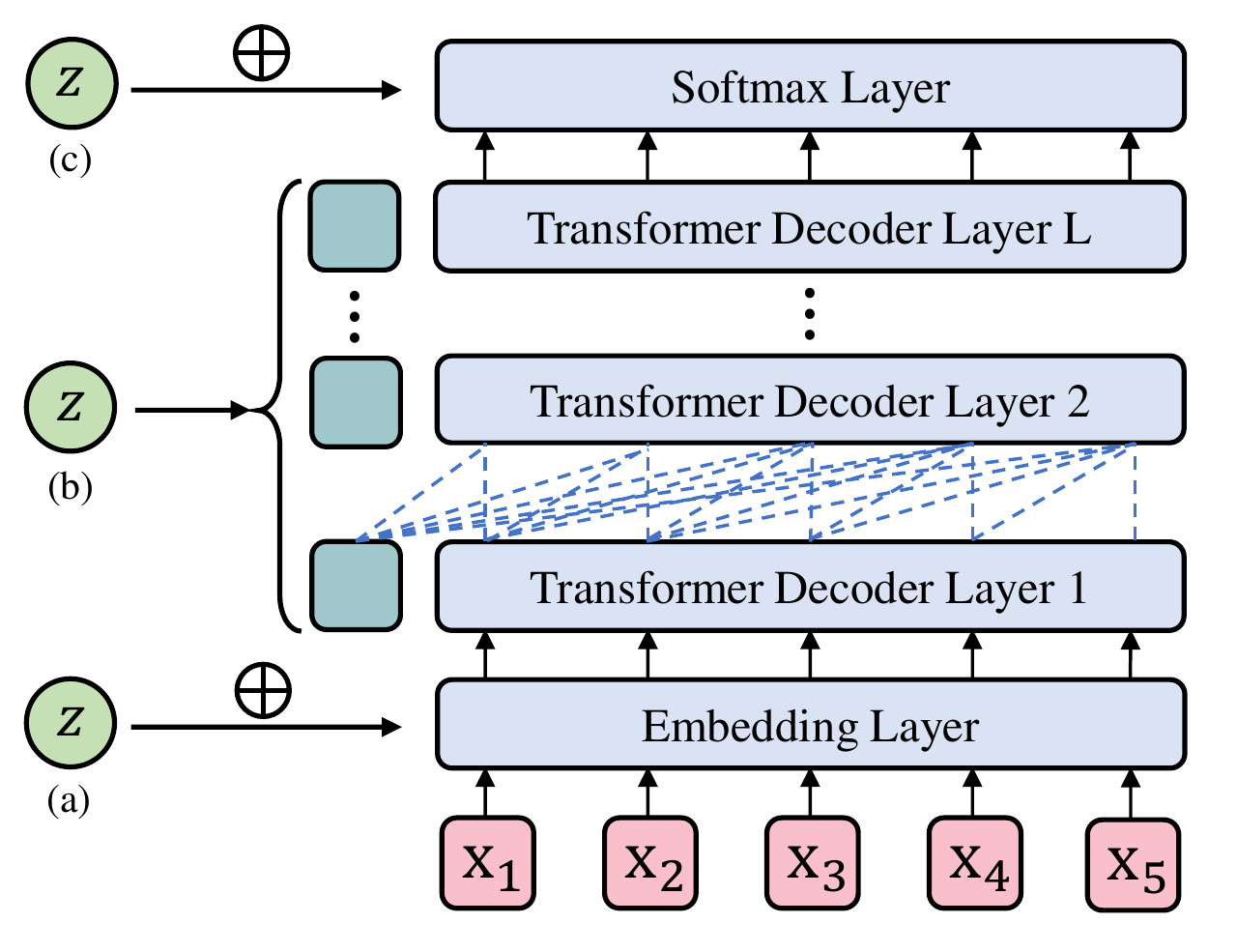}
\caption{Existing paradigms of Transformer VAE.}
\label{fig:baselines}
\end{figure}
The sequential nature of the text leads to typically used auto-regressive decoders in VAE for language generation. However, such strong decoders tend to evade the difficulty of learning meaningful latent codes by heavily relying on previously generated words and hence ignore latent variables~\cite{bowman-etal-2016-generating}, known as \textit{KL vanishing} or \textit{posterior collapse}. This problem causes two drawbacks: (a) the posterior distribution quickly turns into the prior one (usually standard Gaussian), falling to build expressive latent representations; (b) the decoder reduces to a naive language model, resulting in the monotonous generated text \citep{fu-etal-2019-cyclical}.

To ameliorate this problem, researchers have designed various techniques. Among them, three broadly used methods include weakening decoders~\citep{bowman-etal-2016-generating, semeniuta-etal-2017-hybrid, zhao-etal-2017-learning}, KL annealing~\citep{bowman-etal-2016-generating, fu-etal-2019-cyclical} and KL threshold~\citep{kingma2016improved, IrinaHiggins2017betaVAELB, li-etal-2019-surprisingly}. Nonetheless, the weakening of decoders restrains models' language modelling capability; annealing hyperparameters are hard to tune; KL threshold introduces a non-smooth objective with some optimization difficulties.

In the era of RNN, VAE can be easily incorporated by using the latent variable as the initial decoder state, while how to combine VAE with recently prevalent Transformer \citep{vaswani2017attention} architectures, which have made a breakthrough in text generation, still remains an open challenge.

As shown in Fig.\ref{fig:baselines}, existing methods of integrating Transformer into VAE fall into three main paradigms: (a) directly adding latent variables to input token embeddings (abbr. \emph{Embedding})~\citep{li-etal-2020-optimus}; (b) using latent variables as a separate memory token vector to be attended by self-attention in each layer (abbr. \emph{Memory})~\citep{LeFang2021TransformerbasedCV}; (c) combining latent variables with the last-layer decoder states before output softmax (abbr. \emph{Softmax})~\citep{TianmingWang2019TCVAETC}. However, paradigm (a) brings noise for self-attention. In paradigm (b), memory vectors tend to be ignored by attention, even exacerbating KL vanishing. In paradigm (c), latent variables couldn't deeply interfere with the whole computation path. Sec.\ref{sec:baseline_analysis} presents more detailed analyses.

To better incorporate the Transformer into VAE and theoretically ameliorate the \textit{KL vanishing} problem, we propose \modelname\footnote{\ \modelname: \textbf{DE}eply Fused \textbf{L}ayer-wise \textbf{LA}tent Variables}, a novel variational transformer framework. \modelname learns a series of layer-wise latent variables in a Transformer encoder, and each is inferred from those of lower layers and then tightly coupled with the hidden states in the corresponding decoder layer by low-rank tensor product. Our method theoretically stimulates the entanglement of latent variables and hence allows the propagation of undiminished latent information through layers. As a result, \modelname forces posterior latent variables to be deeply fused with the entire computation path and encode richer information of input text, achieving higher KL values even without any annealing or threshold training tricks. 

In summary, our contributions are as follows:
($i$) We are the first to propose layer-wise inferred latent variables in Transformer-based architecture to mitigate KL vanishing; We ($ii$) innovatively inject latent variables using the low-rank tensor product, ($iii$) provide a theoretical validity of our method and ($iv$) demonstrate its effectiveness on four unconditional and three conditional generation tasks. Our codes are available at \href{https://github.com/OpenVLG/DELLA.git}{https://github.com/OpenVLG/DELLA.git}.
\section{Related Work}
Thanks to the representation capacity of latent space, VAE has been widely adopted for both image generation \citep{AaronvandenOord2017NeuralDR, ArashVahdat2020NVAEAD} and text generation \citep{bowman-etal-2016-generating, hu2017toward}. In the early stage, VAE was combined with RNN decoders for generating a broad range of text, varying from dialogue \citep{IulianVladSerban2016AHL}, image caption \citep{LiweiWang2017DiverseAA}, text summarization \citep{AnkushGupta2017ADG} to story \citep{MengHsuanYu2020DraftAE} and poetry \citep{Mixpoet:20}. In this case, latent variables are usually utilized as either the initial decoder state \citep{li-etal-2018-generating-classical} or input at each time step \citep{AnkushGupta2017ADG}. 

In spite of extensive applications, VAE suffered from \textit{KL vanishing} in the scenario of text generation~\citep{bowman-etal-2016-generating}. Several lines of techniques have been proposed to alleviate this problem. The first line is to avoid a too fast decrease of the KL divergence by re-weighting. KL annealing \citep{bowman-etal-2016-generating} linearly increased the weight of KL term from 0 to 1 during the warm-up period. \citet{fu-etal-2019-cyclical} further proposed cyclical annealing, which repeats the warm-up process multiple times. The second line guarantees a positive lower bound of the KL term. KL thresholding~\cite{kingma2016improved} achieved a fixed minimum by combining a hinge loss, while BN-VAE \citep{zhu-etal-2020-batch} learned more flexible ones via batch normalization. $\delta$-VAE \citep{AliRazavi2019PreventingPC} chose to restrain the family of posterior distributions. The third line aims to constrain decoders to force a more informative latent variable. \citet{LiweiWang2017DiverseAA} introduced an auxiliary BOW (bag-of-words) loss. \citet{JunxianHe2019LaggingIN} added additional training loops for the encoder. \citet{yang2017improved} adopted dilated CNN as decoder, and \citet{AdjiBDieng2019AvoidingLV} added skip connections to the decoder. Although the above methods mitigate \textit{KL vanishing} to some extent, it is still challenging for either tuning or optimization.

In these years, the powerful Transformer has been integrated with VAE to benefit diverse tasks, including text classification \citep{gururangan-etal-2019-variational}, story generation \citep{TianmingWang2019TCVAETC, LeFang2021TransformerbasedCV} and dialogue generation \citep{ZhaojiangLin2020VariationalTF}. Optimus \citep{li-etal-2020-optimus} further bridged the pre-trained BERT \citep{devlin-etal-2019-bert} and GPT-2 \citep{Radford2019LanguageMA} with VAE for pre-training. Most existing works inject latent variables into the Transformer decoder by the three paradigms, Embedding~\citep{li-etal-2020-optimus}, Memory~\citep{li-etal-2020-optimus, LeFang2021TransformerbasedCV} and Softmax\citep{TianmingWang2019TCVAETC}, as discussed in Sec. \ref{sec:introduction}, while these methods shallowly fuse the latent variables with hidden states. To achieve deeper fusion and ameliorate KL vanishing, we propose \modelname. 

The most relevant architecture to our model is hierarchical VAE \citep{CasperKaaeSnderby2016LadderVA, AlexejKlushyn2019LearningHP, ArashVahdat2020NVAEAD, child2020very}, which is mainly designed for image generation and not suitable for text. For text generation, hierarchical latent variables are either independent of each other~\citep{IulianVladSerban2016AHL}, or corresponding to different text granularities (sentence or word level), while our \modelname learns conditionally inferred and layer-wise latent variables based on Transformer.

\section{Preliminaries}
\label{sec:preliminaries}
\subsection{Transformer}
Transformer \citep{vaswani2017attention} represents an input sequence $\boldsymbol{x} \!=\!\left\{x_1, \dots, x_i, \dots, x_n\right\}$ as contextualized distributed hidden states $\boldsymbol{h} = \left\{h_1, \dots, h_i, \dots, h_n\right\}$ by a series of stacked layers, and states in the $l$-th layer, $\boldsymbol{h}^{(l)}$, are calculated with scaled dot-product attention: 
\begin{equation}
    \operatorname{Attention}(Q, K, V) =\operatorname{softmax}\left(\frac{Q^{T}K}{\sqrt{d}}\right)V^{T},
\label{eq1}
\end{equation}
% Q is d*n, W is n*n, h is d*n
where $Q, K, V$ stand for Query, Key, Value, respectively, which are projected from outputs of the previous layer:
$Q = W^{q}\boldsymbol{h}^{(l-1)}$, $K= W^{k}\boldsymbol{h}^{(l-1)}$, $V =W^{v} \boldsymbol{h}^{(l-1)}$. $d$ is the dimension of hidden states. In practice, multiple groups of states are calculated with different attention parameters and then concatenated, known as multi-head attention.

\subsection{VAE}
As a kind of generative model, VAE estimates the intractable data distribution $p(\boldsymbol{x})$ by deriving and maximizing its lower bound as:
\begin{align}
    \begin{split}
    &\log p(\boldsymbol{x}) \ge \mathcal{L}_{ELBO}(\boldsymbol{x};\boldsymbol{\theta}, \boldsymbol{\phi})=\\
    &\mathbb{E}_{q_{\boldsymbol{\phi}}(\boldsymbol{z}|\boldsymbol{x})}[\log p_{\boldsymbol{\theta}}(\boldsymbol{x}|\boldsymbol{z})] \!-\! \mathrm{KL}(q_{\boldsymbol{\phi}}(\boldsymbol{z}|\boldsymbol{x}) || p(\boldsymbol{z})),
    \end{split}
\label{eq2}
\end{align}
where $\boldsymbol{z}$ is the latent variable and $p(\boldsymbol{z})$ is the prior distribution of latent variable which is commonly assumed as standard Gaussian; the posterior distribution $p(\boldsymbol{z}|\boldsymbol{x})$ is approximated by an inference network (encoder) $q_{\boldsymbol{\phi}}(\boldsymbol{z}|\boldsymbol{x})$; $p_{\boldsymbol{\theta}}(\boldsymbol{x}|\boldsymbol{z})$ is a generator (decoder) to generate text $\boldsymbol{x}$ from the latent variable $\boldsymbol{z}$; $\boldsymbol{\theta}$ and $\boldsymbol{\phi}$ are corresponding parameters.

The whole lower bound in Eq.(\ref{eq2}), called Evidence Lower BOund (ELBO), consists of two terms: the reconstruction loss,
\begin{align}
\mathcal{L}_{E} &=-\mathbb{E}_{q_{\boldsymbol{\phi}}(\boldsymbol{z}|\boldsymbol{x})}\left[\log p_{\boldsymbol{\theta}}(\boldsymbol{x}|\boldsymbol{z})\right],
\end{align}
which helps reconstruct the input given the posterior latent variable $\boldsymbol{z}$, and the KL divergence,
\begin{align}
\mathcal{L}_{R} =\mathrm{KL}\left(q_{\boldsymbol{\phi}}(\boldsymbol{z}|\boldsymbol{x}) \| p(\boldsymbol{z})\right).
\label{eq4}
\end{align}

In practice, VAE is considered as a regularized Auto-encoder, and a hyper-parameter $\beta$ is introduced to control the strength of KL, $\beta \mathcal{L}_{R}$, usually used in KL annealing methods~\citep{fu-etal-2019-cyclical}.

\subsection{Incorporate Transformer into VAE}
\label{sec:baseline_analysis}
% 是否要加隐向量encoder侧的产生方法
%Incorporating VAE into Transformer is involved two problems: obtaining the latent code from the Transformer encoder and injecting the latent code into the Transformer decoder. Traditional RNN-based encoders commonly take the hidden states of last token. For Transformer encoder, one choice is to take the last-layer hidden states of special token as sentence-level representation \citep{li-etal-2020-optimus}. The other is to make a pooling to all last-layer hidden states \citep{LeFang2021TransformerbasedCV}.

For Transformer encoder, the posterior $\boldsymbol{z}$ is mapped from the text representation, which can be the pooling of all hidden states in the last layer~\citep{LeFang2021TransformerbasedCV}, or state of a special token~\citep{li-etal-2020-optimus}, \textit{e.g.}, [CLS]. Then $\boldsymbol{z}$ is injected into Transformer decoder by the paradigms discussed in Sec. \ref{sec:introduction}. 

Now we take a further step and investigate why intrinsically these three paradigms, namely Embedding, Memory and Softmax, would perform poorly.

\textbf{Embedding}: Define $\boldsymbol{e}_i, \boldsymbol{e}_j$ as two token embeddings and $\alpha_{i,j}$ as the attention weight of $i$-th and $j$-th tokens. From Eq.(\ref{eq1}), we have $\alpha_{i,j} = (W^q\boldsymbol{e}_i)^T(W^k\boldsymbol{e}_j)=\boldsymbol{e}_i^T(W^q)^TW^k\boldsymbol{e}_j$, which is further abbreviated as $\langle \boldsymbol{e}_i, \boldsymbol{e}_j \rangle$. Such Embedding paradigm directly adds $\boldsymbol{z}$ to token embeddings as:
\begin{align}
    \begin{split}
        \alpha'_{i,j} &= \big[W^q(\boldsymbol{e}_i + \boldsymbol{z})\big]^T\big[W^k(\boldsymbol{e}_j + \boldsymbol{z}) \big] \\
                      &= \langle \boldsymbol{e}_i, \boldsymbol{e}_j \rangle + \langle \boldsymbol{e}_i, \boldsymbol{z}\rangle + \langle \boldsymbol{z}, \boldsymbol{e}_j\rangle + \langle \boldsymbol{z}, \boldsymbol{z}\rangle,
    \end{split}
\end{align}
where we can find that a redundant term, $\langle \boldsymbol{z}, \boldsymbol{z}\rangle$, is introduced, bringing extra %computational cost and 
noise for attention mechanism. Moreover, information in $\boldsymbol{z}$ could diminish with propagation through layers (Fig.\,\ref{fig:analysis}), aggravating KL vanishing.

\textbf{Memory}: This paradigm treats $\boldsymbol{z}$ as an additional memory token and places it at the beginning of $\boldsymbol{x}$ to be attended by other tokens via attention. Nevertheless, as mentioned in Sec. \ref{sec:introduction}, the powerful Transformer decoder may only rely on preceding decoded tokens. Consequently, with no explicit constraints (\textit{e.g.}, auxiliary loss), such a memory token is more likely to be ignored by self-attention (Fig. \ref{fig:attn_weight1} \& \ref{fig:attn_weight2}), even exacerbating KL vanishing.

\textbf{Softmax}: This paradigm first adds $\boldsymbol{z}$ to the last-layer hidden states $\boldsymbol{h}$, and then projects $\boldsymbol{z}+\boldsymbol{h}$ into a logit vector $\boldsymbol{p} \in \mathbb{R}^v$ over the vocabulary, where $v$ is vocab size. In this method, latent variables do not interact with hidden states until the last layer, which erodes the effect of latent variables (see Fig.\,\ref{fig:analysis}).

\section{Methodology}
\label{sec:model}
As demonstrated in Sec.\,\ref{sec:preliminaries}, existing three paradigms make latent variables gradually diminish through layers, be ignored by self-attention or inadequately interact with hidden states, which would not mitigate but even worsen the KL vanishing problem.

To deeply fuse latent variables with the whole computation path of Transformer, we propose \modelname to learn a series of layer-wise posterior latent variables which are conditionally inferred in encoder, and injected into hidden states in decoder by low-rank tensor product. We present layer-wise latent variables in Sec. \ref{sec:layer_wise}, describe the tensor product fusion in Sec. \ref{sec:low-rank}, give the theoretical verification of \modelname's effectiveness for ameliorating KL vanishing in Sec. \ref{sec:theorem}, and then extend \modelname to Conditional VAE (CVAE) in Sec. \ref{sec:cvae}.

% \begin{figure}[]
% \centering
% \includegraphics[scale=0.5]{figs/fig2_v1.pdf}
% \caption{Structure of Layer-wise Latent Variable}
% \label{fig:baselines}
% \end{figure}

\subsection{Layer-wise Latent Variables}
\label{sec:layer_wise}
Different from previous work where only one latent variable $\boldsymbol{z}$ is calculated and shared by~\citep{li-etal-2020-optimus} or projected to~\citep{LeFang2021TransformerbasedCV} decoder layers, we involve a series of latent variables $\boldsymbol{z} = \{\boldsymbol{z}_1, \boldsymbol{z}_2, \dots, \boldsymbol{z}_L\}$, where $L$  is the number of Transformer layers. Then we reformulate the prior and posterior distributions as $p(\boldsymbol{z})= \prod_{l=1}^{L} p(\boldsymbol{z}_l|\boldsymbol{z}_{<l})$, $q(\boldsymbol{z}|\boldsymbol{x}) = \prod_{l=1}^{L} q(\boldsymbol{z}_l|\boldsymbol{z}_{<l}, \boldsymbol{x})$, respectively, with each $\boldsymbol{z}_l$ still following Gaussian distribution. Then we rewrite $\mathcal{L}_{R}$ in Eq.(\ref{eq4}) similar to~\citet{ArashVahdat2020NVAEAD}:
\begin{equation}
    \begin{aligned}
        &\mathcal{L}_R=\mathrm{KL}(q(\boldsymbol{z}|\boldsymbol{x}) || p(\boldsymbol{z}))\\
    =&\sum_{l=1}^{L}\mathbb{E}_{q(\boldsymbol{z}_{<l}|\boldsymbol{x})}\left[\mathrm{KL}(q(\boldsymbol{z}_{l}|\boldsymbol{x}, \boldsymbol{z}_{<l})||p(\boldsymbol{z}_{l}|\boldsymbol{z}_{<l}))\right]. 
    \end{aligned}
    \label{eq:lw_kl}
\end{equation}

When $l=1$, $p(\boldsymbol{z}_1|\boldsymbol{z}_{<1}) = p(\boldsymbol{z}_1)$ is the standard Gaussian distribution, $q(\boldsymbol{z}_{1}|\boldsymbol{x}, \boldsymbol{z}_{<1}) = q(\boldsymbol{z}_1|\boldsymbol{x})$. We give detailed derivations in Appendix \ref{apx_sec: proof}. 

These latent variables $\boldsymbol{z}_{l}$ are calculated (inferred) layer by layer using representations of the corresponding layer. Concretely, in $l$-th layer, we use the hidden state of the first token in text $\boldsymbol{x}$, as its $l$-th-layer representation, denoted as $\boldsymbol{x}^{(l)} \in \mathbb{R}^d$, where $d$ is hidden size. Then we represent latent variables in lower layers as $\boldsymbol{z}_{<l}$ and obtain it by:
\begin{equation}
    \boldsymbol{z}_{<l} = \tanh(\boldsymbol{W}_{hh}^{(l)}\boldsymbol{z}_{<l-1} + \boldsymbol{W}_{ih}^{(l)}\boldsymbol{z}_{l-1}),
\end{equation}
where $W_{hh}, W_{ih} \in \mathbb{R}^{p\times p}$, so $\boldsymbol{z}_{<l} \in \mathbb{R}^p$ and $p$ is the dimension of latent variable. $\boldsymbol{z}_0$ and $\boldsymbol{z}_{<0}$ are set as zero vectors. We calculate the mean and variance vectors of $p(\boldsymbol{z}_l|\boldsymbol{z}_{<l})$ and $q(\boldsymbol{z}_l|\boldsymbol{z}_{<l}, \boldsymbol{x})$ by:
\begin{equation}
    \begin{split}
        \begin{pmatrix}
        \boldsymbol{\mu}_p \\
        \log(\boldsymbol{\sigma}^2_p)
    \end{pmatrix} &= \boldsymbol{W}_p^{(l)} \boldsymbol{z}_{<l},\\
    \begin{pmatrix}
        \boldsymbol{\mu}_q \\
        \log(\boldsymbol{\sigma}^2_q)
    \end{pmatrix} &= \boldsymbol{W}_q^{(l)} 
    \begin{pmatrix}
        \boldsymbol{z}_{<l} \\
        \boldsymbol{x}^{(l)}
    \end{pmatrix},
    \end{split}
\end{equation}
where $\boldsymbol{W}_p \in \mathbb{R}^{p \times 2p}$, $\boldsymbol{W}_p \in \mathbb{R}^{p \times 2p}$.

The latent variable $\boldsymbol{z}_{l}$ is sampled from the posterior distribution $q(\boldsymbol{z}_l|\boldsymbol{z}_{<l}, \boldsymbol{x}) = \mathcal{N}(\boldsymbol{\mu}_q, \boldsymbol{\sigma}_q^2\boldsymbol{I})$ for training, and from the prior one $q(\boldsymbol{z}_l|\boldsymbol{z}_{<l})=\mathcal{N}(\boldsymbol{\mu}_p, \boldsymbol{\sigma}_p^2\boldsymbol{I})$ for testing. Since hidden states in each layer belong to different vector spaces, the parameters to calculate each $\boldsymbol{z}_{<l}$, \textit{e.g.}, $\boldsymbol{W}_p^{(l)}$ and $\boldsymbol{W}_q^{(l)}$, \emph{do not} share throughout different layers.

\subsection{Low-rank Tensor Product}
\label{sec:low-rank}
We inject the latent variable $\boldsymbol{z}_l$, which is obtained based on $l$-th encoder layer, into the corresponding $l$-th decoder layer. Instead of simply using $\boldsymbol{z}_l$ as a memory token as discussed in Sec. \ref{sec:baseline_analysis}, we resort to low-rank tensor product, which has been successfully utilized for fusing multimodal representations~\citep{liu-etal-2018-efficient-low}, to deeply fuse latent variables with hidden states in the decoder.

In detail, we conduct low-rank tensor product on $\boldsymbol{z}_l$ and $\boldsymbol{x}_i$'s $l$-th-layer value vector $\boldsymbol{v}_i^{(l)}$ as:
\begin{equation}
    \widetilde{\boldsymbol{v}}_i^{(l)} = (\sum\limits_{j=1}^{r}\boldsymbol{W}_v^{(l, j)}\boldsymbol{v}_i^{(l)}) \circ (\sum\limits_{j=1}^{r}\boldsymbol{W}_z^{(l, j)}\boldsymbol{z}_l),
\end{equation}
where $r$ is a hyper-parameter, $\circ$ means element-wise multiplication, $\boldsymbol{W}_v \in \mathbb{R}^{d\times d}. \boldsymbol{W}_z \in \mathbb{R}^{p\times d}$ are learnable parameters which are shared across all positions ($i$) but not shared with layers ($l$), considering distinct vector spaces in different layers, as mentioned in Sec. \ref{sec:layer_wise}. Then the fused Value $\widetilde{V}^{(l)} = \{\widetilde{\boldsymbol{v}}_1^{(l)}, \dots, \widetilde{\boldsymbol{v}}_n^{(l)}\}$ is used in Eq.(\ref{eq1})

In this way, layer-wise $\boldsymbol{z}_l$ is conditionally inferred from latent variables in previous encoder layers, together with $l$-th-layer text representation, and then explicitly fused with the corresponding decoder layer, yielding a deeper intervention throughout the whole computation path of Transformer.
%Advantage of Layer-wise Latent Variables
\subsection{Why Could \modelname Work Well?}
\label{sec:theorem}
To theoretically interpret the advantage of layer-wise latent variables which contributes most to \modelname (Table \ref{tab:ablation_study}), we give the following theorem:

\begin{theorem}
For an observation $\boldsymbol{x}$ and a sequence of latent variables $\boldsymbol{z}_1, \boldsymbol{z}_2, \dots \boldsymbol{z}_L$, satisfying $p(\boldsymbol{z})= \prod_{l=1}^{L} p(\boldsymbol{z}_l|\boldsymbol{z}_{<l})$, and $q(\boldsymbol{z}|\boldsymbol{x}) = \prod_{l=1}^{L} q(\boldsymbol{z}_l|\boldsymbol{z}_{<l}, \boldsymbol{x})$, then the  expectation of the KL term, $\mathbb{E}_{p(\boldsymbol{x})}[\mathcal{L}_R]$ is an upper bound of:
\begin{equation}
    -\sum\limits_{i=2}^{L-1}I(\boldsymbol{z}_L; \dots ; \boldsymbol{z}_i|\boldsymbol{z}_{i-1}) - I(\boldsymbol{z}_L; \dots ; \boldsymbol{z}_1|\boldsymbol{x}),
\label{eq10}
\end{equation}
where $I$ is the interaction information\footnote{https://en.wikipedia.org/wiki/Interaction\_information}.
\label{thm1}
\end{theorem}
See Appendix \ref{proof:theorem1} for proof. Based on Theorem \ref{thm1}, minimizing $\mathcal{L}_R$ approximatively means maximizing each interaction information term in Eq.(\ref{eq10}), which forces the entanglement of all latent variables $\boldsymbol{z}_1; \dots ; \boldsymbol{z}_L$ given the observation $\boldsymbol{x}$, alleviating the diminishing of information encoded in latent variables when propagating through layers.

\subsection{Extension to CVAE}
\label{sec:cvae}
\modelname could also be applied to CVAE for conditional generation tasks like storytelling. Given an observation $\boldsymbol{x}$ and its condition $\boldsymbol{c}$, we can optimize:
\begin{align}
    \log p(\boldsymbol{x}|\boldsymbol{c}) & \geq
    \mathbb{E}_{q_{\boldsymbol{\phi}}(\boldsymbol{z}|\boldsymbol{x}, \boldsymbol{c})}[\log p_{\boldsymbol{\theta}}(\boldsymbol{x}|\boldsymbol{z}, \boldsymbol{c})] \\ 
    & - \mathrm{KL}(q_{\boldsymbol{\phi}}(\boldsymbol{z}|\boldsymbol{x}, \boldsymbol{c}) || p(\boldsymbol{z}|\boldsymbol{c})) \nonumber,
\end{align}
and then replace the prior distribution $q(\boldsymbol{z}_{l}|\boldsymbol{x}, \boldsymbol{z}_{<l})$ and posterior distribution $p(\boldsymbol{z}_{l}|\boldsymbol{z}_{<l})$ in Eq.(\ref{eq:lw_kl}) with $q(\boldsymbol{z}_{l}|\boldsymbol{x},\boldsymbol{c}, \boldsymbol{z}_{<l})$ and $p(\boldsymbol{z}_{l}|\boldsymbol{z}_{<l}, \boldsymbol{c})$, respectively.

In this case, we encode the condition $\boldsymbol{c}$ with the same encoder. Similarly, we can obtain the representation of $\boldsymbol{c}$ at $l$-th layer, denoted as $\boldsymbol{c}^{(l)} \in \mathbb{R}^d$, and then calculate the mean and log variance of 
$p(\boldsymbol{z}_l|\boldsymbol{z}_{<l}, \boldsymbol{c})$ and $q(\boldsymbol{z}_l|\boldsymbol{z}_{<l}, \boldsymbol{x}, \boldsymbol{c})$ by:
\begin{equation}
    \begin{split}
        \begin{pmatrix}
        \boldsymbol{\mu}_p \\
        \log(\boldsymbol{\sigma}^2_p)
    \end{pmatrix} &= \hat{\boldsymbol{W}}_p^{(l)} \begin{pmatrix}
        \boldsymbol{z}_{<l} \\
        \boldsymbol{c}^{(l)}
    \end{pmatrix},\\
    \begin{pmatrix}
        \boldsymbol{\mu}_q \\
        \log(\boldsymbol{\sigma}^2_q)
    \end{pmatrix} &= \hat{\boldsymbol{W}}_q^{(l)} 
    \begin{pmatrix}
        \boldsymbol{z}_{<l} \\
        \boldsymbol{x}^{(l)} \\
        \boldsymbol{c}^{(l)}
    \end{pmatrix},
    \end{split}
\end{equation}
where $\hat{\boldsymbol{W}}_p^{(l)} \in \mathbb{R}^{(p + d) \times 2p}$, $\hat{\boldsymbol{W}}_q^{(l)} \mathbb{R}^{(p + 2d) \times 2p}$.

\section{Experiment}

\subsection{Dataset}
We consider four datasets for language modelling and unconditional generation, including the Yelp, and Yahoo \citep{yang2017improved, JunxianHe2019LaggingIN}, Penn Treebank (PTB) \citep{marcus-etal-1993-building}, and SNLI \citep{bowman-etal-2015-large}, and three datasets for conditional generation tasks, including summarization generation with CNN/DailyMail (CNN/DM) \citep{see-etal-2017-get}, story generation with WritingPrompts (WP) \citep{fan-etal-2018-hierarchical} and paraphrase generation with Quora \footnote{https://quoradata.quora.com/First-Quora-Dataset-Release-Question-Pairs}. Detailed data statistics are listed in Table \ref{tab:stat_dataset}. Due to the limited computation capability, we use 165,157 samples in CNN/DM and 22,2614 in WP with the max length of 900 for training. 

\subsection{Implementation Details}
We use pretrained language models as the backbone and fine-tune them on each task mentioned above with our \modelname as in~\citep{li-etal-2020-optimus}. For unconditional generation and story generation, encoder and decoder \textbf{shared} the same parameters initialized with 12-layer GPT-2~\citep{Radford2019LanguageMA}. For summarization and paraphrase generation, parameters are \textbf{not shared} and initialized with BART-base \citep{lewis-etal-2020-bart}.  We set the dimension of latent variable as 32 for all VAE-based models and use cyclical annealing for training, following~\citep{li-etal-2020-optimus}. More details are given in Appendix \ref{apx_sec:implemet}.

\subsection{Baseline}
We make a comprehensive comparison with strong Transformer-based baselines. We do not consider RNN-based models that are inferior to Transformer for text generation as shown in~\citep{li-etal-2020-optimus}.

\textbf{Finetuned Pretrained Models}.
To manifest the suitability of \modelname for different pretrained language models, we compare it with fine-tuned GPT2 on unconditional generation and story generation, and with fine-tuned BART-base on summarization generation and paraphrase generation.
%We compare \modelname with fine-tuned GPT2~\citep{Radford2019LanguageMA} on unconditional generation and story generation, and with fine-tuned BART-base~\citep{lewis-etal-2020-bart} on summarization generation and paraphrase generation to manifest the suitability of \modelname for different pretrained language models.

\textbf{Optimus}~\citep{li-etal-2020-optimus}: a large-scale VAE model which takes a pre-trained BERT as encoder and pretrained GPT-2 as decoder. This model is first pretrained as a VAE, which simultaneously utilizes the two paradigms, Embedding and Memory as introduced in Sec. \ref{sec:baseline_analysis}, for injecting latent variables, with both KL annealing and KL threshold tricks, and then fine-tuned on downstream tasks.

\textbf{Transformer-based VAE}. Besides Optimus, we also compre the three paradigms,  namely \textbf{Embedding}~\cite{li-etal-2020-optimus}, \textbf{Memory}~\citep{LeFang2021TransformerbasedCV} and \textbf{Softmax}~\citep{TianmingWang2019TCVAETC}, and incorporate each paradigm into the same pre-trained model as \modelname on each dataset for fair comparison.

\subsection{Metrics}
\emph{For unconditional generation tasks}, we consider three types of metrics. \textbf{(a) Representation Learning Capability}: we report PPL, ELBO, KL, mutual information (MI)~\citep{AlexanderAAlemi2016DeepVI} and activate units (AU)~\citep{YuriBurda2016ImportanceWA}. These metrics measure VAE's ability to mitigate KL vanishing and learn meaningful representations. Different from traditional language models like GPT-2, VAE-based models could not produce exact PPL due to randomness, so we use importance-weighted samples to estimate PPL, following~\citet{JunxianHe2019LaggingIN}. We set the threshold in AU to 0.2 to further distinguish different models. \textbf{(b) Generation Quality}: we report BLEU~\citep{papineni-etal-2002-bleu}, CND~\citep{JianingLi2020OnTR} and MAUVE~\citep{pillutla-etal:mauve:neurips2021}. CND and MAUVE measure the divergence between human-authored text and the generated one. \textbf{(c) Generation Diversity}: we report Self-BLEU \citep{YaomingZhu2018TexygenAB}, Dist \citep{li-etal-2016-diversity} and JS (Jaccard similarity) \citep{KeWang2018SentiGANGS} to assess the diversity and novelty of generated text. 

\emph{For conditional generation tasks}, we report BLEU, Rouge-1, Rouge-2, Rouge-L \citep{lin-hovy-2002-manual}, and BERTScore \citep{TianyiZhang2020BERTScoreET} to evaluate the quality of generated texts, as well as the same diversity metrics used in unconditional generation. We also report KL and AU values to present representation learning capability. More details of metrics are provided in Appendix \ref{apx_sec:metrics}.
%-------------------------------------------------------------
\begin{table*}[htp]
\centering
\scalebox{0.8}{
\begin{tabular}{cccccccccccc}
\toprule
\multicolumn{1}{c|}{\multirow{2}{*}{Model}} & \multicolumn{5}{c|}{Representation   Learning} & \multicolumn{3}{c|}{Generation Quality} & \multicolumn{3}{c}{Generation   Diversity} \\ \cline{2-12} 
\multicolumn{1}{c|}{}  & \multicolumn{1}{c|}{PPL$\downarrow$}   &\multicolumn{1}{c|}{ELBO$\downarrow$}   & \multicolumn{1}{c|}{KL$\uparrow$}    & \multicolumn{1}{c|}{MI$\uparrow$}    & \multicolumn{1}{c|}{AU$\uparrow$} & \multicolumn{1}{c|}{BLEU$\uparrow$}  & \multicolumn{1}{c|}{CND$\downarrow$} &  \multicolumn{1}{c|}{MAUVE$\uparrow$} &  \multicolumn{1}{c|}{SB$\downarrow$}      & \multicolumn{1}{c|}{Dist$\uparrow$} & JS $\downarrow$     \\ \midrule
\multicolumn{12}{c}{Dataset: Yelp}                                                                                                                                                                                                                                                                \\ \midrule
\multicolumn{1}{c|}{GPT-2}                   & \multicolumn{1}{c|}{22.13}  & \multicolumn{1}{c|}{-}       & \multicolumn{1}{c|}{-}      & \multicolumn{1}{c|}{-}      & \multicolumn{1}{c|}{-}   & \multicolumn{1}{c|}{56.92}    & \multicolumn{1}{c|}{0.68}      & \multicolumn{1}{c|}{0.12} & \multicolumn{1}{c|}{65.90}  & \multicolumn{1}{c|}{\textbf{17.96}}      &       0.51       \\ 
\multicolumn{1}{c|}{Optimus}                & \multicolumn{1}{c|}{22.79}   & \multicolumn{1}{c|}{344.10}   & \multicolumn{1}{c|}{15.09}      & \multicolumn{1}{c|}{7.67}      & \multicolumn{1}{c|}{-}       & \multicolumn{1}{c|}{-}      & \multicolumn{1}{c|}{-}      & \multicolumn{1}{c|}{-}   & \multicolumn{1}{c|}{-}  & \multicolumn{1}{c|}{-}      &    -          \\ 
\multicolumn{1}{c|}{Embed}                  & \multicolumn{1}{c|}{19.98} & \multicolumn{1}{c|}{327.28} & \multicolumn{1}{c|}{4.77}  & \multicolumn{1}{c|}{4.14}  & \multicolumn{1}{c|}{6}   & \multicolumn{1}{c|}{56.34}& \multicolumn{1}{c|}{0.31}      & \multicolumn{1}{c|}{0.42}      & \multicolumn{1}{c|}{65.27}   & \multicolumn{1}{c|}{15.59}     &  0.44            \\
\multicolumn{1}{c|}{Memory}                    & \multicolumn{1}{c|}{19.95} & \multicolumn{1}{c|}{326.60}  & \multicolumn{1}{c|}{5.70}   & \multicolumn{1}{c|}{5.30}   & \multicolumn{1}{c|}{11}   & \multicolumn{1}{c|}{\textbf{57.37}}& \multicolumn{1}{c|}{0.27} & \multicolumn{1}{c|}{0.46}      & \multicolumn{1}{c|}{63.90}    & \multicolumn{1}{c|}{16.91}    &    \textbf{0.39}          \\ 
\multicolumn{1}{c|}{Softmax}                & \multicolumn{1}{c|}{20.14} & \multicolumn{1}{c|}{328.13} & \multicolumn{1}{c|}{7.50}   & \multicolumn{1}{c|}{6.29}  & \multicolumn{1}{c|}{13}   & \multicolumn{1}{c|}{56.83}& \multicolumn{1}{c|}{0.30} & \multicolumn{1}{c|}{0.45}  & \multicolumn{1}{c|}{64.26}     & \multicolumn{1}{c|}{16.51}   &    0.40          \\ 
\multicolumn{1}{c|}{\modelname}                  & \multicolumn{1}{c|}{\textbf{12.35}} & \multicolumn{1}{c|}{\textbf{239.83}} & \multicolumn{1}{c|}{\textbf{29.47}} & \multicolumn{1}{c|}{\textbf{10.78}} & \multicolumn{1}{c|}{\textbf{23}}   & \multicolumn{1}{c|}{57.15} & \multicolumn{1}{c|}{\textbf{0.13}}  & \multicolumn{1}{c|}{\textbf{0.55}}  & \multicolumn{1}{c|}{\textbf{60.02}}   & \multicolumn{1}{c|}{17.63}     &    0.43          \\ \midrule
\multicolumn{12}{c}{Dataset: Yahoo}                                                                                                                                                                                                                                                               \\ \midrule
\multicolumn{1}{c|}{GPT-2}                   & \multicolumn{1}{c|}{24.17}   & \multicolumn{1}{c|}{-}       & \multicolumn{1}{c|}{-}      & \multicolumn{1}{c|}{-}      & \multicolumn{1}{c|}{-} & \multicolumn{1}{c|}{44.25}   & \multicolumn{1}{c|}{0.55}      & \multicolumn{1}{c|}{0.15}         & \multicolumn{1}{c|}{54.06}    & \multicolumn{1}{c|}{21.07}    &    \textbf{0.28}          \\ 
\multicolumn{1}{c|}{Optimus}                & \multicolumn{1}{c|}{23.11}    & \multicolumn{1}{c|}{293.34}  & \multicolumn{1}{c|}{17.45}      & \multicolumn{1}{c|}{8.85}      & \multicolumn{1}{c|}{-}       & \multicolumn{1}{c|}{-}      & \multicolumn{1}{c|}{-}      & \multicolumn{1}{c|}{-}   & \multicolumn{1}{c|}{-}    & \multicolumn{1}{c|}{-}    & -             \\
\multicolumn{1}{c|}{Embed}                  & \multicolumn{1}{c|}{22.18} & \multicolumn{1}{c|}{286.85} & \multicolumn{1}{c|}{3.63}  & \multicolumn{1}{c|}{3.03}  & \multicolumn{1}{c|}{3}   & \multicolumn{1}{c|}{42.27} & \multicolumn{1}{c|}{0.45}      & \multicolumn{1}{c|}{0.31}      & \multicolumn{1}{c|}{54.15}    & \multicolumn{1}{c|}{20.80}    &      0.32        \\ 
\multicolumn{1}{c|}{Memory}                    & \multicolumn{1}{c|}{22.03} & \multicolumn{1}{c|}{285.47} & \multicolumn{1}{c|}{4.87}  & \multicolumn{1}{c|}{4.62}  & \multicolumn{1}{c|}{18}   & \multicolumn{1}{c|}{\textbf{45.20}} & \multicolumn{1}{c|}{0.46}      & \multicolumn{1}{c|}{0.37}      & \multicolumn{1}{c|}{54.59}    & \multicolumn{1}{c|}{21.87}    &     0.33            \\ 
\multicolumn{1}{c|}{Softmax}                & \multicolumn{1}{c|}{22.35} & \multicolumn{1}{c|}{287.44} & \multicolumn{1}{c|}{6.35}  & \multicolumn{1}{c|}{5.52}  & \multicolumn{1}{c|}{19}   & \multicolumn{1}{c|}{44.28} & \multicolumn{1}{c|}{0.44}      & \multicolumn{1}{c|}{0.34}      & \multicolumn{1}{c|}{54.49}    & \multicolumn{1}{c|}{21.65}    &    0.32          \\ 
\multicolumn{1}{c|}{\modelname}                  & \multicolumn{1}{c|}{\textbf{11.49}} & \multicolumn{1}{c|}{\textbf{201.34}} & \multicolumn{1}{c|}{\textbf{27.84}} & \multicolumn{1}{c|}{\textbf{12.31}}      & \multicolumn{1}{c|}{\textbf{21}}& \multicolumn{1}{c|}{44.67} & \multicolumn{1}{c|}{\textbf{0.19}}      & \multicolumn{1}{c|}{\textbf{0.38}}         & \multicolumn{1}{c|}{\textbf{48.53}} & \multicolumn{1}{c|}{\textbf{21.88}}    &    0.31      \\ \bottomrule
\end{tabular}
}
\caption{Evaluation results for language modelling and unconditional generation. Results of Optimus are directly copied from the original paper with $\lambda=0.5$. SB means Self-BLEU.}
\label{tab:uncond_task}
\end{table*}
%--------------------------------------------------------
\subsection{Results}
\subsubsection{Unconditional Generation}
\begin{figure}[]
\centering
\includegraphics[scale=0.4]{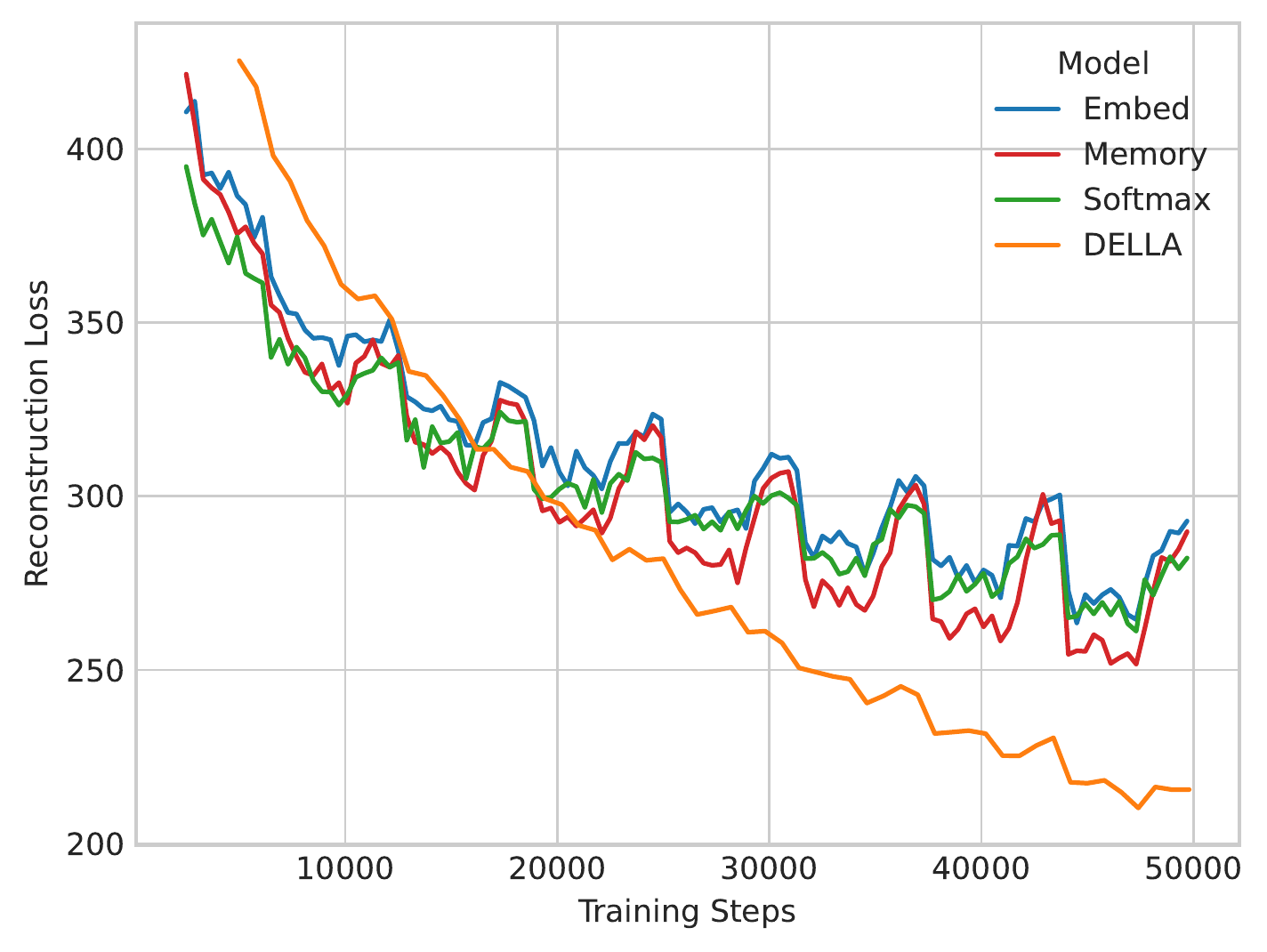} \
\includegraphics[scale=0.4]{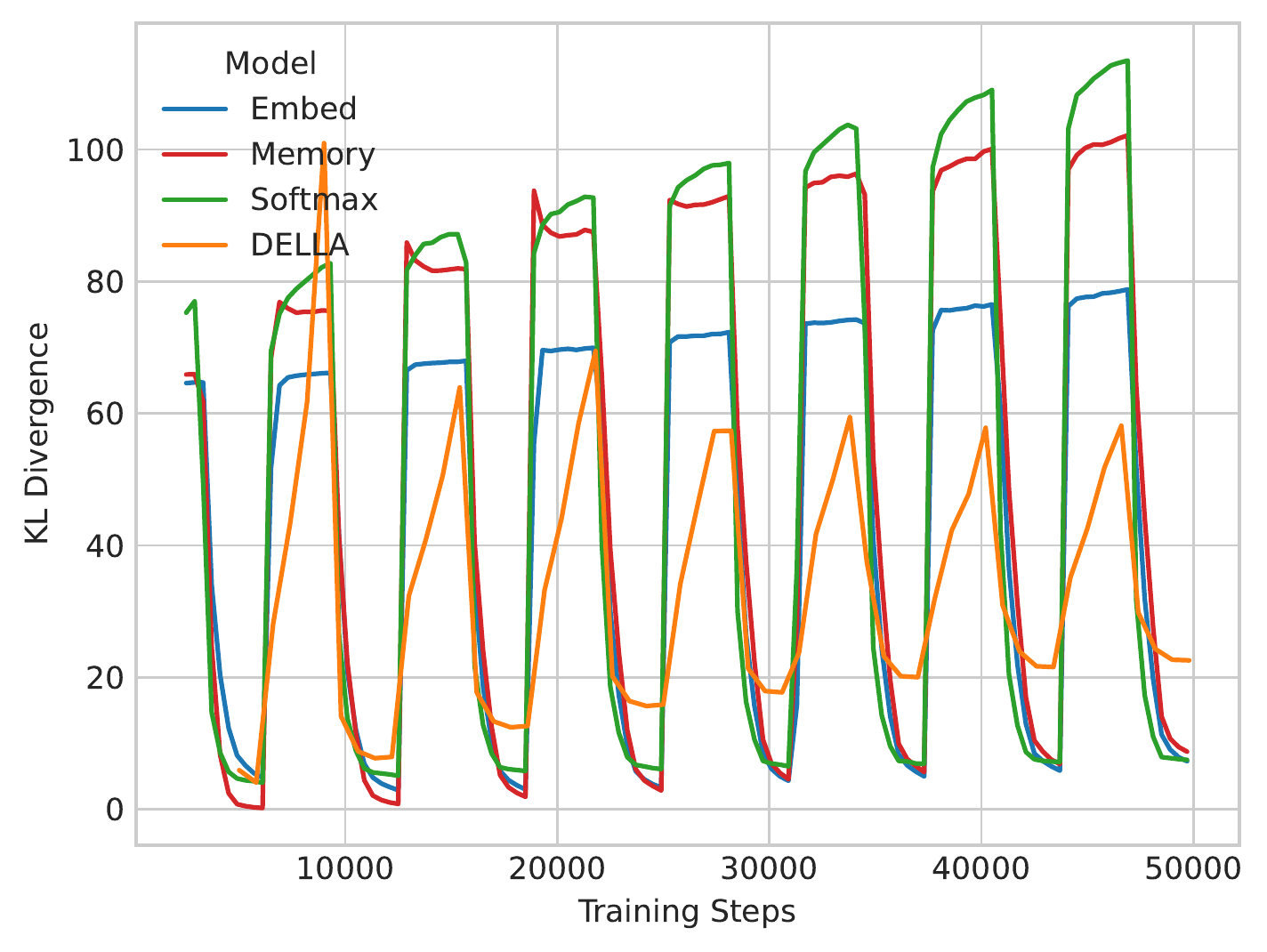} \
\caption{Reconstruction loss and KL Divergence throughout the training process.}
\label{fig:training}
\end{figure}

We present results on Yelp and Yahoo in Table \ref{tab:uncond_task} and leave the those on PTB and SNLI in the Appendix \ref{apx_sec:results} due to space limitations. We also show the learning curves of ELBO and KL in Fig.~\ref{fig:training}.

As shown in Table \ref{tab:uncond_task}, \modelname achieves notable improvement on almost all the metrics, especially superior on representation learning metrics. Much higher KL, MI, AU, and a big gap in PPL obtained by \modelname indicate the latent variables encode more meaningful text information and won't diminish when propagating through Transformer layers, which strongly supports our motivation that fusing latent variables with hidden states more deeply could effectively alleviate the KL vanishing problem. Such results also empirically verify the theoretical advantage of our model (Theorem \ref{thm1}), demonstrating entangled layer-wise latent variables can preserve more encoded knowledge for decoder. We will show that $\boldsymbol{z}$ can involve more information when injected into more layers in Sec. \ref{sec:analysis}.

Besides, \modelname also gets good performance (comparable BLEU and much better CND and MAUVE) on generation quality. With more informative latent variables, \modelname could achieve a better ELBO and hence further boost the learning of data distribution $p(\boldsymbol{x})$ in Eq.(\ref{eq2}), leading to satisfactory quality of generated texts.

Generally, \modelname also outperforms baseline models on generation diversity. The reason is two-fold: randomly sampled latent variables $\boldsymbol{z}$ should bring diversity, while the VAE-based baselines tend to ignore $\boldsymbol{z}$ as mentioned before, losing some randomness. In contrast, latent variables are deeply fused in \modelname, maintaining enough randomness. Besides, each latent variable is sampled in the corresponding layer, and thus such a sampling process accumulates and enhances randomness, further benefiting diversity while keeping good quality.

% \usepackage{multirow}
\subsubsection{Conditional Generation}
%---------------------------------------------
\begin{table*}[htp]
\centering
\scalebox{0.8}{
\begin{tabular}{ccccccccccc}
\toprule
\multicolumn{1}{c|}{\multirow{2}{*}{Model}} & \multicolumn{5}{c|}{Quality}                                                                                                                            & \multicolumn{3}{c|}{Diversity} & \multicolumn{1}{c|}{\multirow{2}{*}{KL$\uparrow$}} & \multicolumn{1}{c}{\multirow{2}{*}{AU$\uparrow$}}   \\ \cline{2-9} 
\multicolumn{1}{c|}{}                       & \multicolumn{1}{c|}{BLEU$\uparrow$} & \multicolumn{1}{c|}{Rouge-1$\uparrow$} & \multicolumn{1}{c|}{Rouge-2$\uparrow$} & \multicolumn{1}{c|}{Rouge-L$\uparrow$} & \multicolumn{1}{c|}{BERTScore$\uparrow$} & \multicolumn{1}{c|}{SB$\downarrow$} & \multicolumn{1}{c|}{Dist$\uparrow$} & \multicolumn{1}{c|}{JS $\downarrow$} &\multicolumn{1}{c|}{}&\multicolumn{1}{c}{}\\ \midrule
\multicolumn{11}{c}{Dataset:   WritingPrompts}\\ \midrule
\multicolumn{1}{c|}{GPT-2}                   & \multicolumn{1}{c|}{27.89}     & \multicolumn{1}{c|}{27.72}        & \multicolumn{1}{c|}{7.96}        & \multicolumn{1}{c|}{14.30}        & \multicolumn{1}{c|}{78.12}          & \multicolumn{1}{c|}{\textbf{53.78}}   &    \multicolumn{1}{c|}{\textbf{22.99}}   &  \multicolumn{1}{c|}{\textbf{0.51}}  & \multicolumn{1}{c|}{-} & -\\ 
\multicolumn{1}{c|}{Embed}                  & \multicolumn{1}{c|}{39.67}     & \multicolumn{1}{c|}{\textbf{36.17}}      & \multicolumn{1}{c|}{7.96}        & \multicolumn{1}{c|}{15.78}        & \multicolumn{1}{c|}{81.64}          & \multicolumn{1}{c|}{64.55}   &    \multicolumn{1}{c|}{14.31}   &   \multicolumn{1}{c|}{0.73}  & \multicolumn{1}{c|}{2.35} & 3 \\ 
\multicolumn{1}{c|}{Memory}                    & \multicolumn{1}{c|}{40.79}     & \multicolumn{1}{c|}{36.13}        & \multicolumn{1}{c|}{8.04}        & \multicolumn{1}{c|}{16.16}        & \multicolumn{1}{c|}{81.68}          & \multicolumn{1}{c|}{67.56}   &    \multicolumn{1}{c|}{12.90}   &  \multicolumn{1}{c|}{0.80}  &  \multicolumn{1}{c|}{0.07} & 0 \\ 
\multicolumn{1}{c|}{Softmax}                & \multicolumn{1}{c|}{41.04}     & \multicolumn{1}{c|}{36.14}        & \multicolumn{1}{c|}{8.12}        & \multicolumn{1}{c|}{16.30}        & \multicolumn{1}{c|}{81.75}          & \multicolumn{1}{c|}{67.02}   &    \multicolumn{1}{c|}{13.08}   &  \multicolumn{1}{c|}{0.78}  &  \multicolumn{1}{c|}{0.32} & 0 \\ 
\multicolumn{1}{c|}{\modelname}                  & \multicolumn{1}{c|}{\textbf{41.39}}     & \multicolumn{1}{c|}{35.46}        & \multicolumn{1}{c|}{\textbf{8.78}}        & \multicolumn{1}{c|}{\textbf{17.20}}        & \multicolumn{1}{c|}{\textbf{81.77}}          & \multicolumn{1}{c|}{56.28}   &    \multicolumn{1}{c|}{20.91}   &  \multicolumn{1}{c|}{0.60}  &  \multicolumn{1}{c|}{\textbf{28.14}} & \textbf{8} \\ \midrule
\multicolumn{11}{c}{Dataset: CNN/DM} \\ \midrule
\multicolumn{1}{c|}{Bart-base}           & \multicolumn{1}{c|}{48.74}   & \multicolumn{1}{c|}{\textbf{41.33}}     & \multicolumn{1}{c|}{ 19.82}        & \multicolumn{1}{c|}{29.63}        & \multicolumn{1}{c|}{87.75}        & \multicolumn{1}{c|}{29.94}   &        \multicolumn{1}{c|}{43.68}   & \multicolumn{1}{c|}{0.10}  & \multicolumn{1}{c|}{-} & - \\ 
\multicolumn{1}{c|}{Embed}                  & \multicolumn{1}{c|}{44.10}     & \multicolumn{1}{c|}{40.43}        & \multicolumn{1}{c|}{19.41}        & \multicolumn{1}{c|}{29.43}        & \multicolumn{1}{c|}{87.60}          & \multicolumn{1}{c|}{29.60}   &        \multicolumn{1}{c|}{44.04}   & \multicolumn{1}{c|}{0.10}  & \multicolumn{1}{c|}{0.0} & 0\\ 
\multicolumn{1}{c|}{Memory}                    & \multicolumn{1}{c|}{46.02}     & \multicolumn{1}{c|}{41.18}        & \multicolumn{1}{c|}{19.74}        & \multicolumn{1}{c|}{29.64}        & \multicolumn{1}{c|}{87.78}          & \multicolumn{1}{c|}{29.79}   &        \multicolumn{1}{c|}{43.92}   & \multicolumn{1}{c|}{0.11}  & \multicolumn{1}{c|}{0.0} & 0\\
\multicolumn{1}{c|}{Softmax}                & \multicolumn{1}{c|}{44.40}     & \multicolumn{1}{c|}{40.94}        & \multicolumn{1}{c|}{19.63}        & \multicolumn{1}{c|}{29.61}        & \multicolumn{1}{c|}{87.00}          & \multicolumn{1}{c|}{29.64}   &        \multicolumn{1}{c|}{44.11}   & \multicolumn{1}{c|}{0.10}  & \multicolumn{1}{c|}{0.0} & 0\\ 
\multicolumn{1}{c|}{\modelname}                  & \multicolumn{1}{c|}{\textbf{49.18}}     & \multicolumn{1}{c|}{41.27}        & \multicolumn{1}{c|}{\textbf{19.85}}        & \multicolumn{1}{c|}{\textbf{29.84}}        & \multicolumn{1}{c|}{\textbf{88.09}}          & \multicolumn{1}{c|}{\textbf{29.07}}   &        \multicolumn{1}{c|}{\textbf{44.24}}   & \multicolumn{1}{c|}{\textbf{0.09}}  &  \multicolumn{1}{c|}{\textbf{0.91}} & \textbf{1} \\ \bottomrule
\end{tabular}}
\caption{Evaluation results for conditional generation.}
\label{tab:cond_task}
\end{table*}
%---------------------------------------------
%We report the results of WP and CNN/DM in Table \ref{tab:cond_task}, and leave those of Quora in Appendix \ref{apx_sec:results}. As we can see, \modelname performs better on most quality metrics, but gets a little worse on diversity compared to GPT. This is because GPT-2 uses top-k sampling for decoding which is more suitable for the quite long text to be generated in WP, while DELLA and other models use beam search. Even so, on both WP and CNN/DM, \modelname still beats all previous VAE paradigms in diversity, manifesting the effectiveness of our \modelname.
We report the results of WP and CNN/DM in Table \ref{tab:cond_task}, and leave those of Quora in Appendix \ref{apx_sec:results}. As we can see, \modelname performs better on most quality metrics, but gets a little worse on diversity compared to GPT-2. This is because GPT-2 may produce some ill-formed contents which `improves' diversity by cheating the metrics but also lead to much worse quality (lower BLEU and Rouge). Even so, on both WP and CNN/DM, \modelname still beats all previous VAE paradigms in diversity, manifesting the effectiveness of our \modelname.

In addition, all baseline methods suffer from severer KL vanishing problems on conditional generation tasks than on unconditional ones. This is because the given condition text could aggravate the reliance of these models on preceding generated tokens and the condition, and therefore bypass latent variables. By contrast, \modelname could learn more informative $\boldsymbol{z}$ and hence keep a relatively higher KL value even given the condition text.
\begin{table}[]
\centering
\scalebox{0.83}{
\begin{tabular}{cccc}
\toprule
\multicolumn{4}{c}{Dataset: WritingPrompts}   \\ \midrule
\multicolumn{1}{c|}{Model}     & \multicolumn{1}{c|}{Fluency} & \multicolumn{1}{c|}{Coherence} & Novelty \\ \midrule
\multicolumn{1}{c|}{GPT2}      & \multicolumn{1}{c|}{1.83}        & \multicolumn{1}{c|}{2.12}        &  2.50     \\ 
\multicolumn{1}{c|}{Embed}     & \multicolumn{1}{c|}{2.16}        & \multicolumn{1}{c|}{2.33}        &  2.67     \\ 
\multicolumn{1}{c|}{Memory}       & \multicolumn{1}{c|}{2.45}        & \multicolumn{1}{c|}{2.28}     &  2.78   \\ 
\multicolumn{1}{c|}{Softmax}   & \multicolumn{1}{c|}{2.48}        & \multicolumn{1}{c|}{\textbf{2.42}}        &  2.85   \\ 
\multicolumn{1}{c|}{\modelname}     & \multicolumn{1}{c|}{\textbf{2.51}}        & \multicolumn{1}{c|}{2.38}   &  \textbf{2.89}       \\ \midrule
\multicolumn{4}{c}{Dataset: CNN/DM}                                                                     \\ \midrule
\multicolumn{1}{c|}{Model}      & \multicolumn{1}{c|}{Informativeness} & \multicolumn{1}{c|}{Coherence} & Novelty \\ \midrule
\multicolumn{1}{c|}{Bart-base}  & \multicolumn{1}{c|}{\textbf{3.12}}    & \multicolumn{1}{c|}{4.32}      &    3.52     \\
\multicolumn{1}{c|}{Embed}      & \multicolumn{1}{c|}{2.88}       & \multicolumn{1}{c|}{4.08}   &    3.50     \\ 
\multicolumn{1}{c|}{Memory}     & \multicolumn{1}{c|}{2.95}       & \multicolumn{1}{c|}{4.23} &    3.48     \\
\multicolumn{1}{c|}{Softmax}    & \multicolumn{1}{c|}{2.91}        & \multicolumn{1}{c|}{\textbf{4.33}} &    3.50     \\ 
\multicolumn{1}{c|}{\modelname} & \multicolumn{1}{c|}{3.05}  & \multicolumn{1}{c|}{\textbf{4.33}} &    \textbf{3.56}     \\ \bottomrule
\end{tabular}}
\caption{Human evaluation results on conditional generation. The scores range from 1 (worst) to 5 (best). The p-value is 0.002 and Kappa score is 0.64 which indicates acceptable inter-annotator agreement. }
\label{tab:human_eval}
\end{table}

\begin{table}[]
\centering
\scalebox{0.85}{
\begin{tabular}{l|c|c|c|c|c}
\toprule
Model             & PPL$\downarrow$ & ELBO$\downarrow$ & KL$\uparrow$ & MI $\uparrow$& AU$\uparrow$ \\ \midrule
\modelname        & \textbf{12.35}  &  \textbf{239.83}  &     \textbf{29.47}  &  \textbf{10.78}  &  \textbf{23}    \\ 
-LTP              & 12.68  &  249.32  &     28.52  &  9.77   &  21    \\ 
-LW               & 19.88  &  324.45  &     20.12  &  7.23   &  18   \\ \midrule
Separate          & 14.17  &  286.30  &     28.82  &  9.88   &  16   \\ 
 $l=1$ KL         & 12.55  &  266.97  &     0.15   &  0.15   &  0  \\ 
 $l=12$ KL        & 12.48  &  263.38  &     0.73   &  0.61   &  0  \\ \midrule
Embed(384)        & 20.11  &  327.29  &     0.55   &  0.38   &  0  \\
Memory(384)       & 20.09  &  326.24  &     0.46   &  0.25   &  0  \\
Softmax(384)      & 20.15  &  330.24  &     5.04   &  7.15   &  0  \\\bottomrule
\end{tabular}
}
\caption{Ablation study on Yelp dataset. LTP: low-rank tensor product. LW: layer-wise latent variables. Separate: latent variables in each layer are independent. $l=1$ or $4$ KL means we only compute KL loss on $\boldsymbol{z}_1$ or $\boldsymbol{z}_L$, respectively. 384 means the dimension of latent variable used in baselines is $12\times32=384$.}
\label{tab:ablation_study}
\end{table}

% \begin{table}[t]
% \centering
% \scalebox{0.85}{
% \begin{tabular}{l|c|c|c|c|c}
% \toprule
% Model            & PPL$\downarrow$ & ELBO$\downarrow$ & KL$\uparrow$ & MI $\uparrow$& AU$\uparrow$ \\ \midrule
% \multicolumn{6}{c}{Backbone: GPT-2 base (12 layers)}   \\ \midrule
% Embed            & 19.98  &  327.28  &  4.77  &  4.14   &  6     \\ 
% Mem              & 19.95  &  326.60  &  5.70  &  5.30   &  11    \\
% Softmax          & 20.14  &  328.13  &  7.50  &  6.29   &  13   \\
% \modelname       & \textbf{12.35}  &  \textbf{239.83}  & \textbf{29.47}   &  \textbf{10.78}   & \textbf{23}  \\\midrule
% \multicolumn{6}{c}{Backbone: GPT-2 medium (24 layers)}   \\ \midrule
% Embed            & 18.33  &  317.44  &  2.13  &  1.44   &  3   \\ 
% Mem              & 18.30  &  317.24  &  4.47  &  4.26   &  10    \\
% Softmax          & 18.47  &  318.80  &  5.80  &  5.03   &  12   \\
% \modelname       & \textbf{11.01}  &  \textbf{230.96}  &  \textbf{17.09} &  \textbf{23.69}  &  \textbf{27} \\
% \bottomrule
% \end{tabular}
% }
% \caption{Performance with different model size.} 
% \label{tab:medium}
% \end{table}

\begin{table}[t]
\centering
\scalebox{0.75}{
\begin{tabular}{l|c|c|c|c|c}
\toprule
Model            & PPL$\downarrow$ & ELBO$\downarrow$ & KL$\uparrow$ & MI $\uparrow$& AU$\uparrow$ \\ \midrule
Embed            & 22.21  &  339.12  &     0.03  &  0.03   &  0    \\ 
+BOW             & 19.98  &  326.51  &     2.75  &  2.48   &  4   \\
+Annealing       & 19.98  &  327.28  &     4.77  &  4.14   &  6    \\
+Annealing + BOW & 20.59  &  332.44  &    19.51  &  9.12   &  \textbf{28}   \\
+Annealing + BN  & 21.14  &  338.59  &    21.09  &  8.98   &  25     \\ \midrule
Memory           & 22.16  &  338.68  &     0.00  &  0.01   &  0    \\ 
+BOW             & 19.87  &  326.00  &     3.89  &  3.59   &  8      \\
+Annealing       & 19.95  &  326.60  &     5.70  &  5.30   &  11    \\
+Annealing + BOW & 20.41  &  331.09  &    18.76  &  9.14   &  \textbf{28}  \\
+Annealing + BN  & 20.25  &  331.59  &    18.11  &  9.07   &  24    \\ \midrule
Softmax          & 22.43  &  333.93  &     0.47  &  0.3    &  0   \\ 
+BOW             & 20.53  &  331.89  &    10.16  &  5.57   &  \textbf{28}  \\
+Annealing       & 20.14  &  328.13  &     7.50  &  6.29   &  13    \\
+Annealing + BOW & 21.14  &  335.48  &    17.51  &  8.46   &  \textbf{28}    \\
+Annealing + BN  & 20.95  &  337.10  &    21.25  &  9.15   &  25     \\ 
\midrule
\modelname       & 17.18  &  312.45  &    9.39   &  5.32   &  6  \\
+BOW             & 13.98  &  289.94  &    11.59  &  9.25   &  8  \\
+Annealing       & \textbf{12.35}  &  \textbf{239.83}  & 29.47   &  10.78   & 23  \\
+Annealing+BOW   & 12.82  &  249.98  &    \textbf{32.79}   &  \textbf{11.26} & 26 \\
\midrule
\multicolumn{6}{c}{Backbone: GPT-2 medium (24 layers)}   \\ \midrule
Embed            & 18.33  &  317.44  &  2.13  &  1.44   &  3   \\ 
Mem              & 18.30  &  317.24  &  4.47  &  4.26   &  10    \\
Softmax          & 18.47  &  318.80  &  5.80  &  5.03   &  12   \\
\modelname       & \textbf{11.01}  &  \textbf{230.96}  &  \textbf{17.09} &  \textbf{23.69}  &  \textbf{27} \\
\bottomrule
\end{tabular}
}
\caption{Results on Yelp for transformer-based VAE with BOW loss, KL annealing, and batch normalization tricks, and use 24-layer GPT2-medium as the backbone. Here we fix $\gamma$ in batch normalization as 1.} 
\label{tab:annealing}
\end{table}

\subsection{Human Evaluation}
To better verify the effectiveness of \modelname, we also conduct human evaluation on the two conditional generation tasks. For each model, we generated 30 samples on each task, and invite 5 competent annotators to score these samples in terms of three criteria, \textbf{Fluency}, \textbf{Coherence} and \textbf{Novelty} for story generation, and \textbf{Informativeness}, \textbf{Coherence} and \textbf{Novelty} for summarization generation.

As shown in Table \ref{tab:human_eval}, \modelname obtains satisfactory performance in quality, and is consistently superior to all baselines on diversity and novelty. See Appendix \ref{apx_sec:humaneval} for more detailed evaluation protocols.
\subsection{Ablation Study}
\begin{figure*}[]
\centering
\includegraphics[scale=0.5]{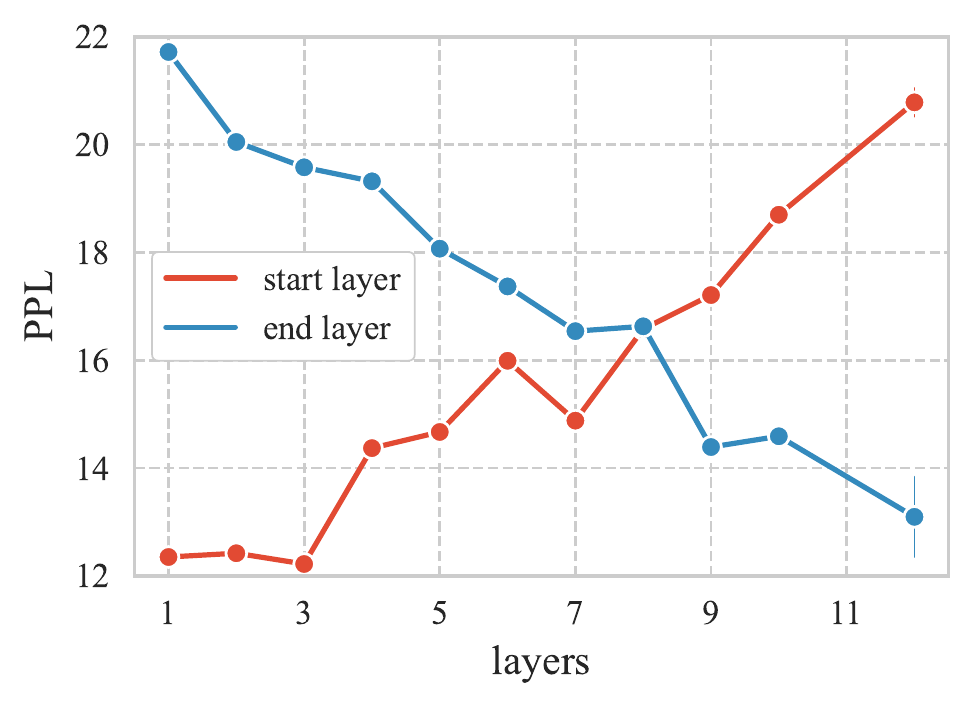} \
\includegraphics[scale=0.5]{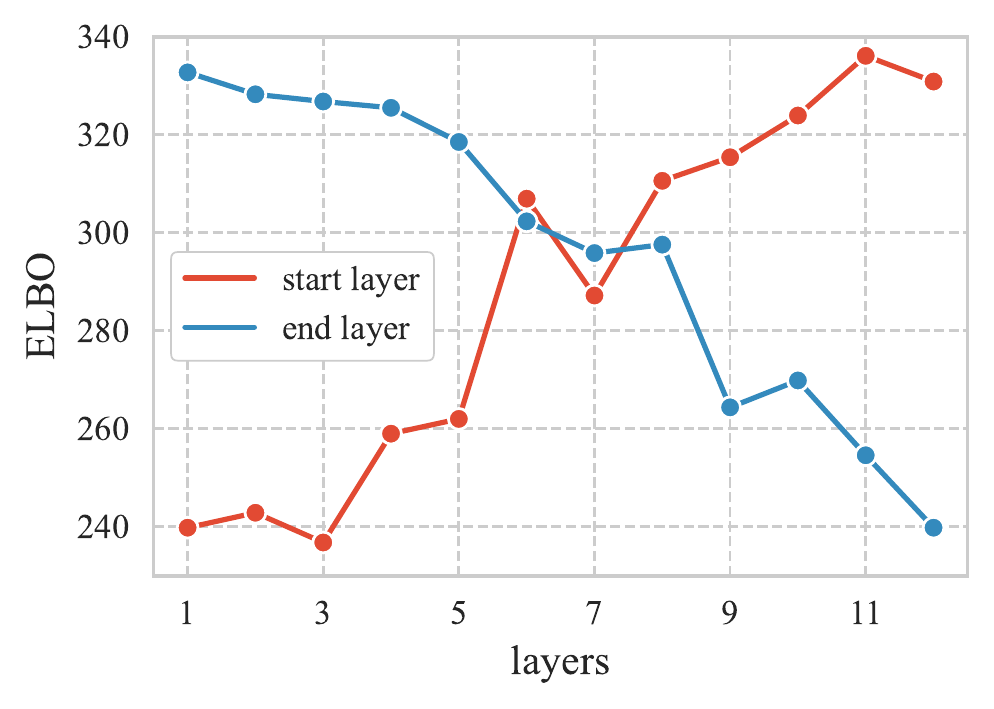} \
\includegraphics[scale=0.5]{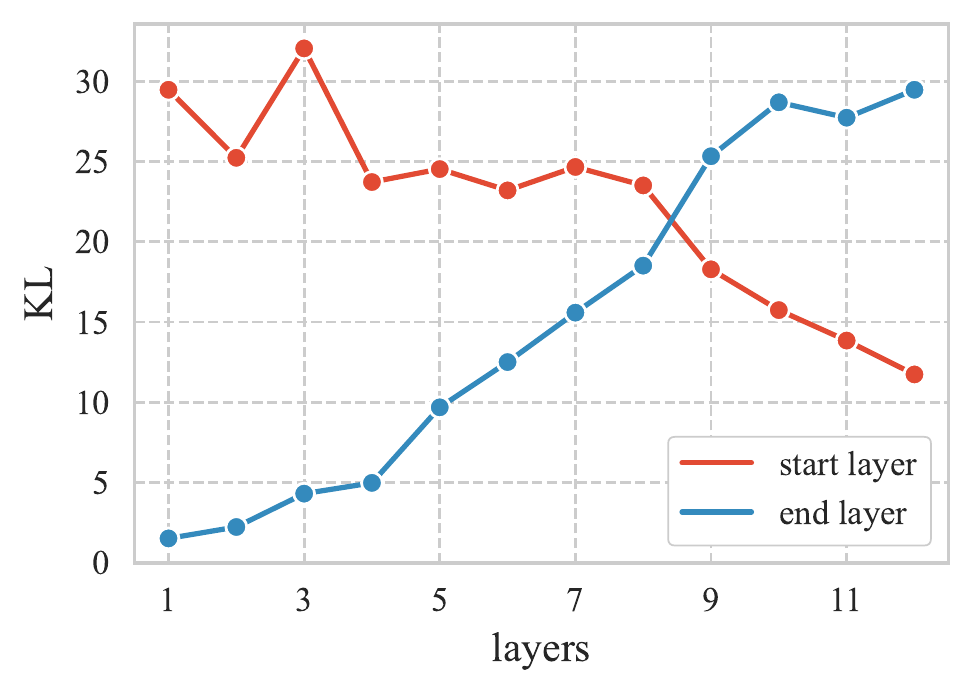} \
\caption{PPL, ELBO end KL on Yelp with different numbers of latent variables. The values \textit{start layer} $i$ and \textit{end layer} $j$ means latent variables are produced and utilized only from $i$-th layer to the last layer, or from the first layer to $j$-th layer of the encoder respectively.}
\label{fig:analysis}
\end{figure*}
Table \ref{tab:ablation_study} shows the results of ablation study on Yelp. We can find both tensor product and the layer-wise latent variables benefit the learning of informative latent variables, while the latter contributes the most to \modelname. To further verify the performance gain originating from Theorem \ref{thm1} instead of simply increasing the number or the dimension of latent variables, we conduct two groups of experiments. 

First, we remove the conditional dependence between layer-wise latent variables by independently sampling each $\boldsymbol{z}_l$ in both training and testing. We can see that removing dependence causes a significant performance drop. Besides, we keep the dependence between $\boldsymbol{z}_l$ but optimize only one of the KL terms in Eq.(\ref{eq:lw_kl}), and find all representation capability metrics deteriorate, especially KL, MI, and AU. Such results effectively demonstrate the necessity of using and optimizing the conditional inference of layer-wise latent variables, supporting our theoretical interpretation of \modelname.

Second, we enlarge the dimension of $\boldsymbol{z}_l$ used in the three paradigms to $384$ ($12 \times 32$), equal to the total latent dimension used in \modelname. The results show that simply increasing the dimension of latent variables brings a more sparse latent space, even exacerbating the KL vanishing problem.

\subsection{Analysis}
\label{sec:analysis}
\textbf{Training Tricks} To reveal the robustness of our model, we evaluate the influence of three commonly used training tricks to relieve KL vanishing, \emph{i.e.}, BOW (bag-of-words) loss \cite{LiweiWang2017DiverseAA}, batch normalization \cite{zhu-etal-2020-batch} and KL annealing \cite{fu-etal-2019-cyclical}, to the performance of \modelname and the three paradigms. As shown in Table \ref{tab:annealing}, previous methods suffer KL vanishing seriously without annealing or BOW loss, getting KL, MI, and AU almost 0. Though not good as using annealing, \modelname still maintains acceptable performance and mitigates KL vanishing even without any training tricks. Bow and batch normalization dramatically prevent low KL divergence, but obstruct the optimization and thus cause higher PPL.

\textbf{Number of Latent Variables} We observe the change of PPL, KL, and ELBO with different numbers of latent variables. We conduct two groups of experiments where we produce and utilize layer-wise latent variables starting from and ending at different layers. As shown in Fig. \ref{fig:analysis}, incorporating more latent variables could continuously improve performance, consistent to our claim in Sec. \ref{sec:model}. With the same number of latent variables, starting from a higher layer is better than ending at a lower layer, which indicates that latent variables generated from higher layers encode more helpful information compared to those from lower layers, manifesting disadvantages of the two previous paradigms, Softmax (starting from the last layer) and Embedding (ending at the first layer).

\textbf{Model size} We compare the performance of \modelname and three paradigms with 24-layer GPT2-medium as the backbone. As shown in Table \ref{tab:annealing}, with the increasing of model size, \modelname consistently achieves better performance than baselines.

\begin{figure*}[]
\centering
\begin{minipage}[t]{0.45\textwidth}
\centering
\includegraphics[scale=0.48]{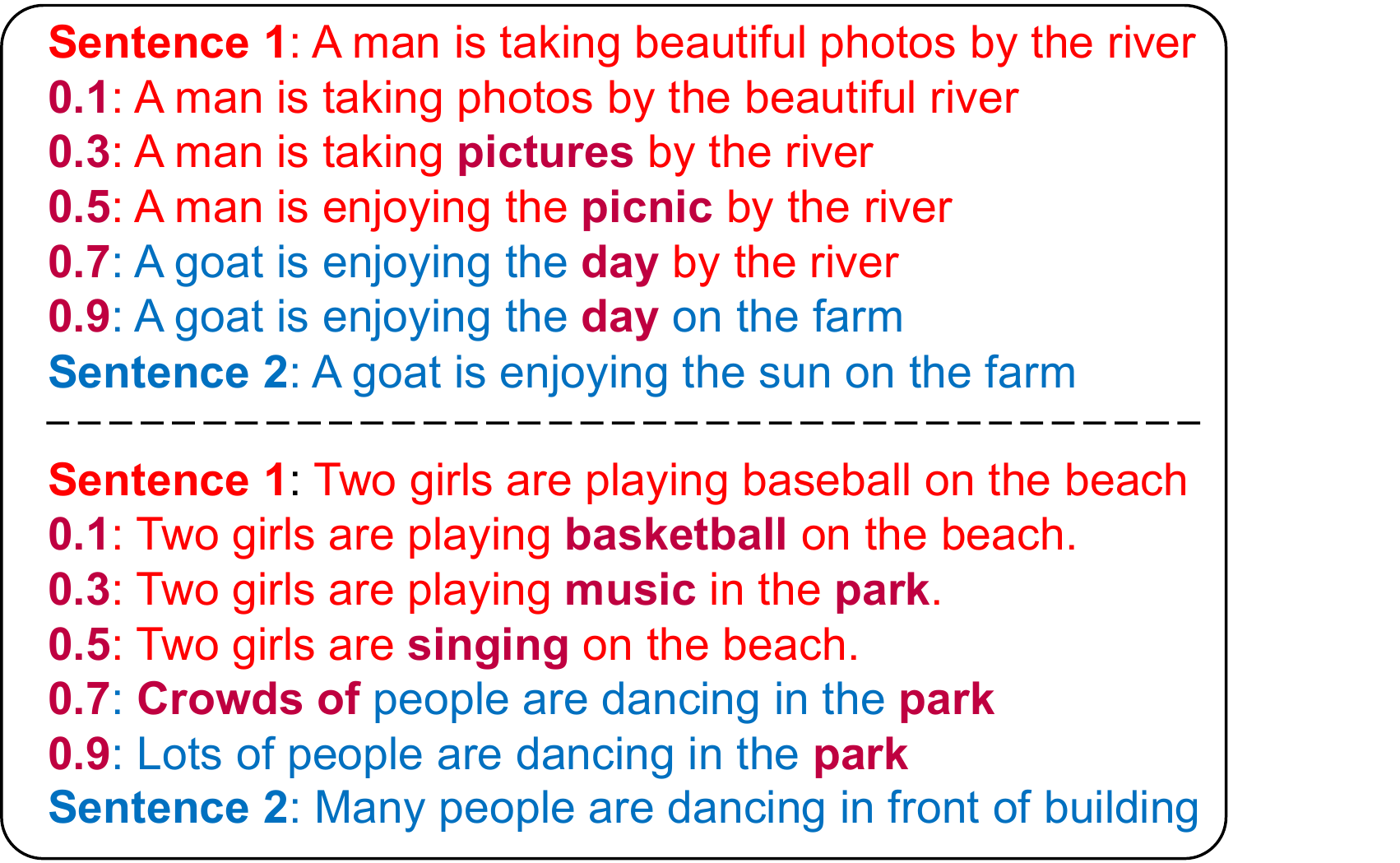}
\caption{Interpolating latent space. The sentence in each row is generated with a latent variable interpolated from those of sentence 1 and sentence 2.}
\label{fig:interpolate}
\end{minipage}
\hfill
\begin{minipage}[t]{0.52\textwidth}
\includegraphics[scale=0.48]{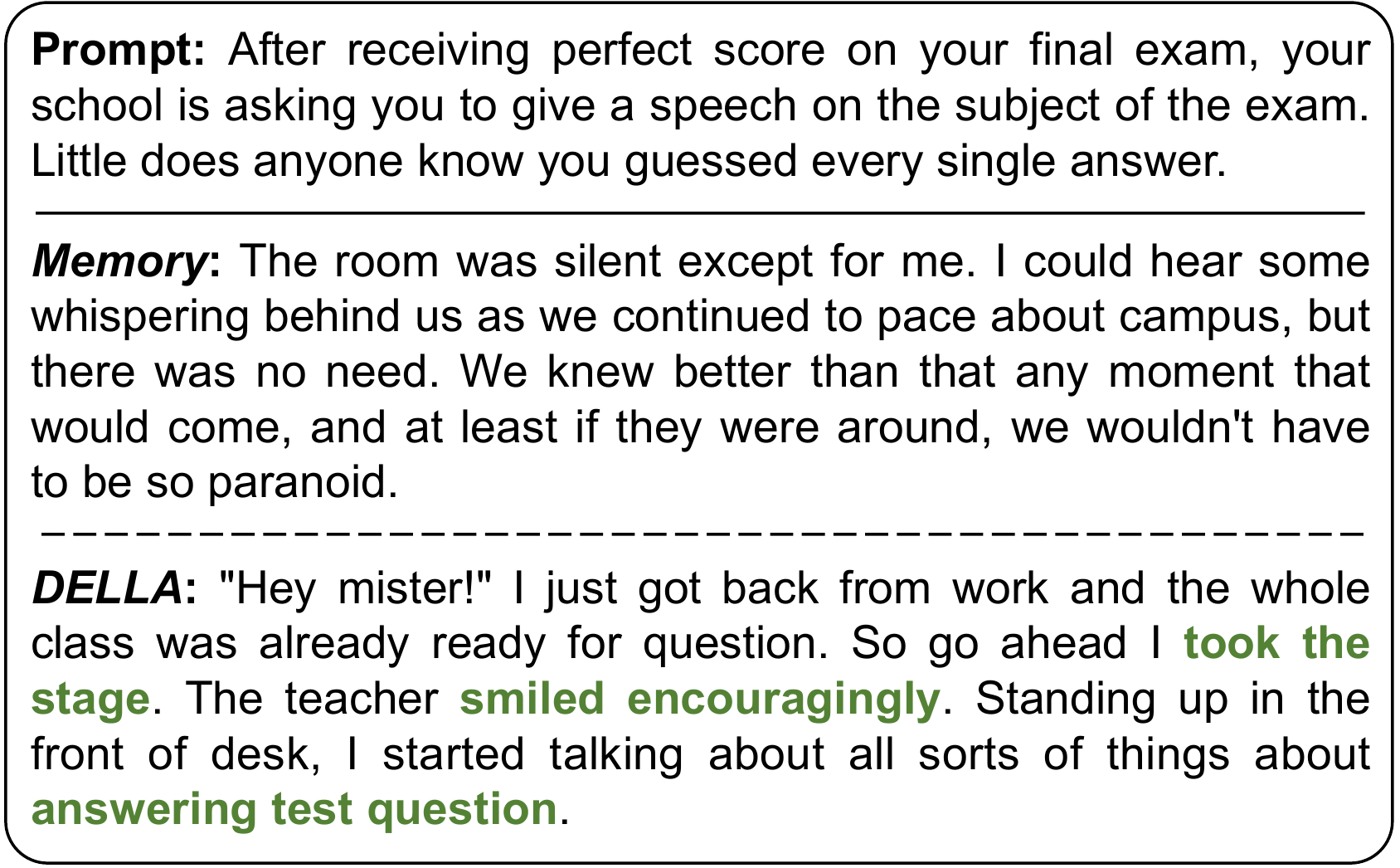}
\caption{Generation examples of Memory and \modelname based on the prompt from test set of WritingPrompts.}
\label{fig:story_case}
\end{minipage}
\end{figure*}

\subsection{Case Study}
VAE captures text representations in smooth latent space. We take two sentences $\boldsymbol{x}_1$ and $\boldsymbol{x}_2$ and sample two posterior latent variables $\boldsymbol{z}^{(1)}$ and $\boldsymbol{z}^{(2)}$ from $p(\boldsymbol{z}^{(1)}|\boldsymbol{x}_1)$ and $p(\boldsymbol{z}^{(2)}|\boldsymbol{x}_2)$, and get interpolated latent variables with $\boldsymbol{z}=\tau\boldsymbol{z}^{(1)}+(1-\tau)\boldsymbol{z}^{(2)}$. We generate multiple sentences with a continuously changed $\tau$ from 0 to 1. As shown in Fig. \ref{fig:interpolate}, sentences generated from interpolated $\boldsymbol{z}$ mix the semantics of the two initial sentences and smoothly change from $\boldsymbol{x}_1$ to $\boldsymbol{x}_2$, showing \modelname's ability of learning a flexible latent space.

Fig. \ref{fig:story_case} shows the generation examples of \modelname and one of baseline, Memory, given the same prompt WritingPrompts. We observe that the generated text of Memory is irrelevant to the prompt, while \modelname generates coherent and vivid text.

\section{Conclusion}
In this paper, we propose a novel variational Transformer framework \modelname. Our framework learns a series of layer-wise latent variables with iterative dependence. These latent variables are conditionally inferred and injected into corresponding decoder layers by low-rank tensor product for deeper fusion. The experiments on both unconditional and conditional generation tasks demonstrate \modelname's ability to significantly mitigate KL vanishing and improve generated text's quality and diversity. In the future, we plan to explore further the potential of \modelname in larger pretrained language models. 

\section*{Acknowledgement}
Thanks for the anonymous reviewers for their comments. This work is supported by the National Key R\&D Program of China (No. 2020AAA0106502) and International Innovation Center of Tsinghua University, Shanghai, China.

\bibliography{anthology,custom}
\bibliographystyle{acl_natbib}
\newpage
\appendix
\section{Experiment Details}
\subsection{Implementation Details}
\label{apx_sec:implemet}
We load pretrained model GPT-2 \citep{Radford2019LanguageMA} as initial parameters for unconditional generation and story generation, and pretrained BART-base \citep{lewis-etal-2020-bart} for summarization and paraphrasing generation tasks. For the summarization and paraphrasing generation, we keep the encoder-decoder attention block. No encoder-decoder attention is used in unconditional generation and story generation tasks. The number of layers and dimensions of hidden states in \modelname is consistent with the configurations of corresponding pretrained models (GPT-2 has 12 layers and Bart-base has 6-layer encoder and 6-layer decoder. The hidden size of both is 768). We use the state of a special token to obtain the representation in the encoder. We utilize cyclical annealing tricks to train \modelname and other VAE baselines. Specifically, two epochs are one annealing period. In one period, $\beta$ (the weight of KL term in ELBO) keeps 1e-5 in the first half, then linearly increases to 1 in the next quarter, then keeps at 1 for the last quarter. We select batch size over $\{16, 32\}$ and learning rate over $\{\text{5e-5, 7e-5}\}$. We use beam search for \modelname and top-k sampling for compared baseline models for the unconditional generation and story generation. For the summarization and paraphrasing generation, we use beam search in all the models.

We implement \modelname and other VAE baselines based on Huggingface Transformers \citep{wolf-etal-2020-transformers} library of v4.10.0 and use NVIDIA GeForce RTX 3090 to train our model. The total number of training GPU hours on different datasets is in Table ~\ref{tab:GPU Hours}. The number of parameters for our model is 193,353,984 in the unconditional generation setting and 195,180,114 in the conditional generation one. All experimental results are trained and tested in a single run.

\begin{table}[htbp]
\centering
\scalebox{0.9}{
\begin{tabular}{l|c}
\toprule
Dataset & Training Time \\\midrule
Yelp   & 20h \\ 
Yahoo  & 20h \\ 
PTB    & 6h  \\ 
SNLI   & 12h \\ 
CNN/DM & 40h \\ 
WP     & 170h \\ 
Quora  & 5h  \\ \bottomrule
\end{tabular}}
\caption{GPU hours of training \modelname with RTX3090}
\label{tab:GPU Hours}
\end{table}

\subsection{Datasets Details}
The detailed dataset statistics are in Table ~\ref{tab:stat_dataset}. For the licenses of the datasets we use, CNN/DM and WritingPrompts use MIT License, while SNLI uses CC BY-SA 4.0. Meanwhile, PTB, Quora, and Yelp use their own license: LDC User Agreement, Yelp Data Agreement, and Quora's Terms of Service, respectively. All of these licenses and agreements allow their data for academic use. Unfortunately, we did not find the license for the Yahoo Dataset.

\begin{table}[]
\centering
\scalebox{0.85}{
\begin{tabular}{l|c|c|c|cc}

\toprule
Dataset        & \# Train     & \# Dev   & \# Test   & \multicolumn{2}{c}{Avarage Length}      \\ \toprule
Yelp           & 100k   & 10k & 10k & \multicolumn{2}{c}{96}                  \\ 
Yahoo          & 100k   & 10k & 10k & \multicolumn{2}{c}{79}                  \\ 
PTB            & 42k    & 3k  & 3k  & \multicolumn{2}{c}{21}                  \\ 
SNLI           & 100k  & 10k & 10k & \multicolumn{2}{c}{10}                  \\ \midrule
CNN/DM         & 287k   & 13k & 11k & \multicolumn{1}{c|}{S: 790} & T: 61  \\ 
WP & 272k   & 15k & 15k & \multicolumn{1}{c|}{S: 28}  & T: 674 \\ 
Quora          & 134k   & 5k  & 10k & \multicolumn{1}{c|}{S: 10}  & T: 10  \\ \bottomrule
\end{tabular}
}
\caption{\textbf{Statistics of datasets}. We present the size of train/dev/test sets and the average length for 7 datasets. S means source text and T means target text. }
\label{tab:stat_dataset}
\end{table}

\subsection{Metrics Details}
\label{apx_sec:metrics}
Here we provide more details of the metrics used in our experiments.

\textbf{Perplexity (PPL)}. $\text{PPL} = p(\boldsymbol{x})^{-1/n}$ is commonly used to evaluate the performance of language models, where $n$ is number of tokens $\boldsymbol{x}$ contains. For VAE-based model, we can only obtain the lower bound of $\log p(\boldsymbol{x})$. We consider $k$ latent variables $z_1, z_2, \dots, z_k$ sampled from the posterior distribution $q(\boldsymbol{z}_i|\boldsymbol{x})$. Based on the fact that average importance weights are an unbiased estimator of $\log p(\boldsymbol{x})$ \citep{YuriBurda2016ImportanceWA} and Jensen’s Inequality, we have:
\begin{align}
    \mathcal{L}_{k} &=\mathbb{E}\left[\log \frac{1}{k} \sum_{i=1}^{k} \frac{p(\boldsymbol{x},             \boldsymbol{z}_i)}{q(\boldsymbol{z}_i|\boldsymbol{x})}\right] \\
                    &\leq \log \mathbb{E}\left[\frac{1}{k} \sum_{i=1}^{k} \frac{p(\boldsymbol{x}, \boldsymbol{z}_i)}{q(\boldsymbol{z}_i|\boldsymbol{x})}\right]=\log p(\boldsymbol{x}). \nonumber
\end{align}
We use $\mathcal{L}_{k}$ to estimate $\log p(\boldsymbol{x})$ and calculate PPL.
    
\textbf{ELBO}. The ELBO is the sum of reconstruction loss and KL divergence.

\textbf{KL}. The KL divergence of the posterior and prior distribution.
    
\textbf{Mutual Information(MI)} \citep{AlexanderAAlemi2016DeepVI}. Mutual Information $\mathcal{I}(\boldsymbol{x}, \boldsymbol{z})$ is defined as:
\begin{align}
&\mathcal{I}_{q}(\boldsymbol{x}, \boldsymbol{z}) \\
=&\mathbb{E}_{p(\boldsymbol{x})} \mathbb{E}_{q(\boldsymbol{z}|\boldsymbol{x})} \log q(\boldsymbol{z}|\boldsymbol{x})-\mathbb{E}_{q(\boldsymbol{z})} \log q(\boldsymbol{z}) \nonumber
\end{align}
where $q_{(\boldsymbol{z})} = \mathbb{E}_{p(\boldsymbol{x})}q(\boldsymbol{z}|\boldsymbol{x})$ is called the aggregated posterior.

\textbf{Activate Units(AU)} \citep{YuriBurda2016ImportanceWA}. AU is the active units in latent variables, defined as $A_{\boldsymbol{z}} = \text{Cov}_{\boldsymbol{x}}(\mathbb{E}_{z\sim q(z|\boldsymbol{x})}[z]) > \delta$, where $\delta$ is a threshold, commonly set as 0.01. However, we find that with $\delta=0.01$, all VAE models in our experiments have full active unitsunconditional. So we increase the threshold to 0.2 to distinguish the performance of different models on this metric. \textbf{Please note} that \modelname incorporates latent variables in all layers, and hence we calculate AU for the latent variable in each layer and then report the average.

\textbf{BLEU} \citep{papineni-etal-2002-bleu}. BLEU measures the n-gram overlap of generated sequences and the reference ones. For the unconditional setting, we regard all samples in the test set as references to each generated example.

\textbf{CND} \citep{JianingLi2020OnTR}. CND approximates the divergence of the empirical reference distribution and generated text distribution in n-gram spaces.

\textbf{MAUVE} \citep{pillutla-etal:mauve:neurips2021}. MAUVE measures the gap between reference text and generated text using divergence frontiers.

\textbf{Self-BLEU} \citep{YaomingZhu2018TexygenAB}. Self-Bleu calculates the BLEU score on the generated samples, which averages the BLEU score of each generated sequence calculated with other generated ones as references. This metric measures the diversity of a set of generated sequences. Higher Self-BLEU means these generated sequences are more distinguishable from each other.

\textbf{Dist} \citep{li-etal-2016-diversity}. Dist measures the proportion of distinct n-grams on generated samples.

\textbf{Jaccard Similarity(JS)} \citep{KeWang2018SentiGANGS}. JS calculates the average n-gram Jaccard similarity between every two generated sequences.

\textbf{Rouge} \citep{lin-hovy-2002-manual}. Rouge computes the n-gram overlap of generated examples with given target samples. We use rouge-score v0.0.4 to evaluate the rouge score of our model and the baselines.

\textbf{BERTScore} \citep{TianyiZhang2020BERTScoreET}. BERTScore uses pre-trained BERT \cite{devlin-etal-2019-bert} to obtain the vector representations of generated and reference text and calculates their cosine similarity. We use bert-score v0.3.10 to calculate the BERTScore of our model and the baselines.

\subsection{Human Evaluation Details}
\label{apx_sec:humaneval}
Due to the relatively long length of generated text, we randomly sample 30 examples in the test set of WP and CNN/DM as input to \modelname and other compared baseline models to generate the target. We invite five graduate students proficient in English to score the generated text. The criteria for story generation include fluency, coherence, and novelty, and the criteria for summarization generation include informativeness, consistency, and novelty. Specifically, fluency measures whether the generated sentences are syntactically fluent; coherence measures whether the generated text is logically structured and consistent with the input text; novelty measures whether the content is novel and attractive; informativeness measures to what extent the generated summarization summarizes the general idea of the article. 

% When conducting the human evaluation, we informed the participants as follows: 
% \begin{itemize}
%     \item The following contents are generated by the automatic models. Some of them may be offensive or contain improper arguments. Please be conscious of these risks and evaluate these contents equitably and adequately.
% 	\item The evaluation you provide will be used only for academic use and will never be used commercially.
% \end{itemize}
% Every evaluator will sign their signature below these warnings to confirm that they have read those words. After finishing the annotation, they will receive \$25. This amount is determined by the time of the whole annotation process and the estimation of average hourly income. The ethics review board for data collection protocol is not essential in our country, so we did not conduct this review for our data collection protocol.

\subsection{Additional Experimental Result}
\label{apx_sec:results}
Table \ref{tab:uncond_task_add} and Table \ref{tab:quora_res} report the results on PTB, SNLI and Quora dataset.%, which give the consistent conclusion as the results on the datasets reported in the the body text of our paper.

% We further make a comparison on decoding algorithms between sampling and beam search, which is a trade-off of quality of diversity. We report the results in Table \ref{tab:decoding}. With beam search, although sampling from prior can brings randomness to the model, only one latent variable is limited. Therefore, three baseline VAE models get poor performance, especially on diversity, with beam search decoding. Only no more than ten plain sentences will these model generate after removing duplicate. So sampling method is the best choice for three baseline VAE models (only choice for GPT-2). 

% In contrast, \modelname can generate sentences with high quality and diversity with beam search. While sampling will further improve diversity, quality will distinctly drop. Therefore, beam search is the best choice for \modelname. 
\subsection{Case Study Details}
We take two sentences $\boldsymbol{x}_1$ and $\boldsymbol{x}_2$ and sample two groups of latent variables $\boldsymbol{z}^{(1)}= \{\boldsymbol{z}^{(1)}_{1}, \boldsymbol{z}^{(1)}_{2}, \dots, \boldsymbol{z}^{(1)}_{L}\}$ and $\boldsymbol{z}^{(2)}= \{\boldsymbol{z}^{(2)}_{1}, \boldsymbol{z}^{(2)}_{2}, \dots, \boldsymbol{z}^{(2)}_{L}\}$ from posterior distributions $p(\boldsymbol{z}^{(1)}|\boldsymbol{x}_1)$ and $p(\boldsymbol{z}^{(2)}|\boldsymbol{x}_2)$. We obtain the weighted latent variables $\hat{\boldsymbol{z}} = \{\hat{\boldsymbol{z}}_1, \hat{\boldsymbol{z}}_2, \dots, \hat{\boldsymbol{z}}_L\}$ by taking weighted sum at each corresponding element in two groups, i.e. $\hat{\boldsymbol{z}}_i = \tau * {\boldsymbol{z}}^{(1)}_i + (1 - \tau) * {\boldsymbol{z}}^{(2)}_i$. The mixed sentence $\hat{\boldsymbol{x}}$ is generated conditioned on $p(\hat{\boldsymbol{x}}|\hat{\boldsymbol{z}})$ by the decoder.

% \subsection{Potential Risks and Limitations of our work}
% Due to the unclean corpus (especially in the WP dataset) we use where slang repeatedly appears, the model training on this corpus may also output some rude expressions during generation. Also, the text generated in the unconditional generation task is not controllable, which may contain some bias or politically sensitive expression. Besides, since our model significantly improves the quality and diversity of generated, it can produce more plausible texts like news, which could be possibly utilized to create fake news or disinformation. However, on the other hand, our model could benefit fairness in language generation. Previous text generation models tend to produce biases like gender or nationality biases, which means only the majority would be appropriately described while the minority may be ignored. These biases are mainly caused by the biased training corpus. With the same data, our model can improve the diversity of generated text, which is also potential for mitigating these biased. We will try to develop debiased language generation systems in future work to avoid these risks harming society.

% While \modelname shows good performance on text generation, it has one limitation: training efficiency. \modelname brings more parameters compared with three baseline methods. Training efficiency needs to be considered if we further explore the performance of \modelname on the large pretrained model.

\begin{table*}[]
\centering
\scalebox{0.9}{
\begin{tabular}{cccccccccccc}
\toprule
\multicolumn{1}{c|}{\multirow{2}{*}{Model}} & \multicolumn{5}{c|}{Representation   Learning} & \multicolumn{3}{c|}{Generation Quality} & \multicolumn{3}{c}{Generation   Diversity} \\ \cline{2-12} 
\multicolumn{1}{c|}{}  & \multicolumn{1}{c|}{PPL$\downarrow$}   &\multicolumn{1}{c|}{ELBO$\downarrow$}   & \multicolumn{1}{c|}{KL$\uparrow$}    & \multicolumn{1}{c|}{MI$\uparrow$}    & \multicolumn{1}{c|}{AU$\uparrow$} & \multicolumn{1}{c|}{BLEU$\uparrow$}  & \multicolumn{1}{c|}{CND$\downarrow$} &  \multicolumn{1}{c|}{MAUVE$\uparrow$} &  \multicolumn{1}{c|}{SB$\downarrow$}      & \multicolumn{1}{c|}{Dist$\uparrow$} & JS $\downarrow$     \\ \midrule
\multicolumn{12}{c}{Dataset: PTB}                                                                                                                                                                                                                                                                \\ \midrule
\multicolumn{1}{c|}{GPT-2}                   & \multicolumn{1}{c|}{25.80}  & \multicolumn{1}{c|}{-}       & \multicolumn{1}{c|}{-}      & \multicolumn{1}{c|}{-}      & \multicolumn{1}{c|}{-}   & \multicolumn{1}{c|}{27.91}    & \multicolumn{1}{c|}{1.12}      & \multicolumn{1}{c|}{\textbf{0.73}} & \multicolumn{1}{c|}{41.55}  & \multicolumn{1}{c|}{37.79}      &       0.30       \\ 
\multicolumn{1}{c|}{Optimus}                & \multicolumn{1}{c|}{22.79}   & \multicolumn{1}{c|}{344.10}   & \multicolumn{1}{c|}{15.09}      & \multicolumn{1}{c|}{7.67}      & \multicolumn{1}{c|}{-}       & \multicolumn{1}{c|}{-}      & \multicolumn{1}{c|}{-}      & \multicolumn{1}{c|}{-}   & \multicolumn{1}{c|}{-}  & \multicolumn{1}{c|}{-}      &    -          \\ 
\multicolumn{1}{c|}{Embed}                  & \multicolumn{1}{c|}{19.98} & \multicolumn{1}{c|}{327.28} & \multicolumn{1}{c|}{4.77}  & \multicolumn{1}{c|}{4.14}  & \multicolumn{1}{c|}{6}   & \multicolumn{1}{c|}{28.04}& \multicolumn{1}{c|}{1.38}      & \multicolumn{1}{c|}{0.69}      & \multicolumn{1}{c|}{41.32}   & \multicolumn{1}{c|}{34.46}     &  0.33            \\
\multicolumn{1}{c|}{Memory}                    & \multicolumn{1}{c|}{24.41} & \multicolumn{1}{c|}{90.25}  & \multicolumn{1}{c|}{1.22}   & \multicolumn{1}{c|}{1.17}   & \multicolumn{1}{c|}{4}   & \multicolumn{1}{c|}{21.31}& \multicolumn{1}{c|}{1.21} & \multicolumn{1}{c|}{0.58}      & \multicolumn{1}{c|}{26.58}    & \multicolumn{1}{c|}{38.28}    &    \textbf{0.08}          \\ 
\multicolumn{1}{c|}{Softmax}                & \multicolumn{1}{c|}{24.04} & \multicolumn{1}{c|}{90.63} & \multicolumn{1}{c|}{2.13}   & \multicolumn{1}{c|}{1.89}  & \multicolumn{1}{c|}{21}   & \multicolumn{1}{c|}{\textbf{28.59}}& \multicolumn{1}{c|}{1.39} & \multicolumn{1}{c|}{0.72}  & \multicolumn{1}{c|}{42.15}     & \multicolumn{1}{c|}{33.91}   &    0.30         \\ 
\multicolumn{1}{c|}{\modelname}                  & \multicolumn{1}{c|}{\textbf{10.28}} & \multicolumn{1}{c|}{\textbf{58.43}} & \multicolumn{1}{c|}{\textbf{12.46}} & \multicolumn{1}{c|}{\textbf{12.35}} & \multicolumn{1}{c|}{\textbf{22}}   & \multicolumn{1}{c|}{28.15} & \multicolumn{1}{c|}{\textbf{0.63}}  & \multicolumn{1}{c|}{0.68}  & \multicolumn{1}{c|}{\textbf{24.87}}   & \multicolumn{1}{c|}{\textbf{41.84}}     &    0.17          \\ \midrule
\multicolumn{12}{c}{Dataset: SNLI}                                                                                                                                                                                                                                                               \\ \midrule
\multicolumn{1}{c|}{GPT-2}                   & \multicolumn{1}{c|}{20.19}   & \multicolumn{1}{c|}{-}       & \multicolumn{1}{c|}{-}      & \multicolumn{1}{c|}{-}      & \multicolumn{1}{c|}{-} & \multicolumn{1}{c|}{\textbf{63.57}}   & \multicolumn{1}{c|}{1.95}      & \multicolumn{1}{c|}{0.71}         & \multicolumn{1}{c|}{75.34}    & \multicolumn{1}{c|}{19.11}    &    0.58         \\ 
\multicolumn{1}{c|}{Optimus}                & \multicolumn{1}{c|}{16.67}    & \multicolumn{1}{c|}{38.50}  & \multicolumn{1}{c|}{16.35}      & \multicolumn{1}{c|}{8.89}      & \multicolumn{1}{c|}{-}       & \multicolumn{1}{c|}{-}      & \multicolumn{1}{c|}{-}      & \multicolumn{1}{c|}{-}   & \multicolumn{1}{c|}{-}    & \multicolumn{1}{c|}{-}    & -             \\
\multicolumn{1}{c|}{Embed}                  & \multicolumn{1}{c|}{13.79} & \multicolumn{1}{c|}{32.97} & \multicolumn{1}{c|}{3.24}  & \multicolumn{1}{c|}{3.16}  & \multicolumn{1}{c|}{20}   & \multicolumn{1}{c|}{59.26} & \multicolumn{1}{c|}{0.98}      & \multicolumn{1}{c|}{\textbf{0.72}}      & \multicolumn{1}{c|}{65.59}    & \multicolumn{1}{c|}{20.89}    &      0.44        \\ 
\multicolumn{1}{c|}{Memory}                    & \multicolumn{1}{c|}{13.78} & \multicolumn{1}{c|}{32.62} & \multicolumn{1}{c|}{2.13}  & \multicolumn{1}{c|}{2.08}  & \multicolumn{1}{c|}{10}   & \multicolumn{1}{c|}{62.80} & \multicolumn{1}{c|}{1.24}      & \multicolumn{1}{c|}{0.67}  & \multicolumn{1}{c|}{54.59}    & \multicolumn{1}{c|}{21.87}    &     0.33            \\ 
\multicolumn{1}{c|}{Softmax}                & \multicolumn{1}{c|}{14.21} & \multicolumn{1}{c|}{33.18} & \multicolumn{1}{c|}{2.70}  & \multicolumn{1}{c|}{2.65}  & \multicolumn{1}{c|}{16}   & \multicolumn{1}{c|}{60.51} & \multicolumn{1}{c|}{1.94}      & \multicolumn{1}{c|}{0.71}      & \multicolumn{1}{c|}{71.84}    & \multicolumn{1}{c|}{18.59}    &    0.57         \\ 
\multicolumn{1}{c|}{\modelname}                  & \multicolumn{1}{c|}{\textbf{5.13}} & \multicolumn{1}{c|}{\textbf{10.23}} & \multicolumn{1}{c|}{\textbf{5.86}} & \multicolumn{1}{c|}{\textbf{16.58}}      & \multicolumn{1}{c|}{\textbf{23}}& \multicolumn{1}{c|}{62.94} & \multicolumn{1}{c|}{\textbf{0.85}}      & \multicolumn{1}{c|}{0.69}         & \multicolumn{1}{c|}{\textbf{36.85}} & \multicolumn{1}{c|}{\textbf{32.61}}    &    \textbf{0.21}       \\ \bottomrule
\end{tabular}
}
\caption{Additional results for language model and unconditional generation task. The results of Optimus are copied from original paper with $\lambda=0.5$.}
\label{tab:uncond_task_add}
\end{table*}

\begin{table*}
\centering
\scalebox{0.95}{
\begin{tabular}{l|c|c|c|c|c|c}
\toprule
Model            & BLEU$\uparrow$ & Rouge-1$\uparrow$ & Rouge-2$\uparrow$ & Rouge-L$\uparrow$ & Bertscore$\uparrow$ & KL$\uparrow$ \\ \midrule
Bart-base        & 64.34   & 63.27     & 39.83        & 60.28       & 94.72  & -\\
Embed            & 63.94   & 63.12     & 39.42        & 60.22       & 94.66  & 0.0 \\ 
Mem              & 63.78   & 62.86     & 39.18        & 59.96       & 94.65  & 0.0  \\ 
Softmax          & 64.30   & 63.25     & 39.92        & 60.39       & 94.71  & 0.0 \\ 
\modelname       & \textbf{64.40}      & \textbf{63.80}        & \textbf{40.58}   & \textbf{61.03} &  \textbf{94.84} & \textbf{3.88} \\ \bottomrule
\end{tabular}}
\caption{Results on Quora dataset. Because the sentences in Quora are quite short and constrained, the results of the three diversity metrics on all baselines are almost the same. So we omit them here.} 
\label{tab:quora_res}
\end{table*}

\newpage
\onecolumn
\section{Additional Proof}
\subsection{Derivation of KL Divergence of Layer-Wise Latent Variables}
\label{apx_sec: proof}
KL divergence of layer-wise latent variables
\begin{equation}
    \begin{aligned}
    &KL(q(\boldsymbol{z}|\boldsymbol{x})||p(\boldsymbol{z})) \\
   =&\int q(\boldsymbol{z}|\boldsymbol{x})\log \frac{q(\boldsymbol{z}|\boldsymbol{x})}{p(\boldsymbol{z})}\dif\boldsymbol{z}   \\
   =&\int \prod_{l=1}^{L}q(\boldsymbol{z}_l|\boldsymbol{x}, \boldsymbol{z}_{<l}) \log\frac{\prod_{l=1}^{L}q(\boldsymbol{z}_l|\boldsymbol{x}, \boldsymbol{z}_{<l})}{\prod_{l=1}^{L}p(\boldsymbol{z}_l| \boldsymbol{z}_{<l})} \dif\boldsymbol{z}_1\dif\boldsymbol{z}_2\dots\dif\boldsymbol{z}_L\\
   =&\sum_{i=1}^{L}\int \prod_{l=1}^{L}q(\boldsymbol{z}_l|\boldsymbol{x}, \boldsymbol{z}_{<l}) \log \frac{q(\boldsymbol{z}_l|\boldsymbol{x}, \boldsymbol{z}_{<l})}{p(\boldsymbol{z}_l| \boldsymbol{z}_{<l})}\dif\boldsymbol{z}_1\dif\boldsymbol{z}_2\dots\dif\boldsymbol{z}_L \\
   =&\sum_{l=1}^{L}\int q(\boldsymbol{z}_{<l}|\boldsymbol{x}) q(\boldsymbol{z}_l|\boldsymbol{x}, \boldsymbol{z}_{<l}) \log \frac{q(\boldsymbol{z}_l|\boldsymbol{x}, \boldsymbol{z}_{<l})}{p(\boldsymbol{z}_l| \boldsymbol{z}_{<l})}\dif\boldsymbol{z}_1\dif\boldsymbol{z}_2\dots\dif\boldsymbol{z}_l  \\
   =&\sum_{l=1}^{L}\mathbb{E}_{q(\boldsymbol{z}_{<l}|\boldsymbol{x})} \mathrm{KL}(q(\boldsymbol{z}_l|\boldsymbol{x}, \boldsymbol{z}_{<l}) || p(\boldsymbol{z}_l| \boldsymbol{z}_{<l}))
\end{aligned}
\end{equation}
\subsection{Proof of Theorem 1}
\label{proof:theorem1}
First, we consider on term in the summation and can obtain:

\begin{equation}
    \begin{aligned}
    &\mathbb{E}_{p(\boldsymbol{x})}\mathbb{E}_{q(\boldsymbol{z}_{<l}|\boldsymbol{x})}[\mathrm{KL}(q(\boldsymbol{z}|\boldsymbol{x}, \boldsymbol{z}_{<l})||p(\boldsymbol{z}_l|\boldsymbol{z}_{<l}))] \\ 
    =& \int q(\boldsymbol{x}) q(\boldsymbol{z}_{<l}|\boldsymbol{x}) q(\boldsymbol{z}_{l}|\boldsymbol{x}, \boldsymbol{z}_{<l})\log\frac{q(\boldsymbol{z}_{l}|\boldsymbol{x}, \boldsymbol{z_l})}{p(\boldsymbol{z}_l|\boldsymbol{z}_{<l})} \dif\boldsymbol{x}\dif\boldsymbol{z}_{l}\dif\boldsymbol{z}_{<l} \\
    =& \int q(\boldsymbol{x}, \boldsymbol{z}_l, \boldsymbol{z}_{<l})\log\frac{q(\boldsymbol{z}_{l}|\boldsymbol{x}, \boldsymbol{z_l})}{p(\boldsymbol{z}_l|\boldsymbol{z}_{<l})} \dif\boldsymbol{x}\dif\boldsymbol{z}_{l}\dif\boldsymbol{z}_{<l}  \\
    =&  \int q(\boldsymbol{x}, \boldsymbol{z}_l, \boldsymbol{z}_{<l})\log\Big(\frac{q(\boldsymbol{z}, \boldsymbol{x}|\boldsymbol{z}_{<l})}{q(\boldsymbol{x}|\boldsymbol{z}_{<l})q(\boldsymbol{z}_l|\boldsymbol{z}_{<l})} \frac{q(\boldsymbol{z}_l|\boldsymbol{z}_{<l})}{p(\boldsymbol{z}_l|\boldsymbol{z}_{<l})}\Big) \dif\boldsymbol{x}\dif\boldsymbol{z}_{l}\dif\boldsymbol{z}_{<l} \\
    =& \int q(\boldsymbol{z}_{<l})q(\boldsymbol{x}, \boldsymbol{z}_l|\boldsymbol{z}_{<l})\log\frac{q(\boldsymbol{z}, \boldsymbol{x}|\boldsymbol{z}_{<l})}{q(\boldsymbol{x}|\boldsymbol{z}_{<l})q(\boldsymbol{z}_l|\boldsymbol{z}_{<l})}\dif\boldsymbol{x}\dif\boldsymbol{z}_{l}\dif\boldsymbol{z}_{<l} + \\
    &  \int q(\boldsymbol{x}|\boldsymbol{z}_l, \boldsymbol{z}_{<l})q(\boldsymbol{z}_l|\boldsymbol{z}_{<l}) q(\boldsymbol{z}_{<l}) \log\frac{q(\boldsymbol{z}_l|\boldsymbol{z}_{<l})}{p(\boldsymbol{z}_l|\boldsymbol{z}_{<l})}\dif\boldsymbol{x}\dif\boldsymbol{z}_{l}\dif\boldsymbol{z}_{<l}  \\
    =& \int q(\boldsymbol{x}, \boldsymbol{z}_l|\boldsymbol{z}_{<l})\log\frac{q(\boldsymbol{z}, \boldsymbol{x}|\boldsymbol{z}_{<l})}{q(\boldsymbol{x}|\boldsymbol{z}_{<l})q(\boldsymbol{z}_l|\boldsymbol{z}_{<l})}\dif\boldsymbol{x}\dif\boldsymbol{z}_{l} + \\
    &\int q(\boldsymbol{z}_l|\boldsymbol{z}_{<l}) q(\boldsymbol{z}_{<l}) \log\frac{q(\boldsymbol{z}_l|\boldsymbol{z}_{<l})}{p(\boldsymbol{z}_l|\boldsymbol{z}_{<l})}\dif\boldsymbol{z}_{l}\dif\boldsymbol{z}_{<l}  \\
    =&H(\boldsymbol{z}_l|\boldsymbol{z}_{<l}) - H(\boldsymbol{z}_l|\boldsymbol{z}_{<l}, \boldsymbol{x}) + \mathbb{E}_{q(\boldsymbol{z}_{<l})}\mathrm{KL}(q(\boldsymbol{z}_l|\boldsymbol{z}_{<l}||p(\boldsymbol{z}_l|\boldsymbol{z}_{<l})) \\
    \ge & H(\boldsymbol{z}_l|\boldsymbol{z}_{<l}) - H(\boldsymbol{z}_l|\boldsymbol{z}_{<l}, \boldsymbol{x})
    \end{aligned}
    \label{eq:16}
\end{equation}
where $H$ is the Shannon entropy. Then, the summation has a lower bound:
\begin{equation}
    \begin{aligned}
    &\sum\limits_{i=1}^{L}\mathbb{E}_{p(\boldsymbol{x})}\mathbb{E}_{q(\boldsymbol{z}_{<l}|\boldsymbol{x})}[\mathrm{KL}(q(\boldsymbol{z}|\boldsymbol{x}, \boldsymbol{z}_{<l})||p(\boldsymbol{z}_l|\boldsymbol{z}_{<l}))] \\
    \ge& \sum\limits_{i=1}^{L} H(\boldsymbol{z}_l|\boldsymbol{z}_{<l}) - H(\boldsymbol{z}_l|\boldsymbol{z}_{<l}, \boldsymbol{x}) \\
    =&H(\boldsymbol{z}_1, \dots, \boldsymbol{z}_L) - H(\boldsymbol{z}_1, \dots, \boldsymbol{z}_L|\boldsymbol{x}) \\
    =&I(\boldsymbol{x};\boldsymbol{z}_1, \dots, \boldsymbol{z}_L) 
    \end{aligned}
    \label{eq:17}
\end{equation}
where $I$ is mutual information. Next, we prove the following inequality with induction:
\begin{equation}
    I(\boldsymbol{x};\boldsymbol{z}_1, \dots, \boldsymbol{z}_L) \ge I(\boldsymbol{x};\boldsymbol{z}_1; \dots; \boldsymbol{z}_L)
\end{equation}
When $L = 2$, we proof $I(\boldsymbol{x};\boldsymbol{z}_1, \boldsymbol{z}_2) \ge I(\boldsymbol{x};\boldsymbol{z}_1; \boldsymbol{z}_2)$. Actually, we have the following facts:
\begin{equation}
\begin{aligned}
    &I(\boldsymbol{x};\boldsymbol{z}_1, \boldsymbol{z}_2) \\
    =&H(\boldsymbol{x}) + H(\boldsymbol{z}_1, \boldsymbol{z}_2) - H(\boldsymbol{x}, \boldsymbol{z}_1, \boldsymbol{z}_2) 
\end{aligned}
\end{equation}
\begin{equation}
    \begin{aligned}
    &I(\boldsymbol{x};\boldsymbol{z}_1;\boldsymbol{z}_2) \\
    =&H(\boldsymbol{x}) + H(\boldsymbol{z}_1) + H(\boldsymbol{z}_2) + H(\boldsymbol{x}, \boldsymbol{z}_1, \boldsymbol{z}_2) \\
    &-  H(\boldsymbol{z}_1, \boldsymbol{z}_2) - H(\boldsymbol{x}, \boldsymbol{z}_1) - H(\boldsymbol{x}, \boldsymbol{z}_2)
    \end{aligned}
\end{equation}
Based on the facts above, we have:
\begin{align}
    I(\boldsymbol{x};\boldsymbol{z}_1, \boldsymbol{z}_2) &\ge I(\boldsymbol{x};\boldsymbol{z}_1; \boldsymbol{z}_2) \\
    \Leftrightarrow 2H(\boldsymbol{z}_1, \boldsymbol{z}_2) + H(\boldsymbol{x}, \boldsymbol{z}_1) + H(\boldsymbol{x}, \boldsymbol{z}_2) &\ge H(\boldsymbol{z}_1) + H(\boldsymbol{z}_2) + 2H(\boldsymbol{x}, \boldsymbol{z}_1, \boldsymbol{z}_2)
\end{align}
It's true because we have:
\begin{equation}
    \begin{aligned}
    &H(\boldsymbol{z}_1, \boldsymbol{z}_2) + H(\boldsymbol{x}, \boldsymbol{z}_1) \\
    =& H(\boldsymbol{z}_2|\boldsymbol{z}_1) + H(\boldsymbol{x}|\boldsymbol{z}_1) + 2H(\boldsymbol{z}_1) \\
    \ge &H(\boldsymbol{x}, \boldsymbol{z}_2|\boldsymbol{z}_1) + 2H(\boldsymbol{z}_1) \\
    =& H(\boldsymbol{x}, \boldsymbol{z}_1, \boldsymbol{z}_2) + H(\boldsymbol{z}_1) 
    \end{aligned}
    \label{eq:23}
\end{equation}
Similarly, the following inequality also holds true:
\begin{equation}
    H(\boldsymbol{z}_1, \boldsymbol{z}_2) + H(\boldsymbol{x}, \boldsymbol{z}_2) \ge H(\boldsymbol{x}, \boldsymbol{z}_1, \boldsymbol{z}_2) + H(\boldsymbol{z}_2) 
    \label{eq:24}
\end{equation}
Therefore, making sum to Eq.(\ref{eq:23}) and Eq.(\ref{eq:24}), we conclude that $I(\boldsymbol{x};\boldsymbol{z}_1, \boldsymbol{z}_2) \ge I(\boldsymbol{x};\boldsymbol{z}_1; \boldsymbol{z}_2)$. Hence, we finish the proof of the $L=2$ case. 

% Because $I(\boldsymbol{x};\boldsymbol{z}_1, \boldsymbol{z}_2) = I(\boldsymbol{x};\boldsymbol{z}_1) + I(\boldsymbol{x};\boldsymbol{z}_2|\boldsymbol{z}_1)$, $I(\boldsymbol{x};\boldsymbol{z}_1;\boldsymbol{z}_2) = I(\boldsymbol{x};\boldsymbol{z}_1) - I(\boldsymbol{x};\boldsymbol{z}_1|\boldsymbol{z}_2)$, we have
% \begin{equation}
%     I(\boldsymbol{x};\boldsymbol{z}_2|\boldsymbol{z}_1) \ge - I(\boldsymbol{x};\boldsymbol{z}_1|\boldsymbol{z}_2)
% \end{equation}

When $L = k$, suppose $I(\boldsymbol{x};\boldsymbol{z}_1, \dots, \boldsymbol{z}_k) \ge I(\boldsymbol{x};\boldsymbol{z}_1; \dots; \boldsymbol{z}_k)$, we consider $L = k+1$. In this case, based on the inductive assumption, we have:
\begin{equation}
    I(\boldsymbol{x};\boldsymbol{z}_1, \dots, \boldsymbol{z}_{k+1}) \ge I(\boldsymbol{x};\boldsymbol{z}_1, \dots, \boldsymbol{z}_{k}) \ge  I(\boldsymbol{x};\boldsymbol{z}_1; \dots; \boldsymbol{z}_{k}) \ge I(\boldsymbol{x};\boldsymbol{z}_1; \dots; \boldsymbol{z}_{k+1})
    \label{eq:25}
\end{equation}
Hence, the case of $L=k+1$ also holds true. Therefore, we conclude that $I(\boldsymbol{x};\boldsymbol{z}_1, \dots, \boldsymbol{z}_L) \ge I(\boldsymbol{x};\boldsymbol{z}_1; \dots; \boldsymbol{z}_L)$.

Now, we consider the interaction information and can obtain:
\begin{equation}
    \begin{aligned}
    &I(\boldsymbol{x};\boldsymbol{z}_1; \dots; \boldsymbol{z}_L)  \\
    =&I(\boldsymbol{z}_L, \boldsymbol{z}_{L-1}) - \sum\limits_{i=2}^{L-1}I(\boldsymbol{z}_L; \dots ; \boldsymbol{z}_i|\boldsymbol{z}_{i-1}) - I(\boldsymbol{z}_L; \dots ; \boldsymbol{z}_1|\boldsymbol{x}) \\
    \ge& \sum\limits_{i=2}^{L-1}I(\boldsymbol{z}_L; \dots ; \boldsymbol{z}_i|\boldsymbol{z}_{i-1}) - I(\boldsymbol{z}_L; \dots ; \boldsymbol{z}_1|\boldsymbol{x})
    \end{aligned}
    \label{eq:26}
\end{equation}

Finally, based on Eq.(\ref{eq:16}), (\ref{eq:17}), (\ref{eq:25}), (\ref{eq:26}), we can conclude:
\begin{equation}
\begin{aligned}
    \mathbb{E}_{p(\boldsymbol{x})}[\mathcal{L}_R] &= \sum\limits_{i=1}^{L}\mathbb{E}_{p(\boldsymbol{x})}\mathbb{E}_{q(\boldsymbol{z}_{<l}|\boldsymbol{x})}[\mathrm{KL}(q(\boldsymbol{z}|\boldsymbol{x}, \boldsymbol{z}_{<l})||p(\boldsymbol{z}_l|\boldsymbol{z}_{<l}))] \\
    &\ge I(\boldsymbol{x};\boldsymbol{z}_1, \dots, \boldsymbol{z}_L) \\
    &\ge I(\boldsymbol{x};\boldsymbol{z}_1; \dots; \boldsymbol{z}_L) \\
    & \ge \sum\limits_{i=2}^{L-1}I(\boldsymbol{z}_L; \dots ; \boldsymbol{z}_i|\boldsymbol{z}_{i-1}) - I(\boldsymbol{z}_L; \dots ; \boldsymbol{z}_1|\boldsymbol{x})
\end{aligned}
\end{equation}

\begin{figure*}[]
\centering
\includegraphics[scale=0.35]{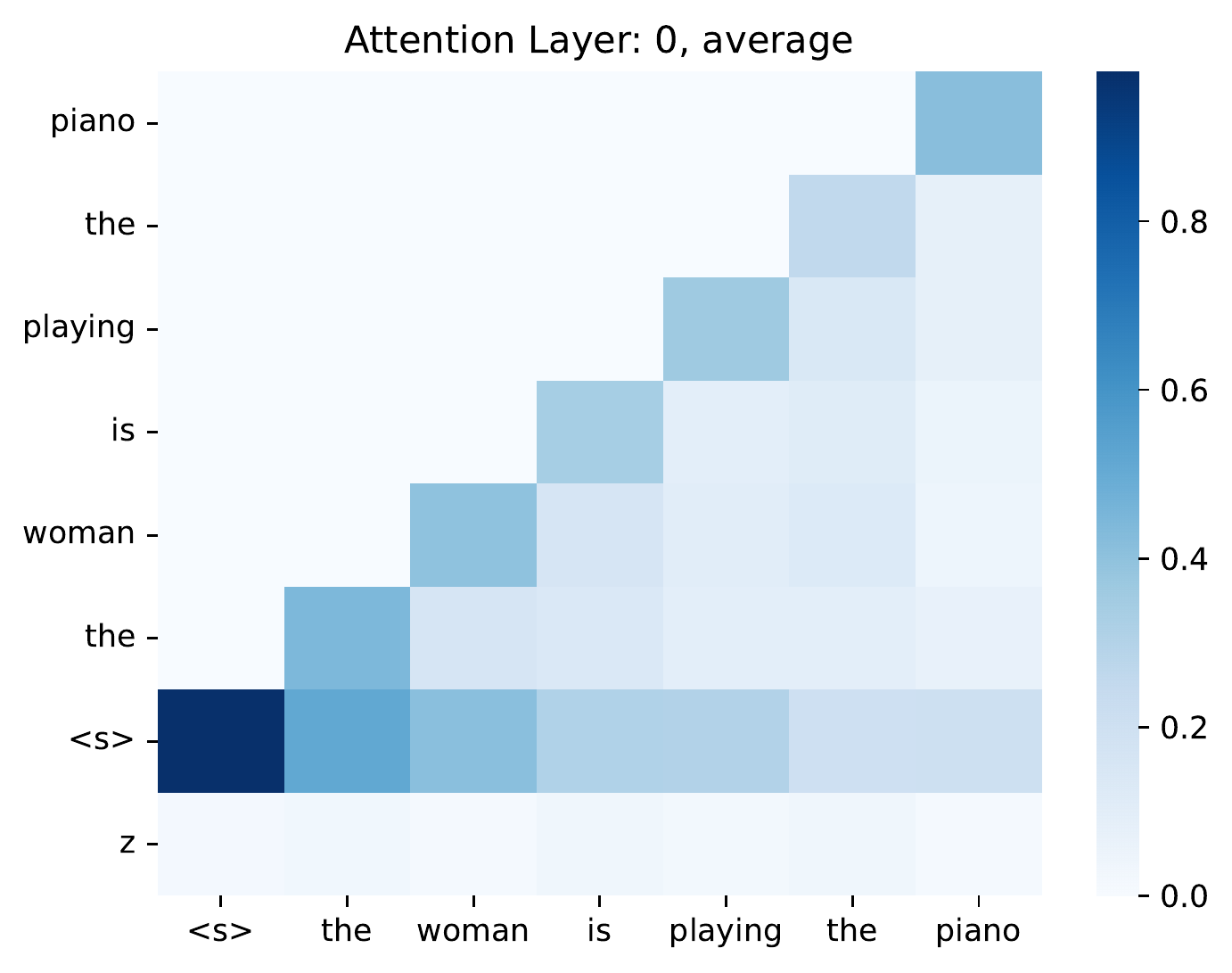} \
\includegraphics[scale=0.35]{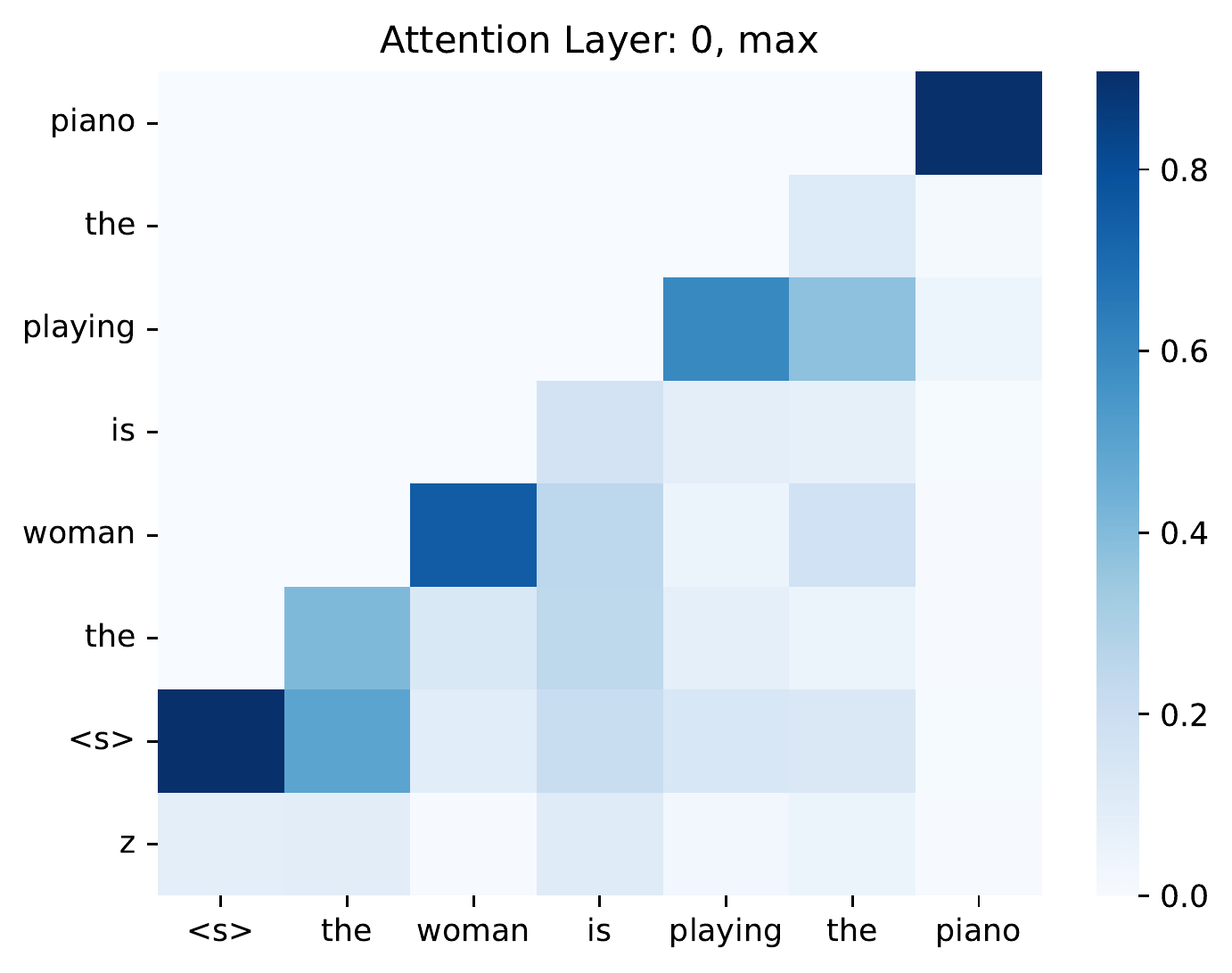} \
\includegraphics[scale=0.35]{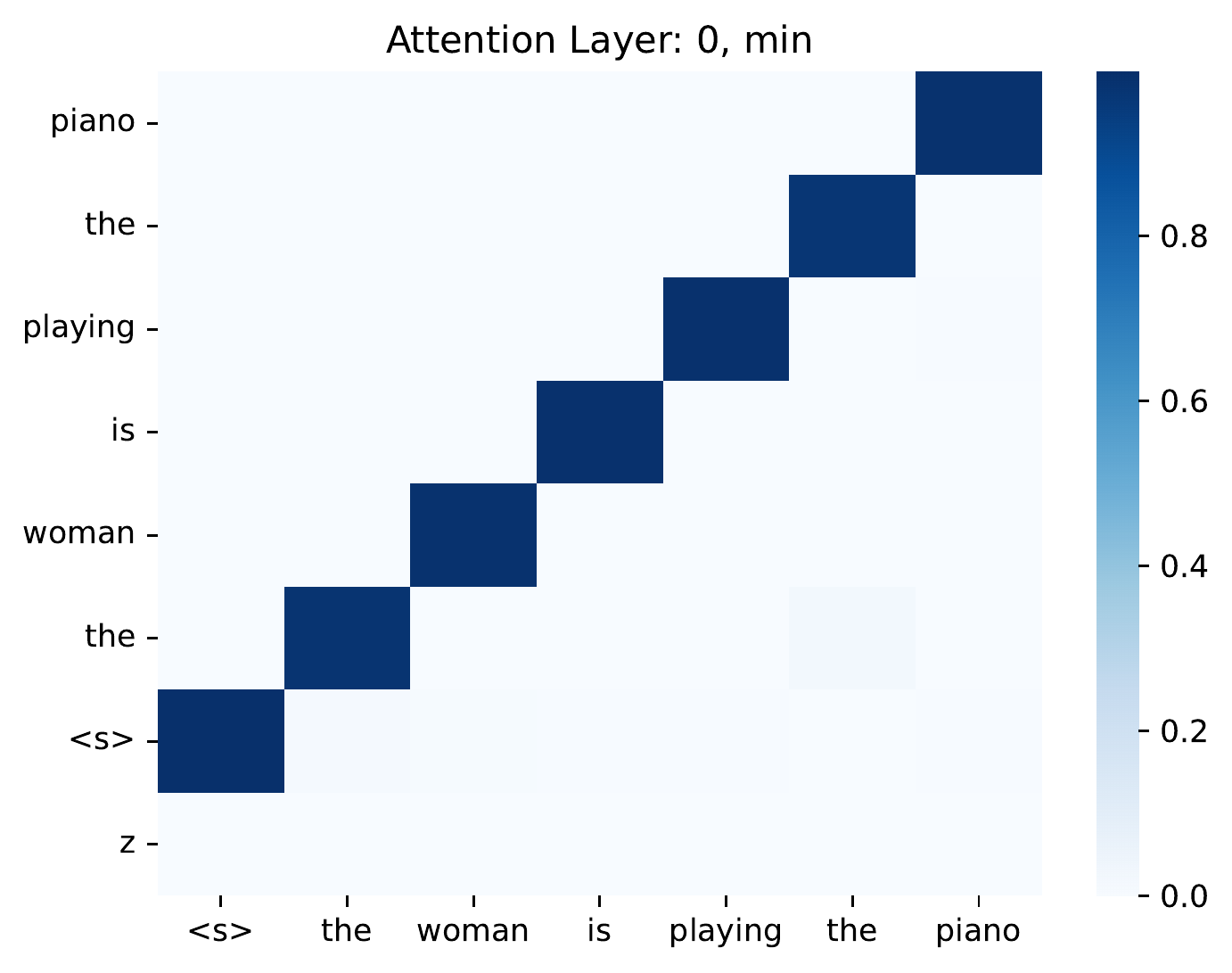} \
\includegraphics[scale=0.35]{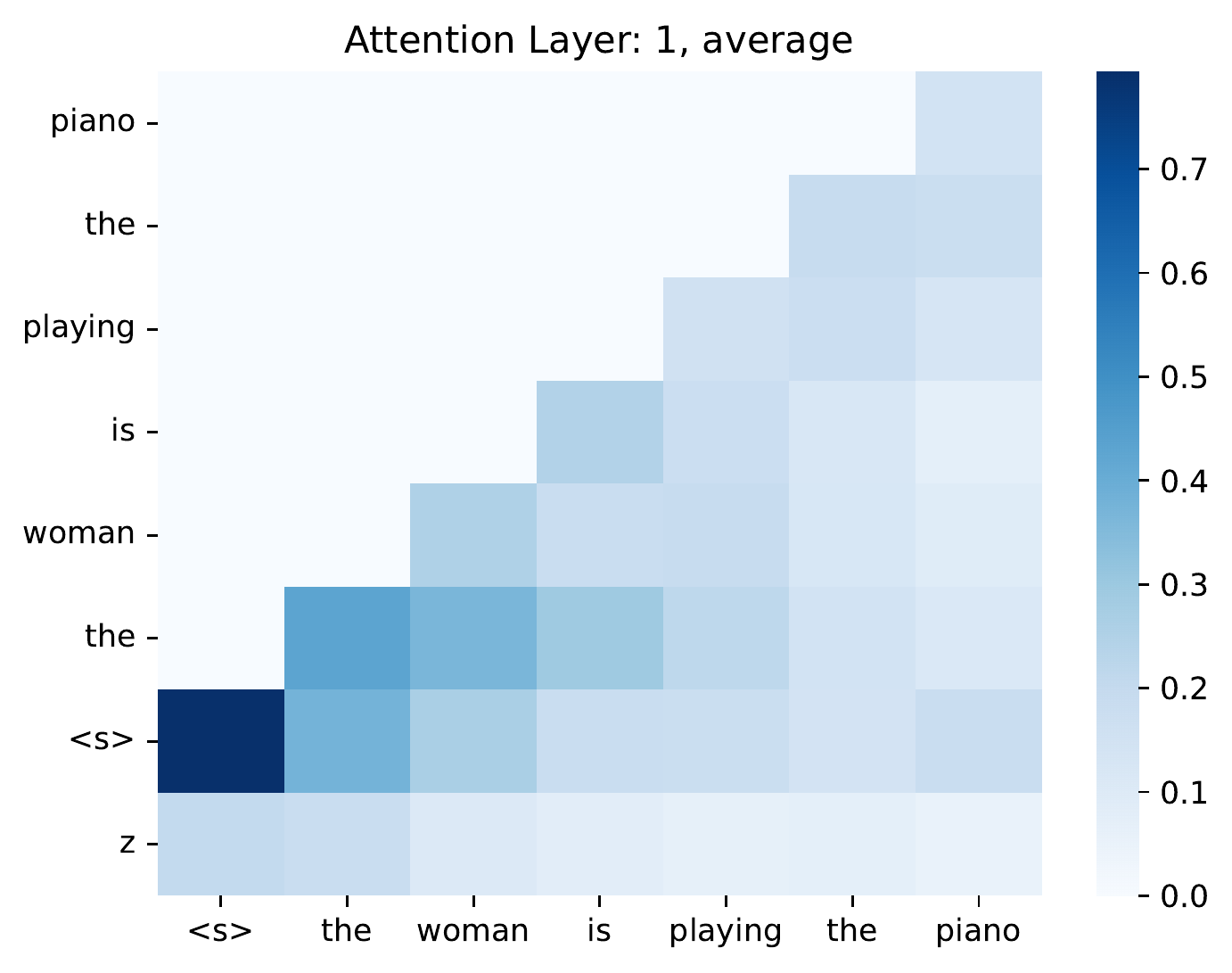} \
\includegraphics[scale=0.35]{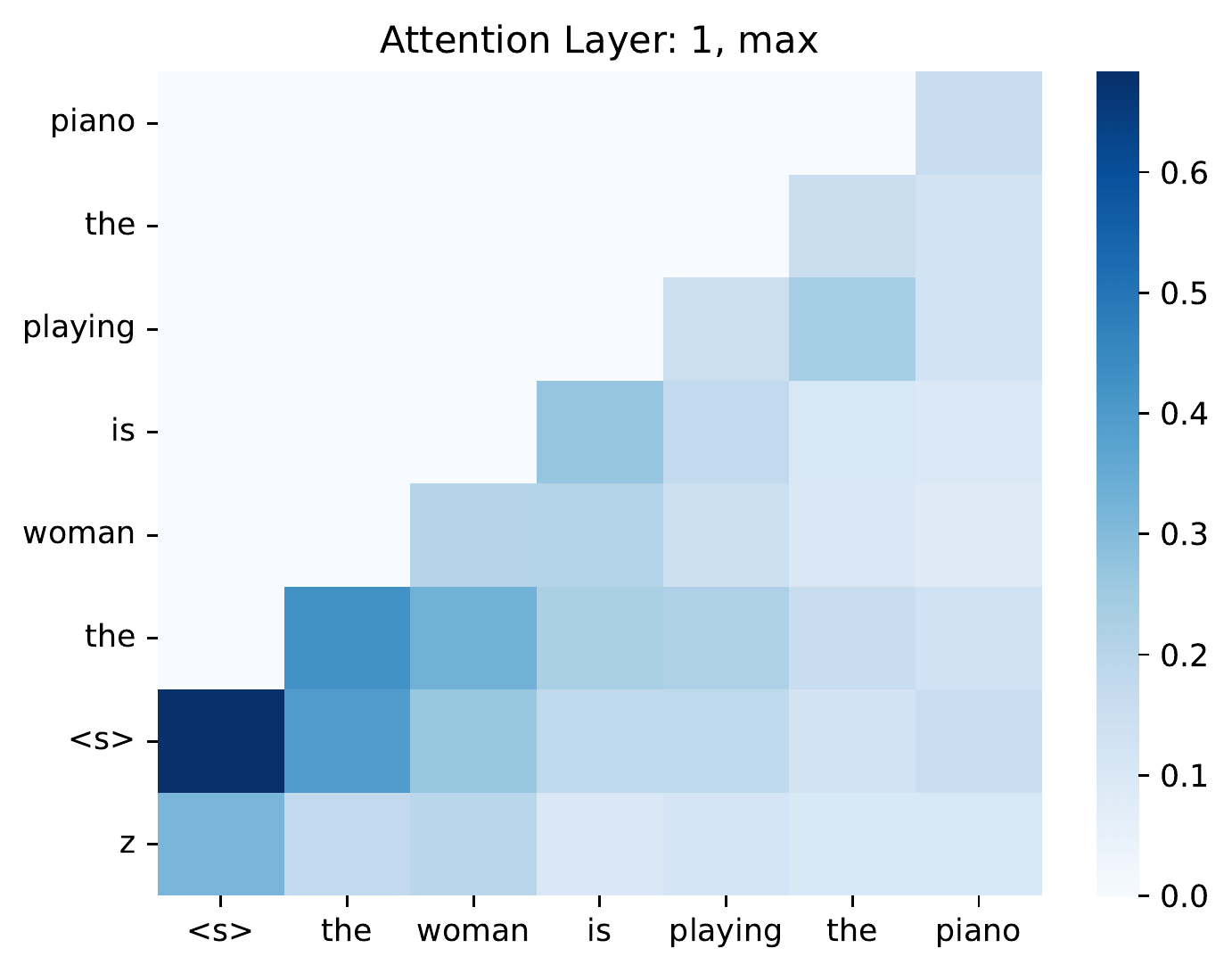} \
\includegraphics[scale=0.35]{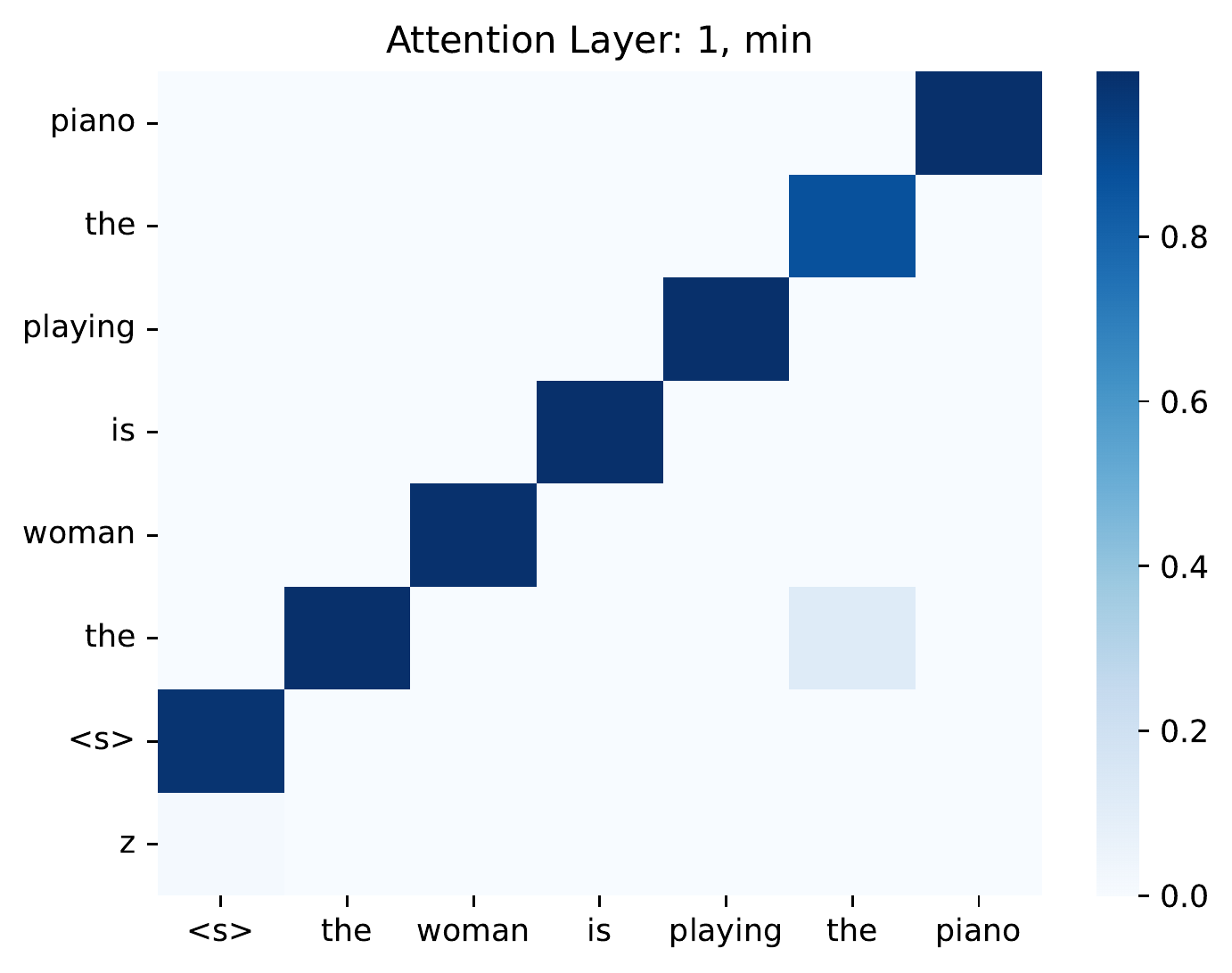} \
\includegraphics[scale=0.35]{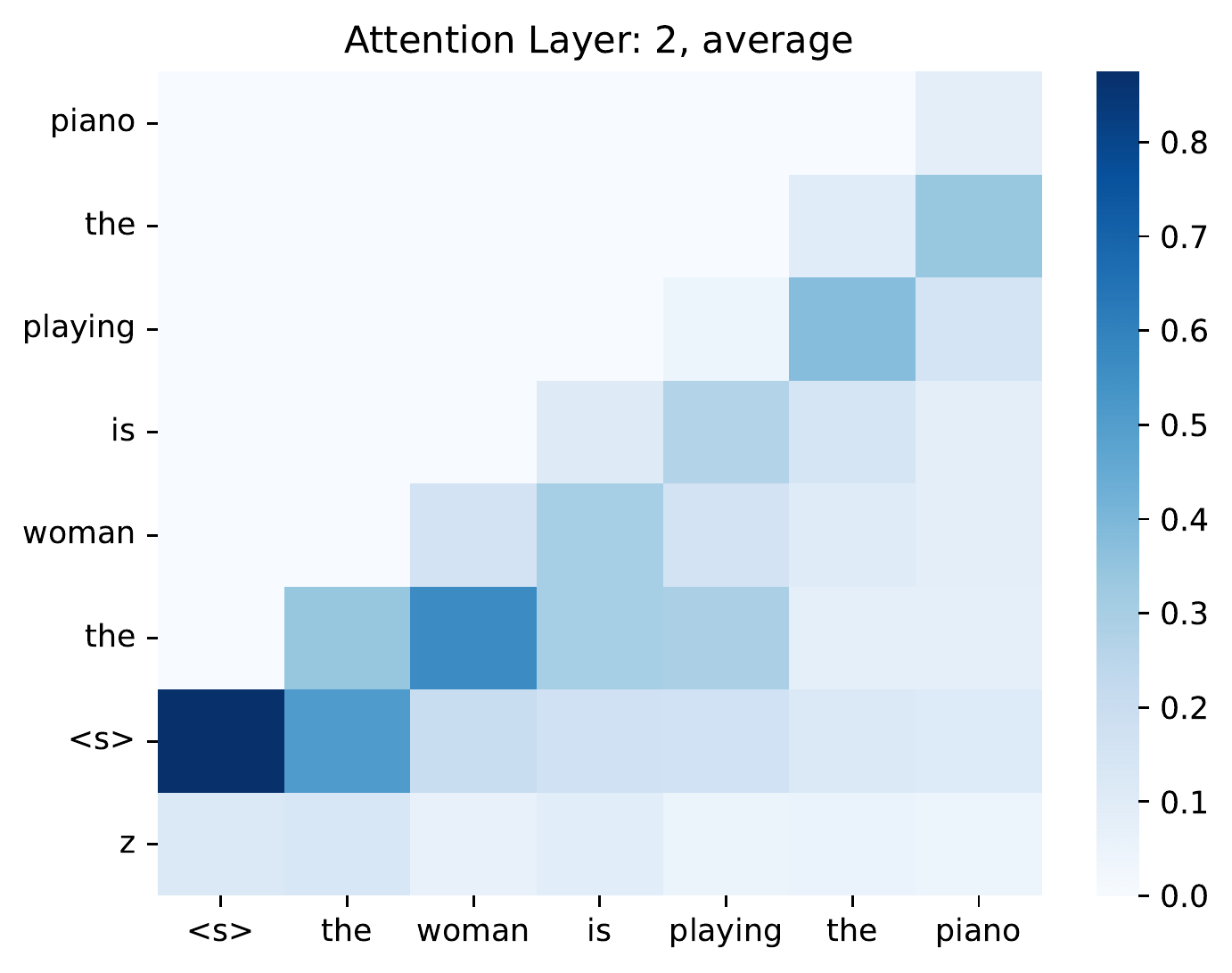} \
\includegraphics[scale=0.35]{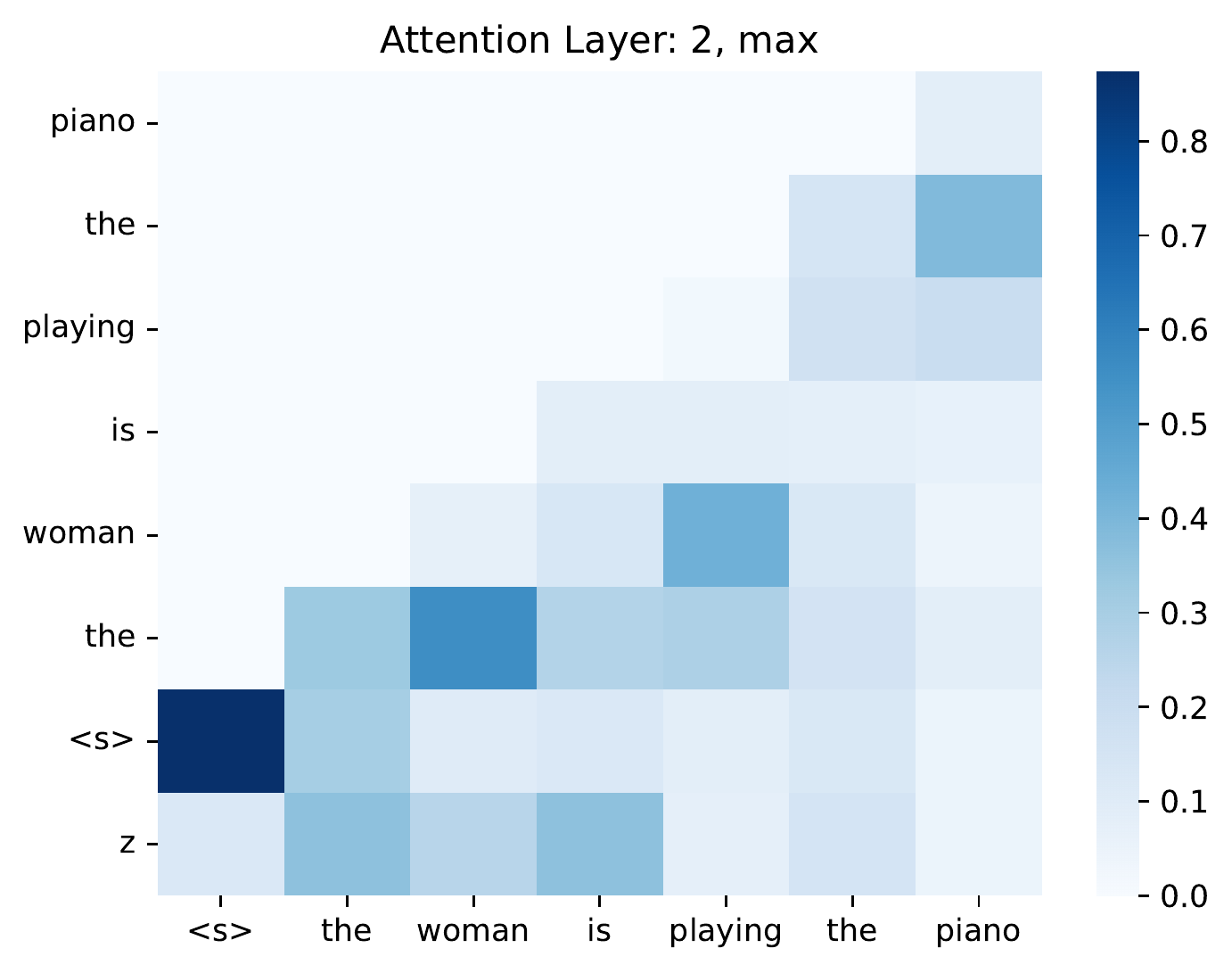} \
\includegraphics[scale=0.35]{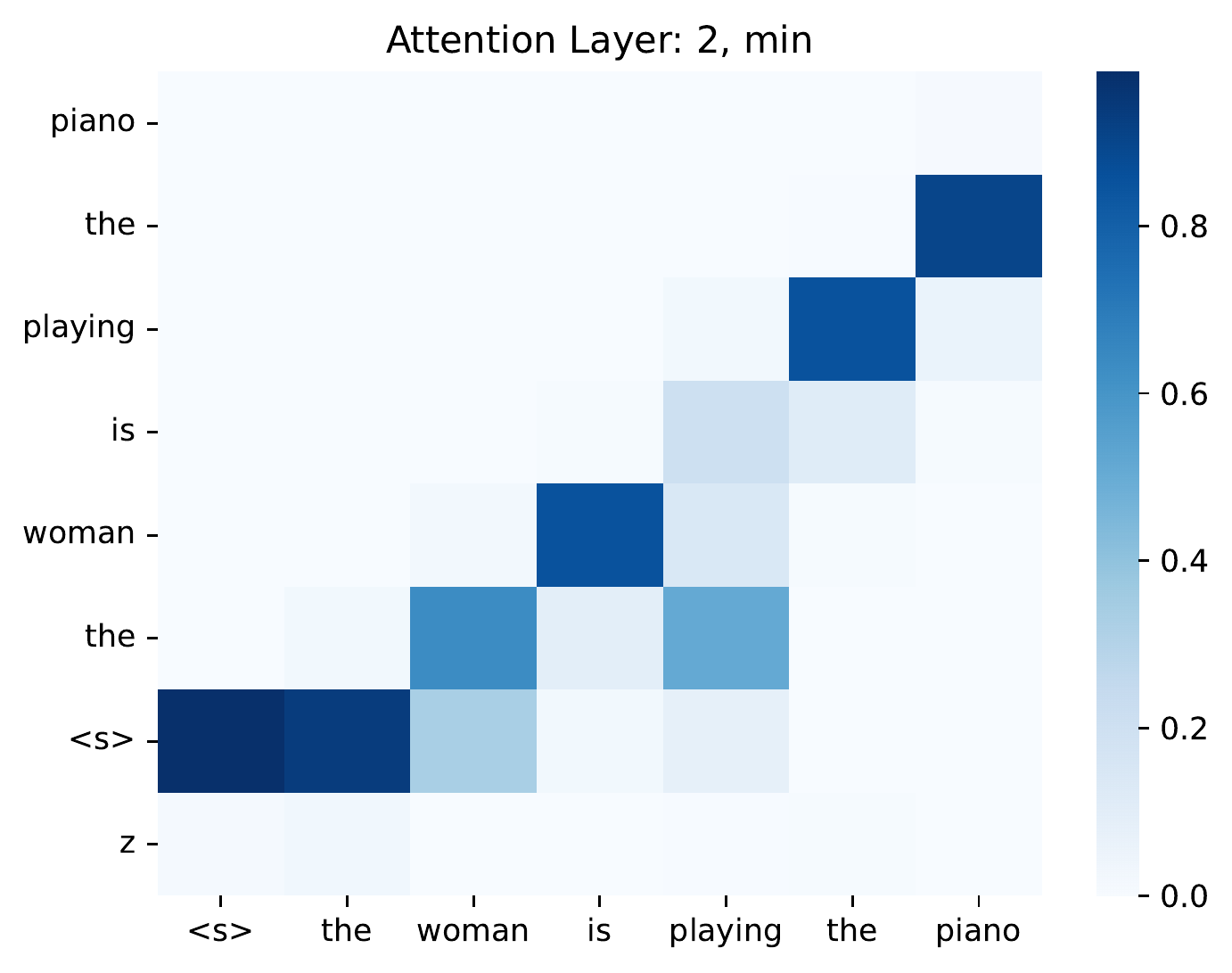} \
\includegraphics[scale=0.35]{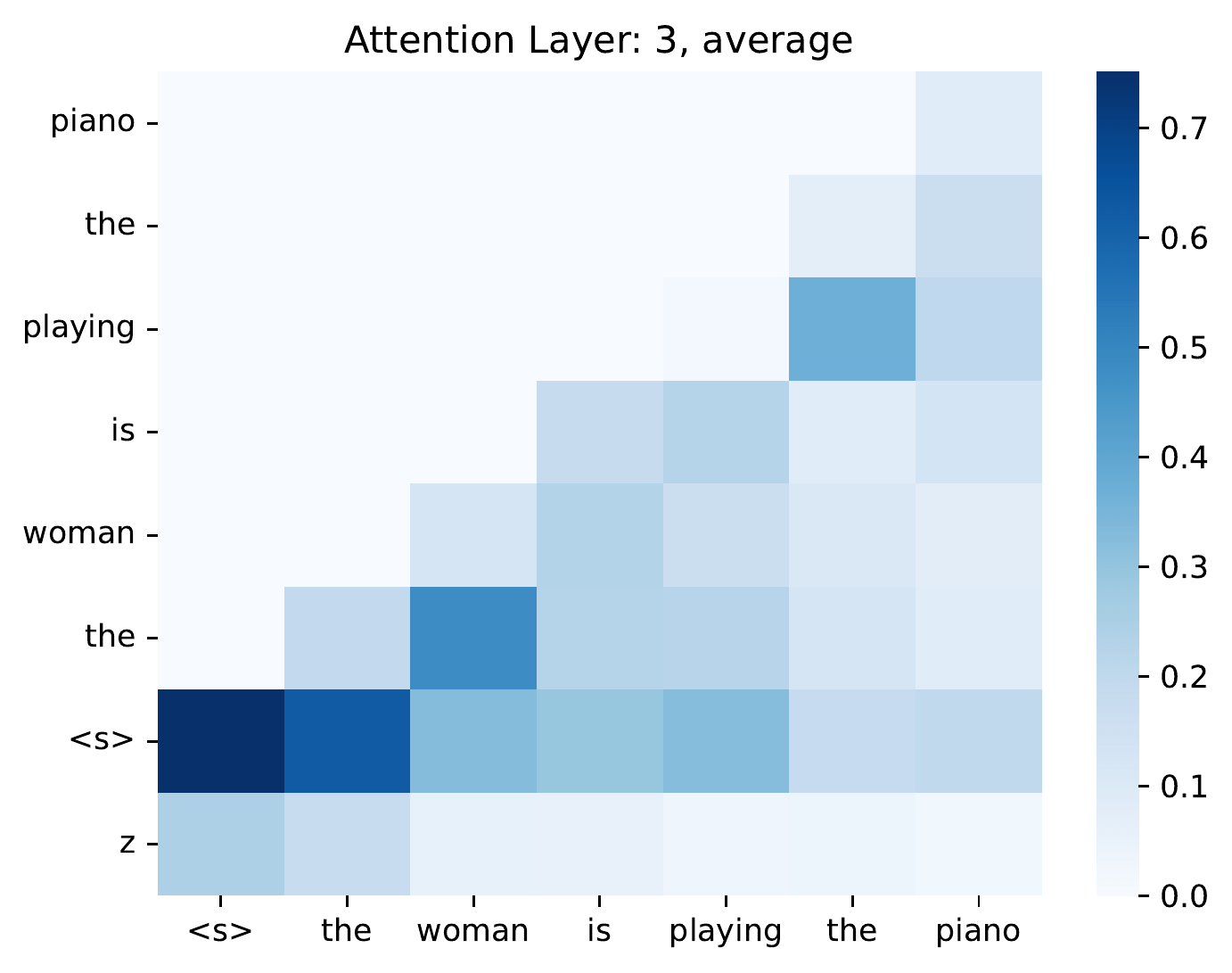} \
\includegraphics[scale=0.35]{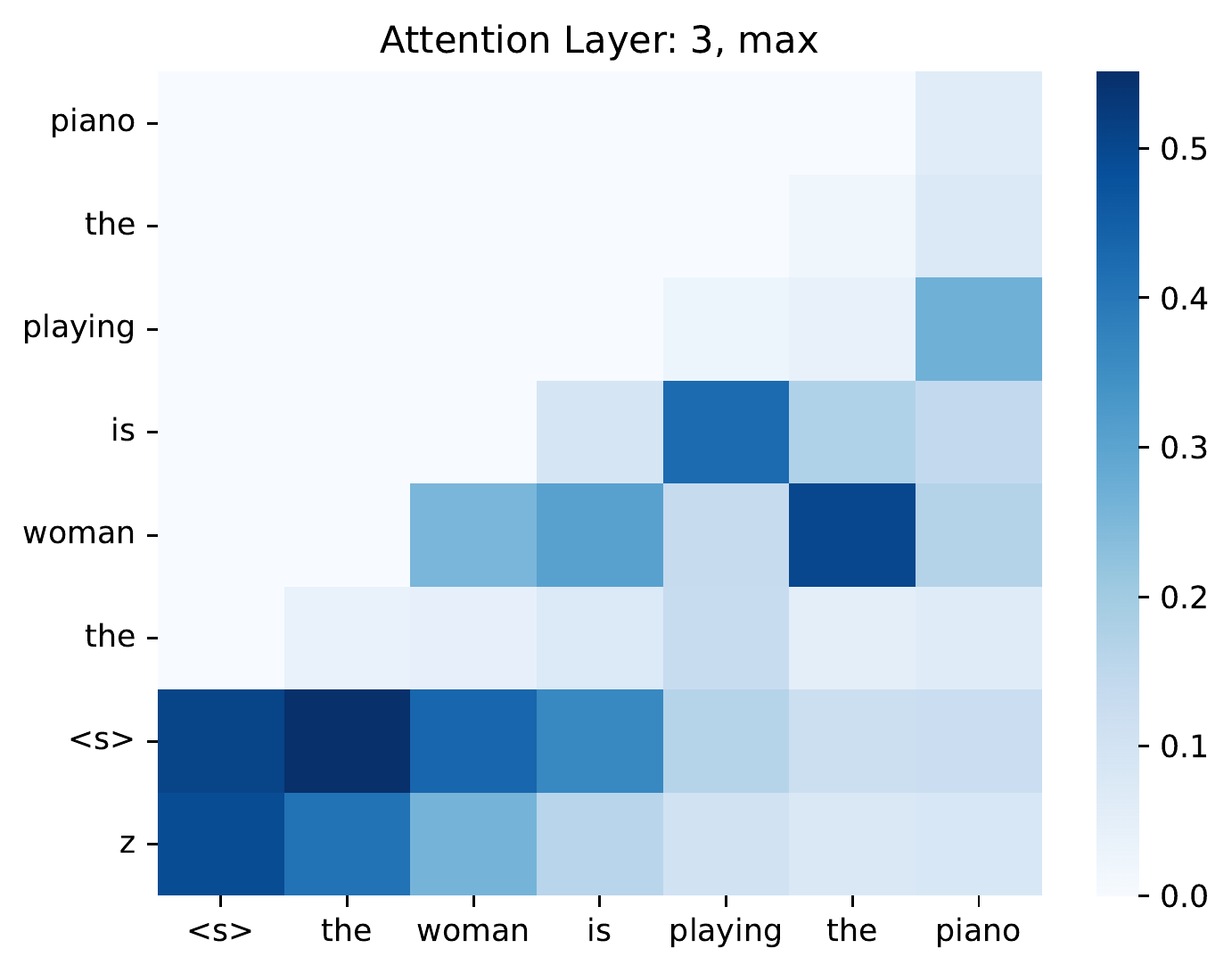} \
\includegraphics[scale=0.35]{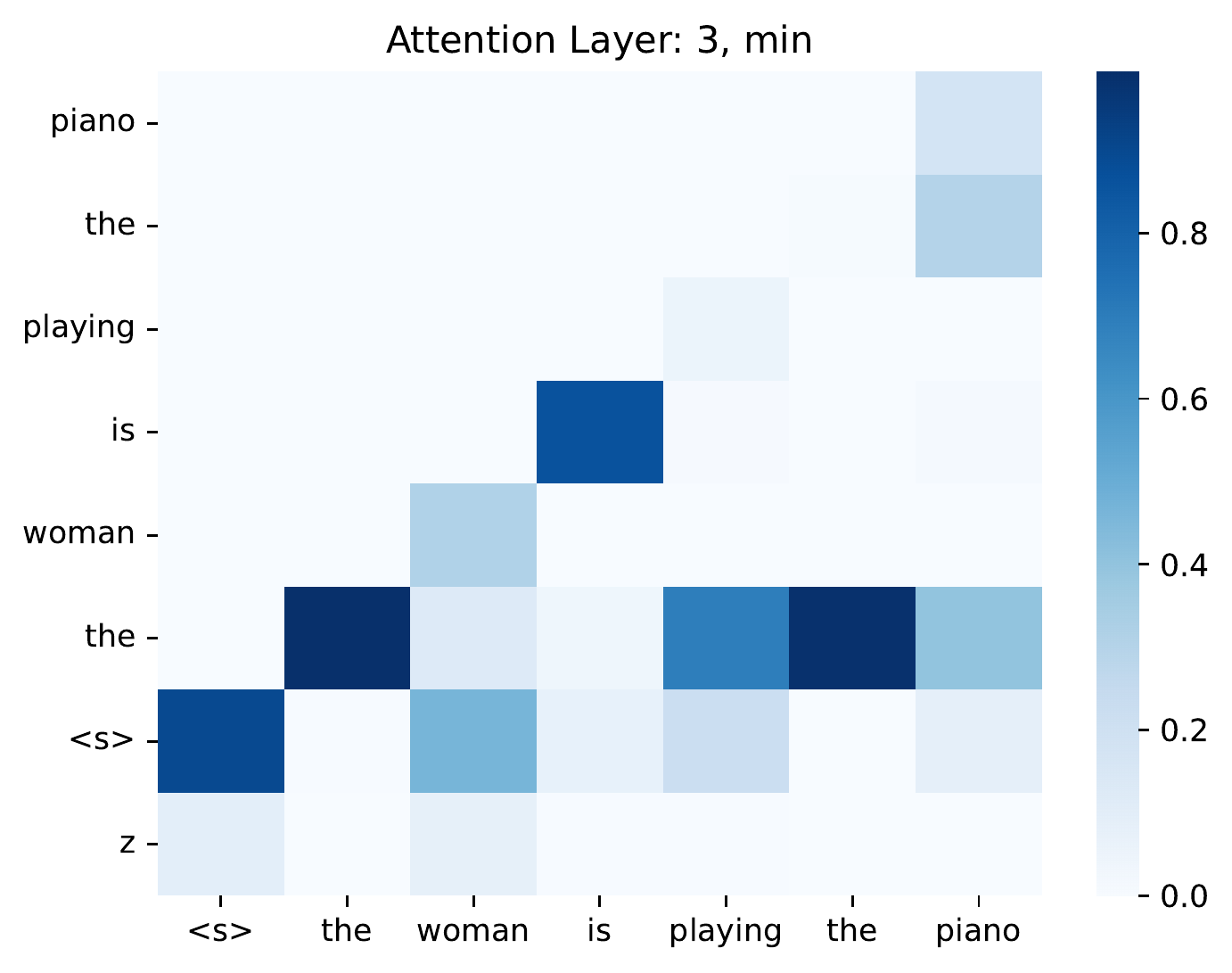} \
\includegraphics[scale=0.35]{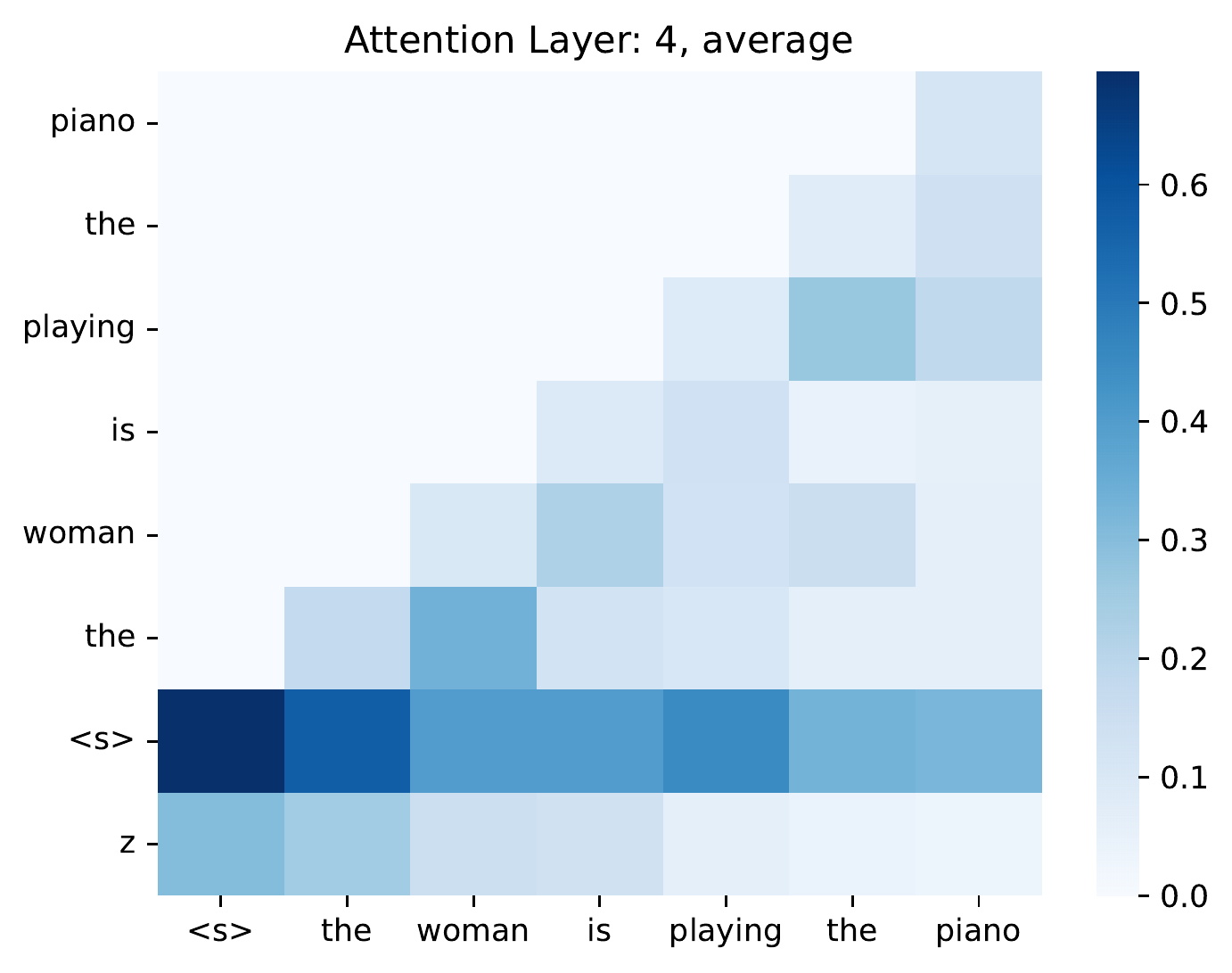} \
\includegraphics[scale=0.35]{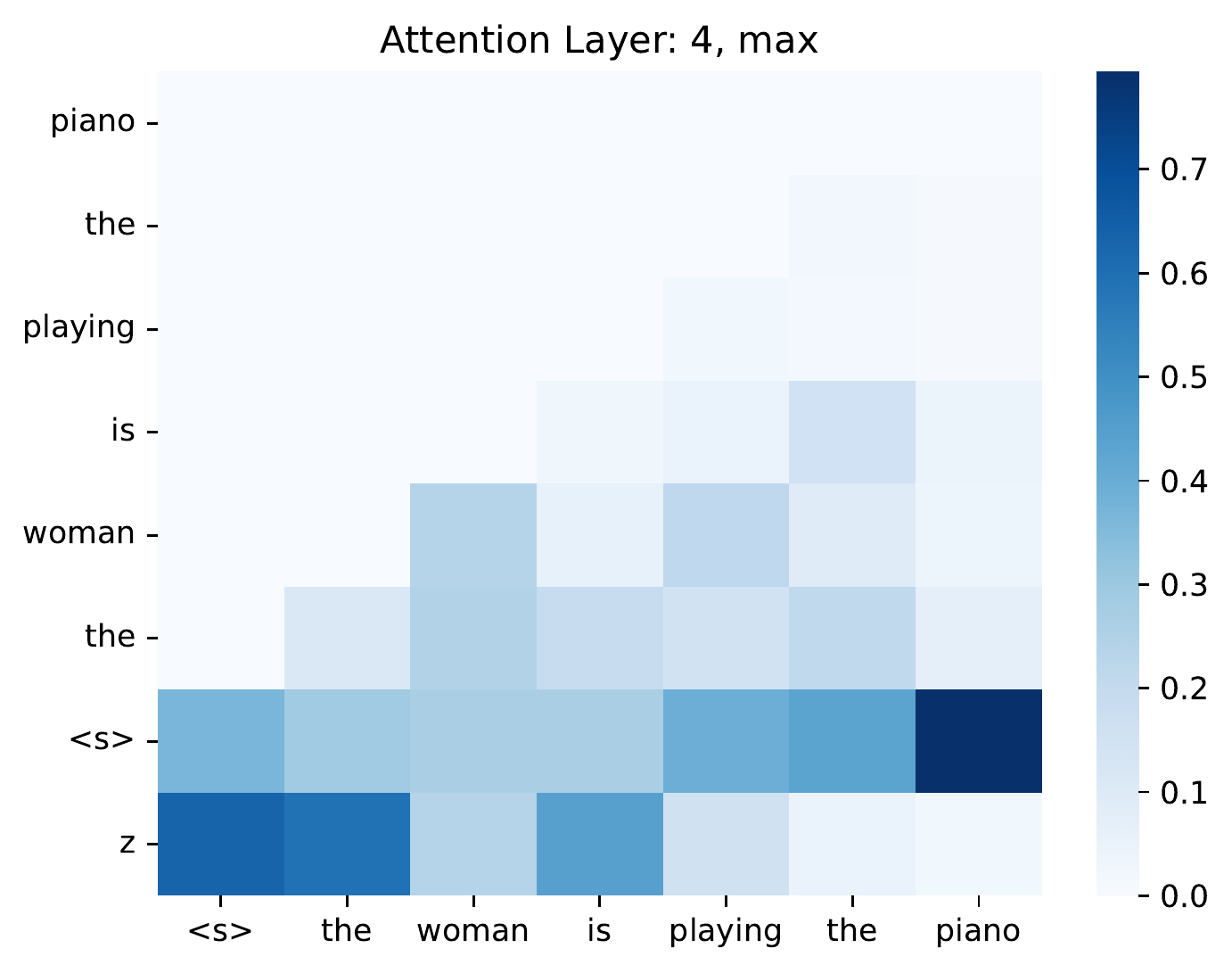} \
\includegraphics[scale=0.35]{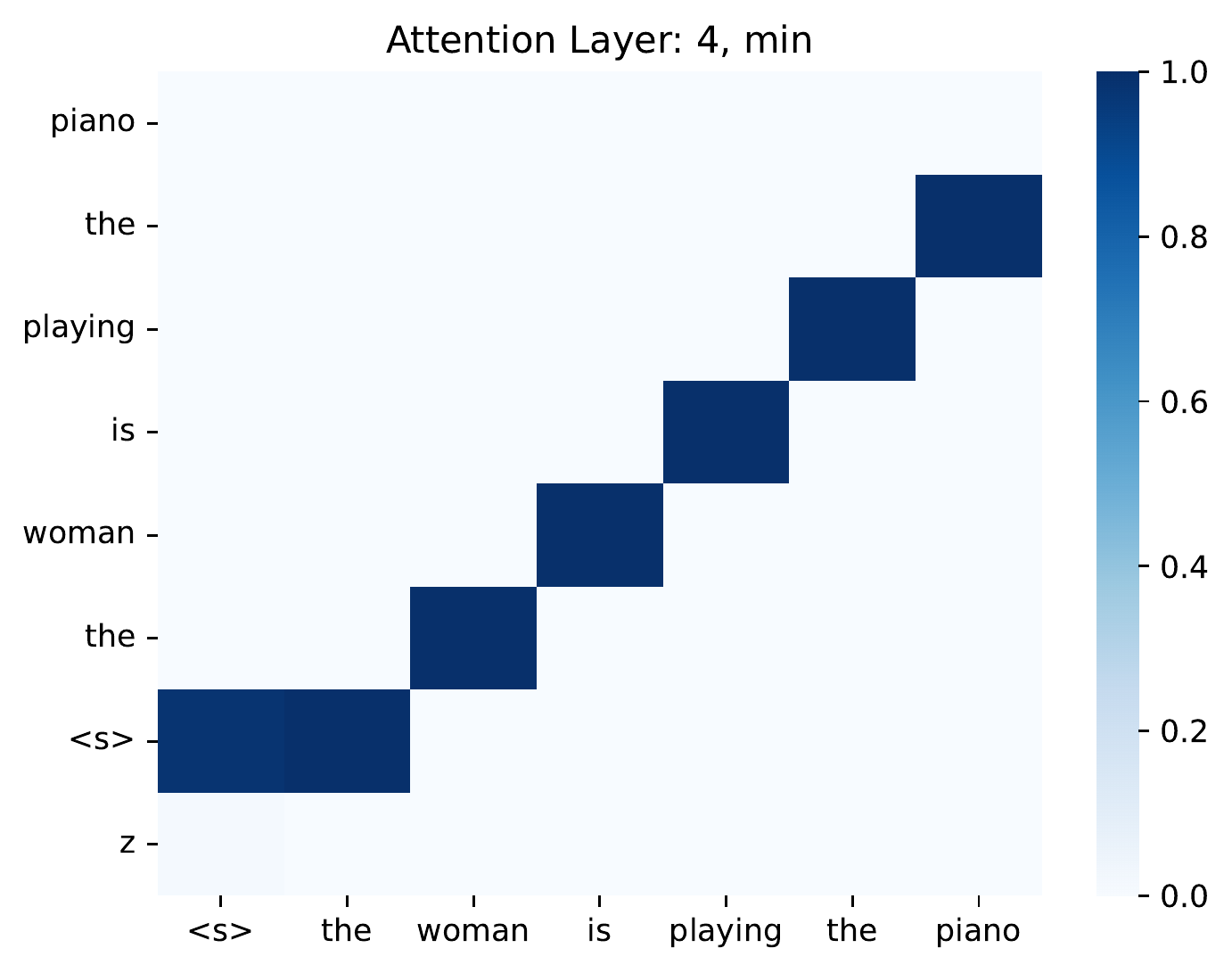} \
\includegraphics[scale=0.35]{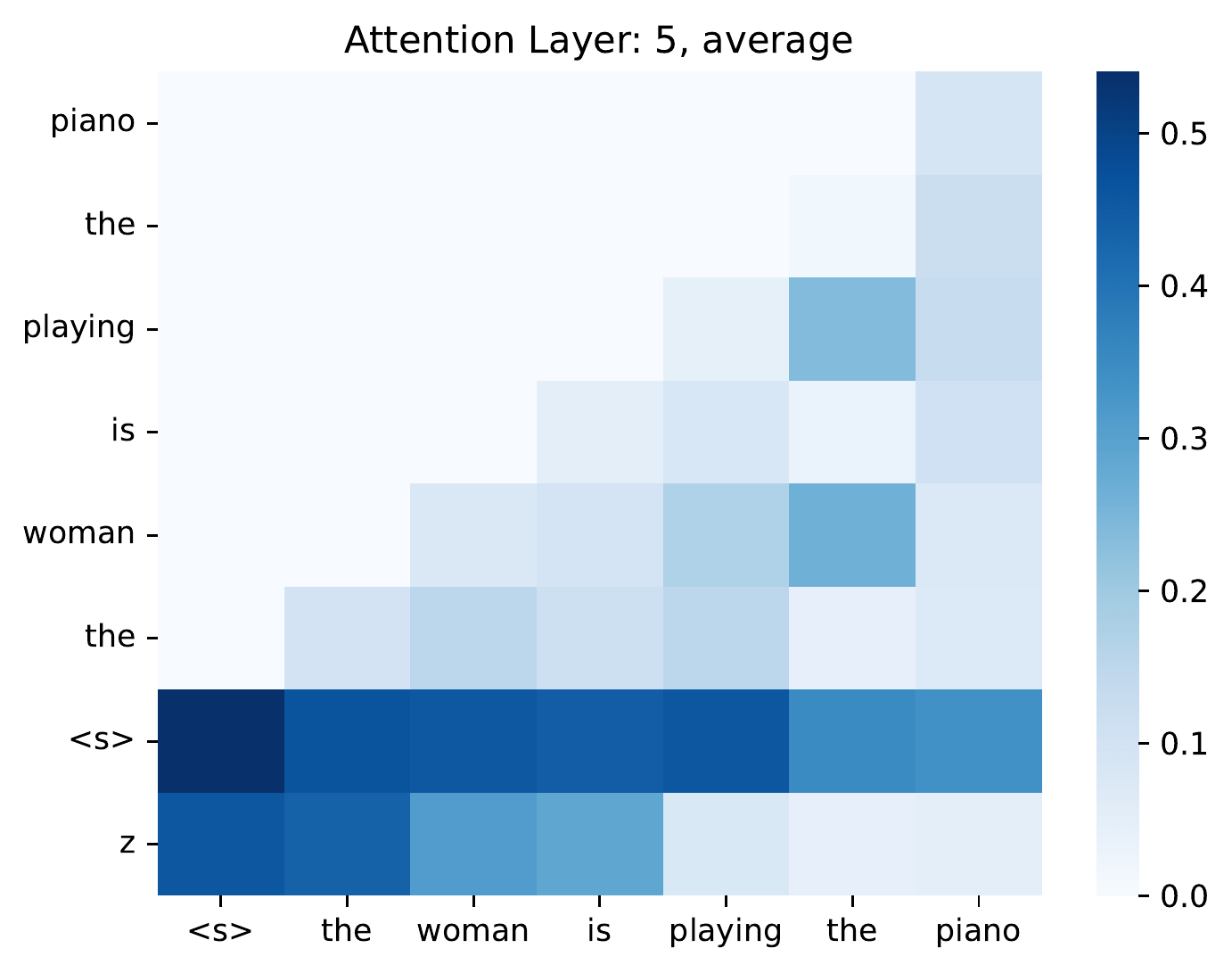} \
\includegraphics[scale=0.35]{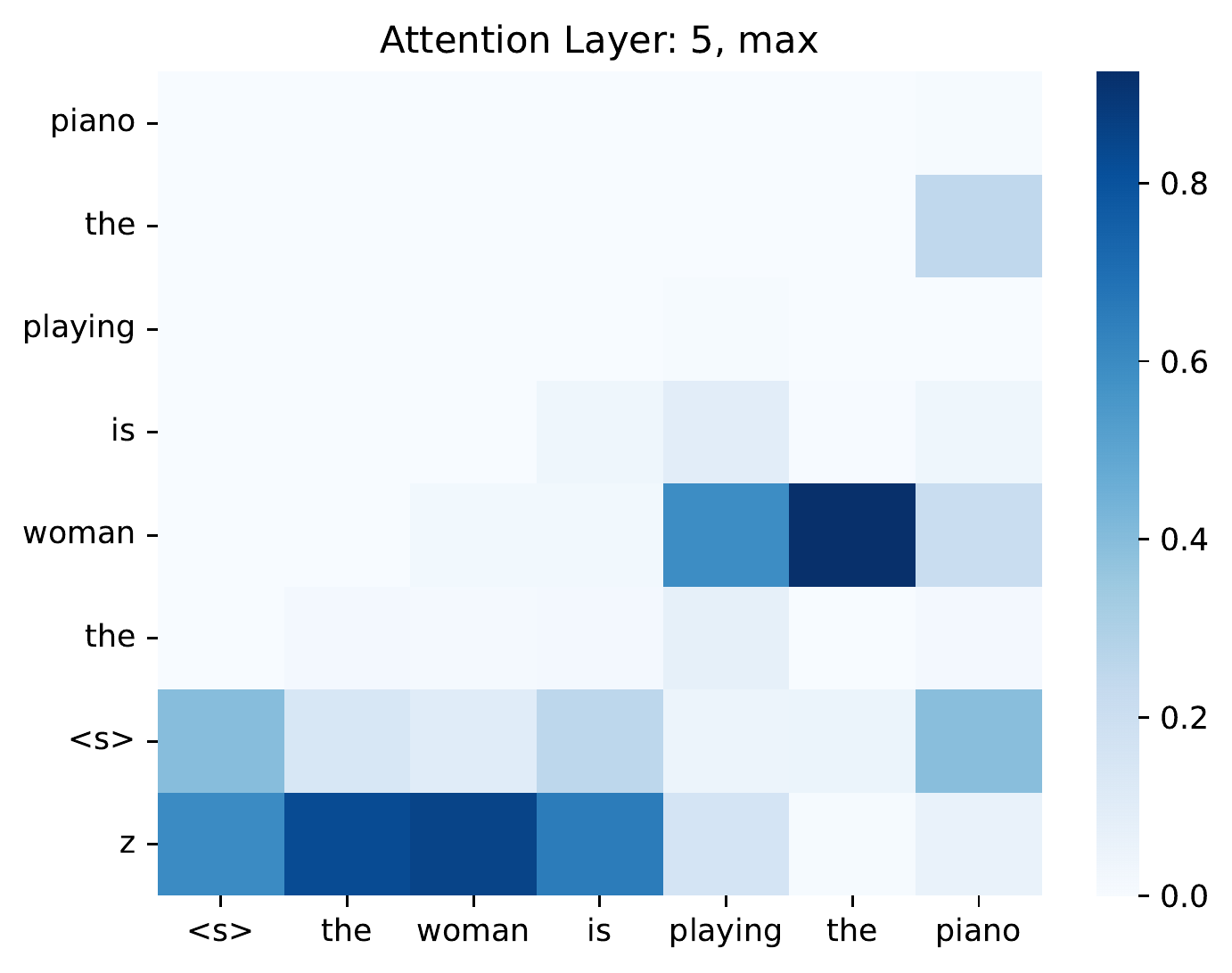} \
\includegraphics[scale=0.35]{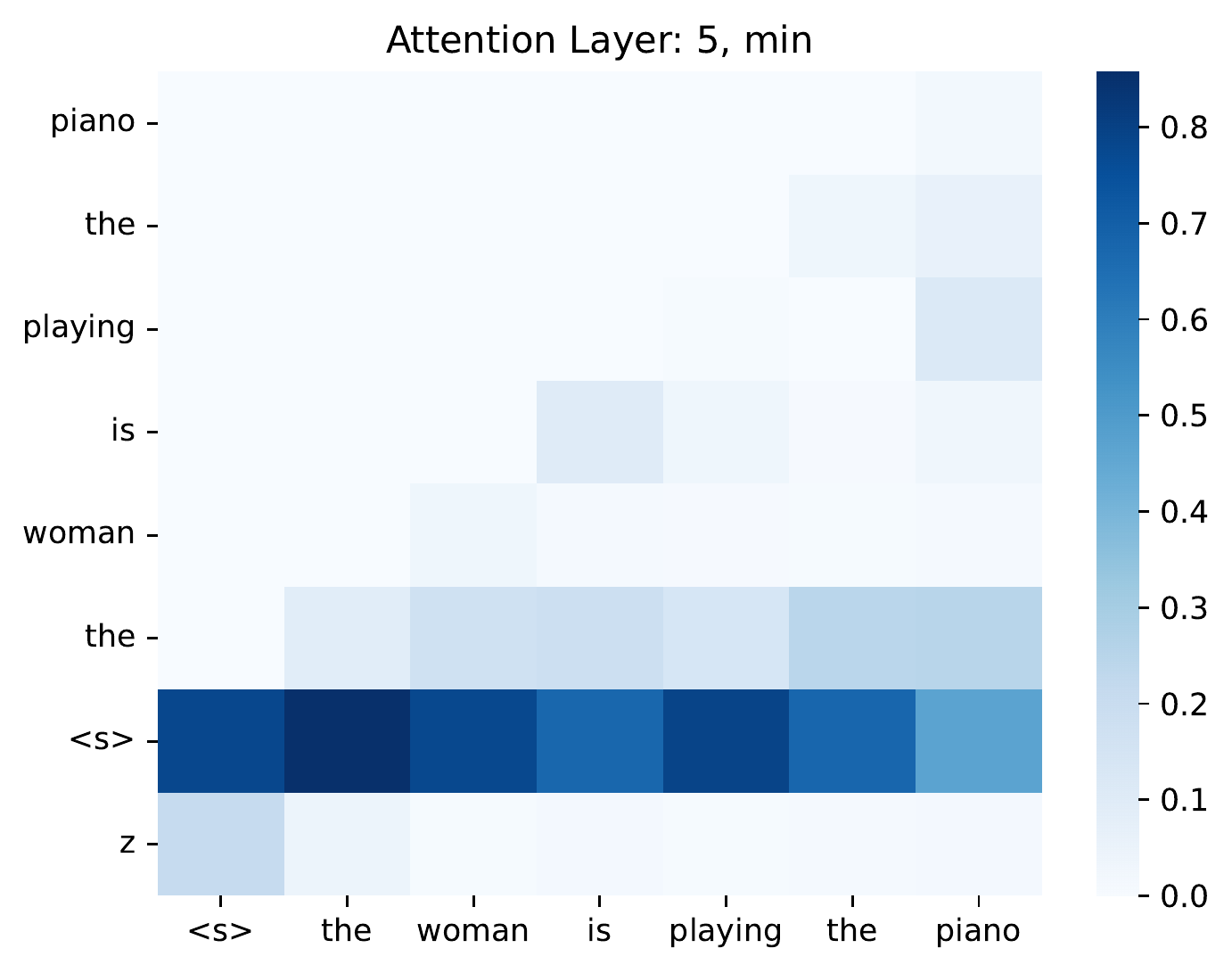} \
\caption{Attention weight of the Memory paradigm for layer 0 to layer 5. We plot three heatmaps in each layer. Average means averaging weights throught all head. Max and min means we select the head with max and min attention weight on the memory token (latent variable).We can see the memory token tends to be ignored by most heads especially in lower layers.}
\label{fig:attn_weight1}
\end{figure*}

\begin{figure*}[]
\centering
\includegraphics[scale=0.35]{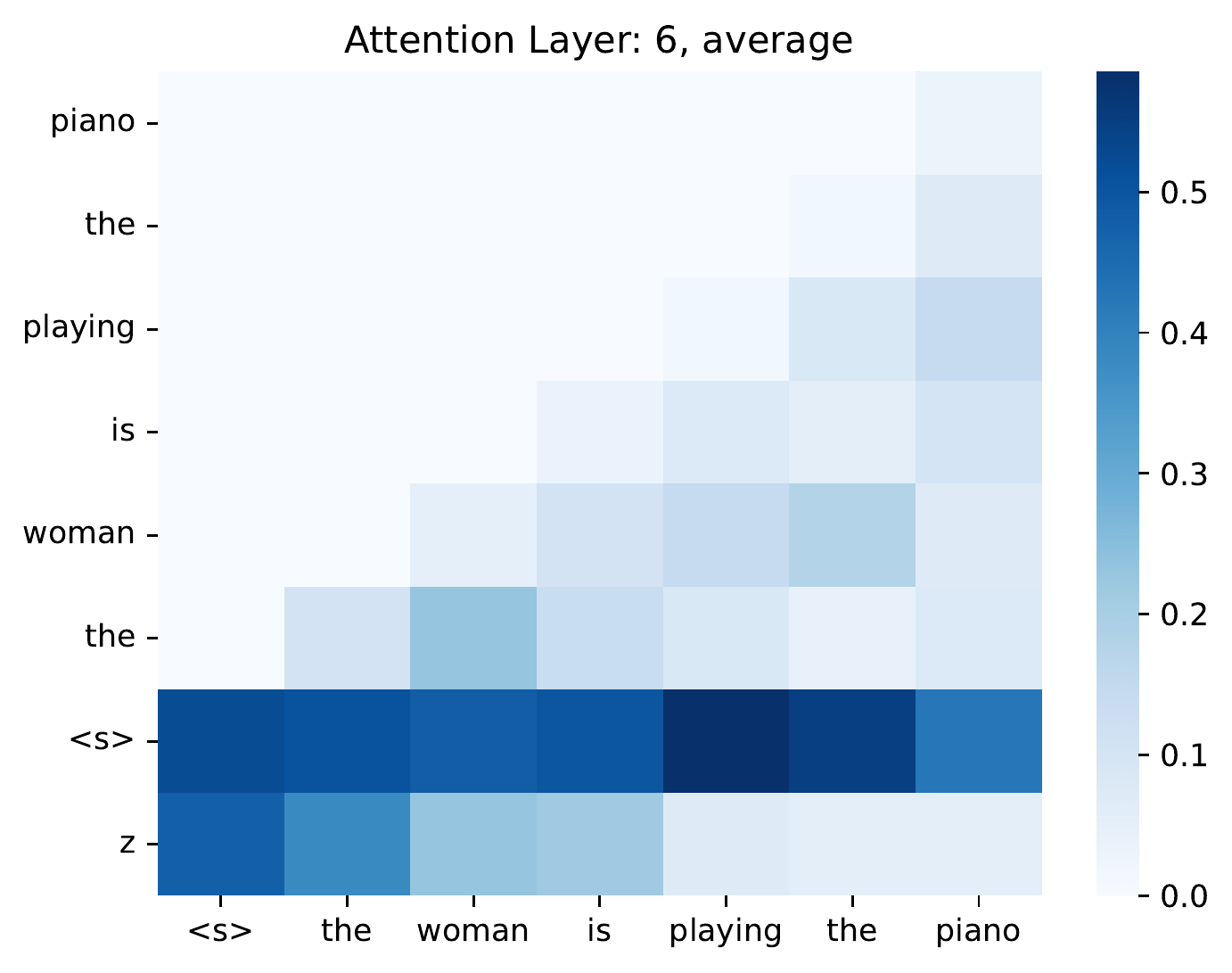} \
\includegraphics[scale=0.35]{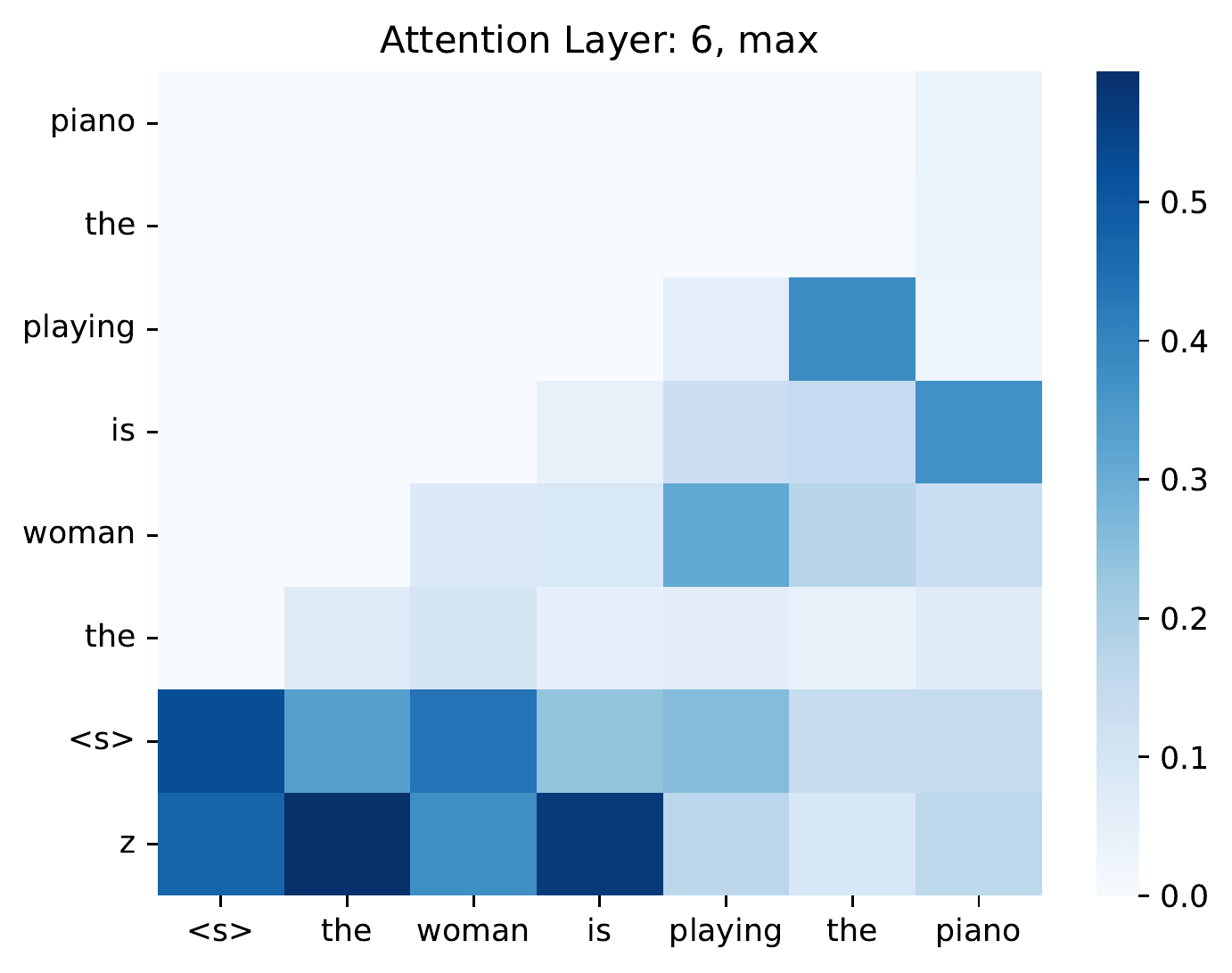} \
\includegraphics[scale=0.35]{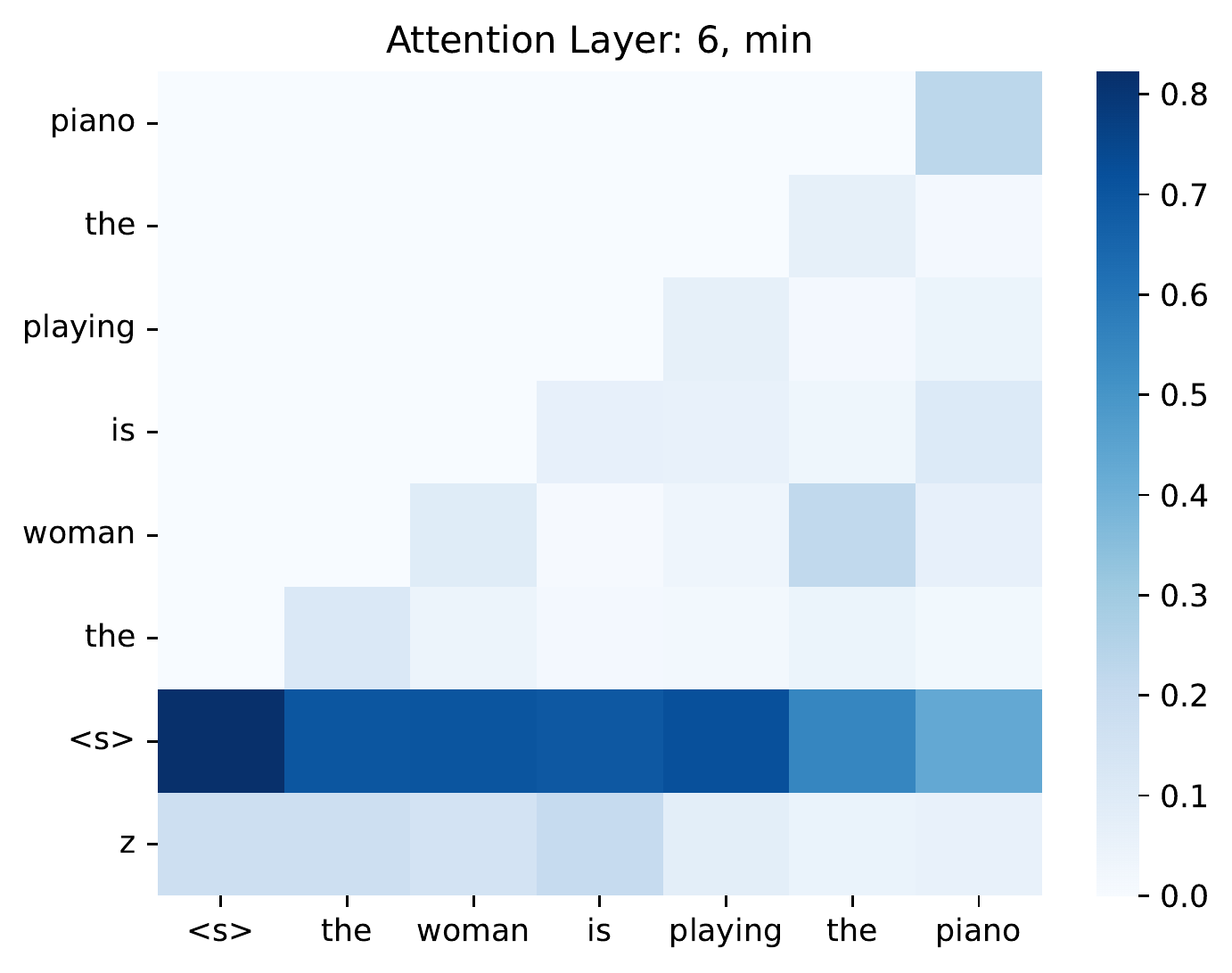} \
\includegraphics[scale=0.35]{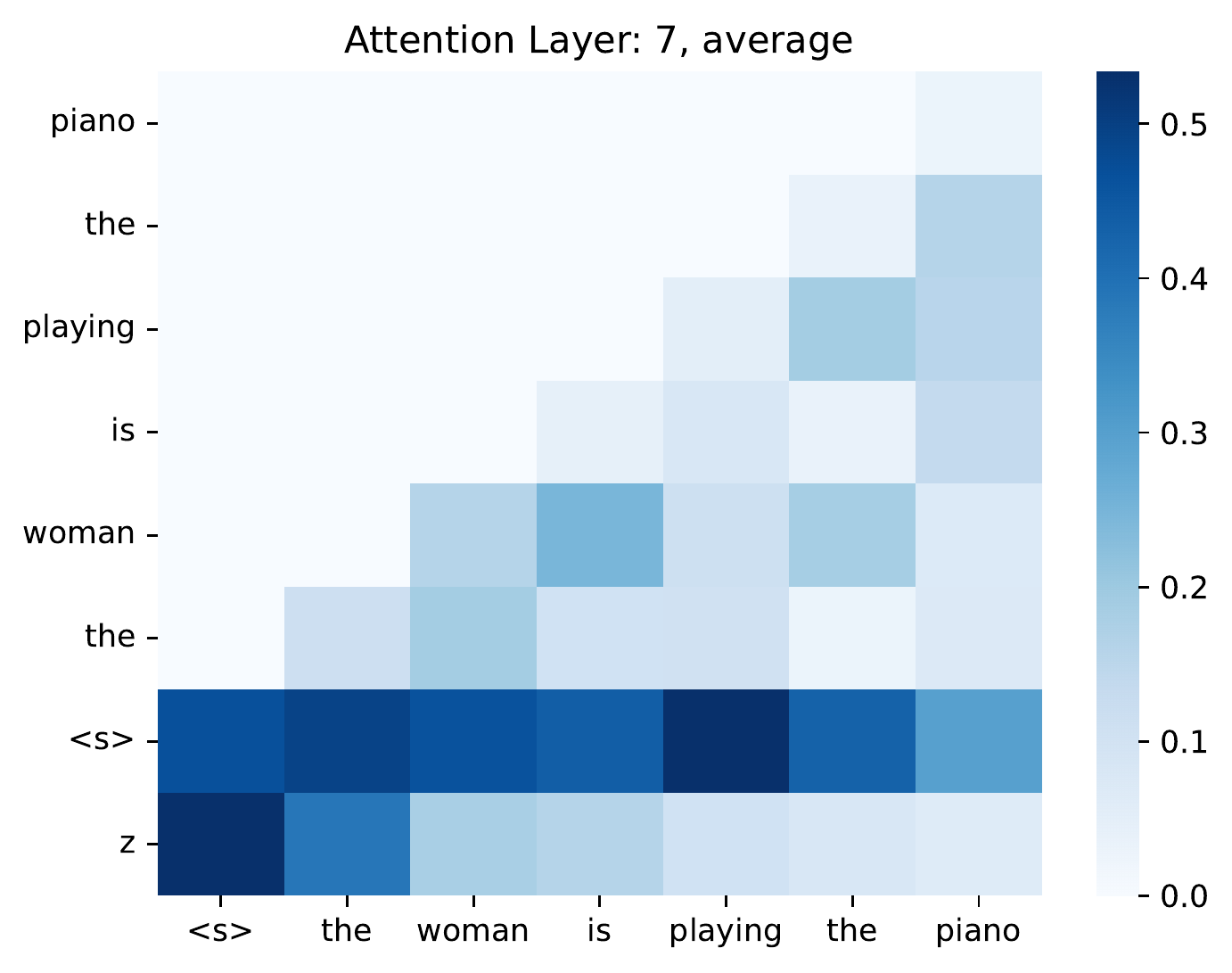} \
\includegraphics[scale=0.35]{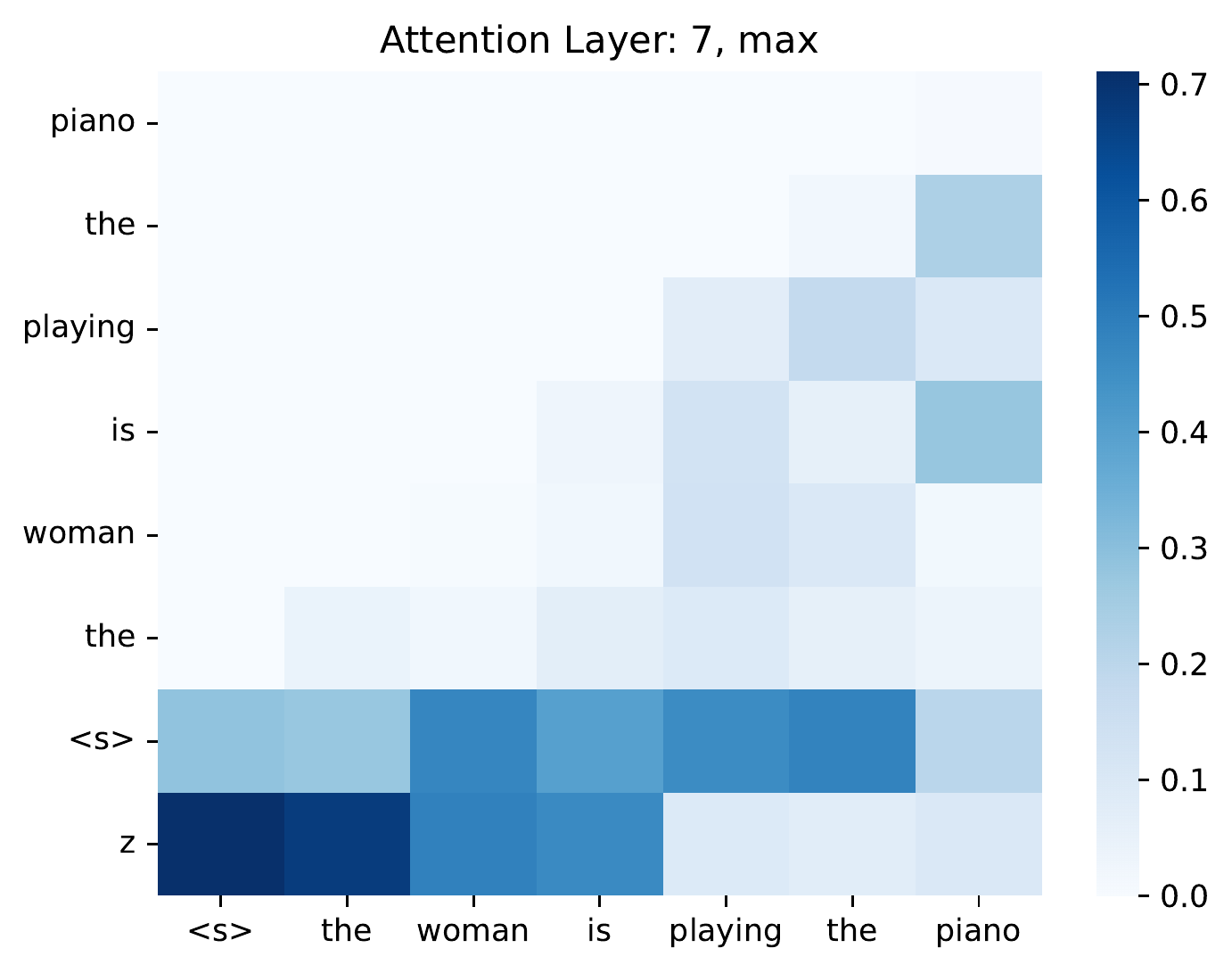} \
\includegraphics[scale=0.35]{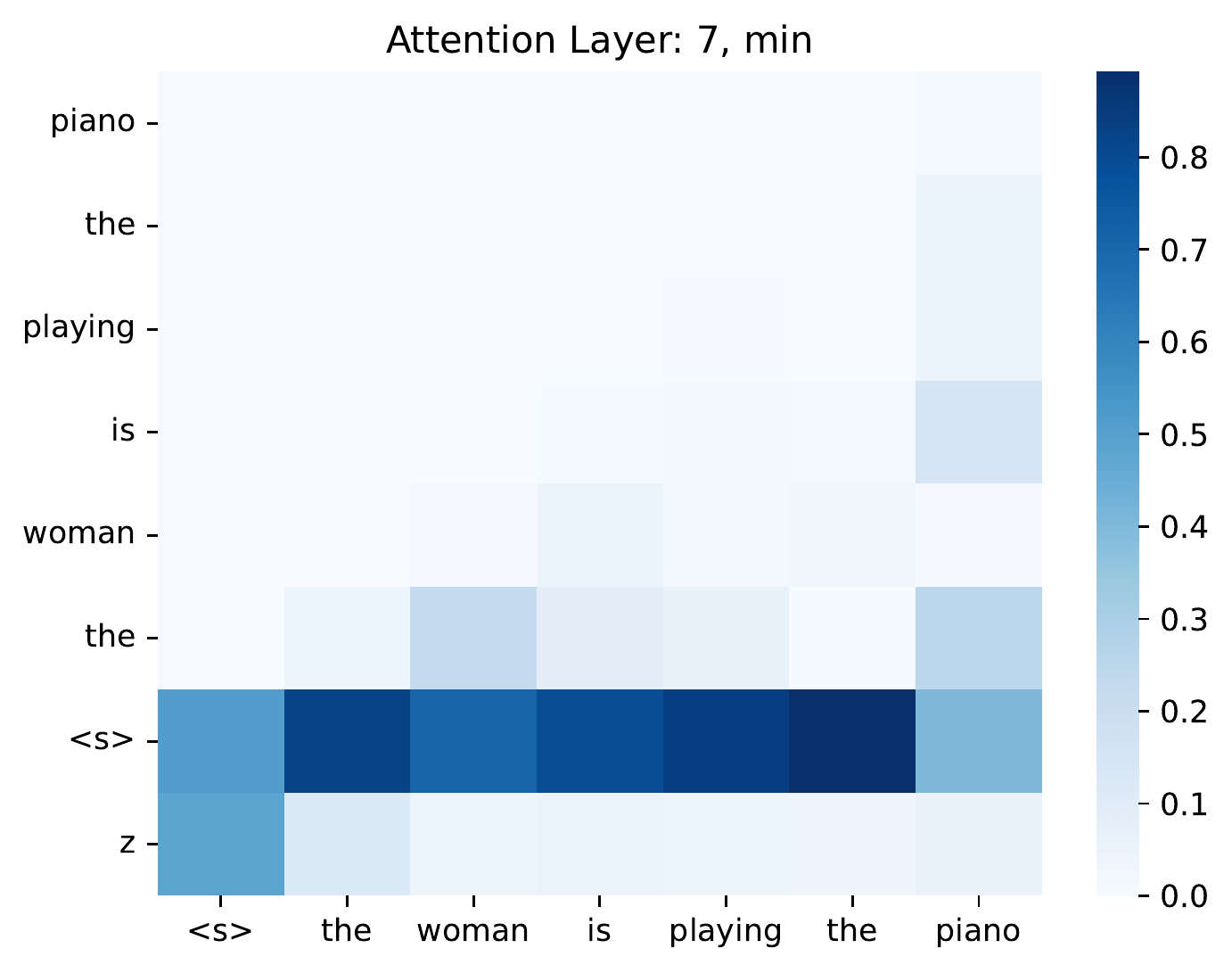} \
\includegraphics[scale=0.35]{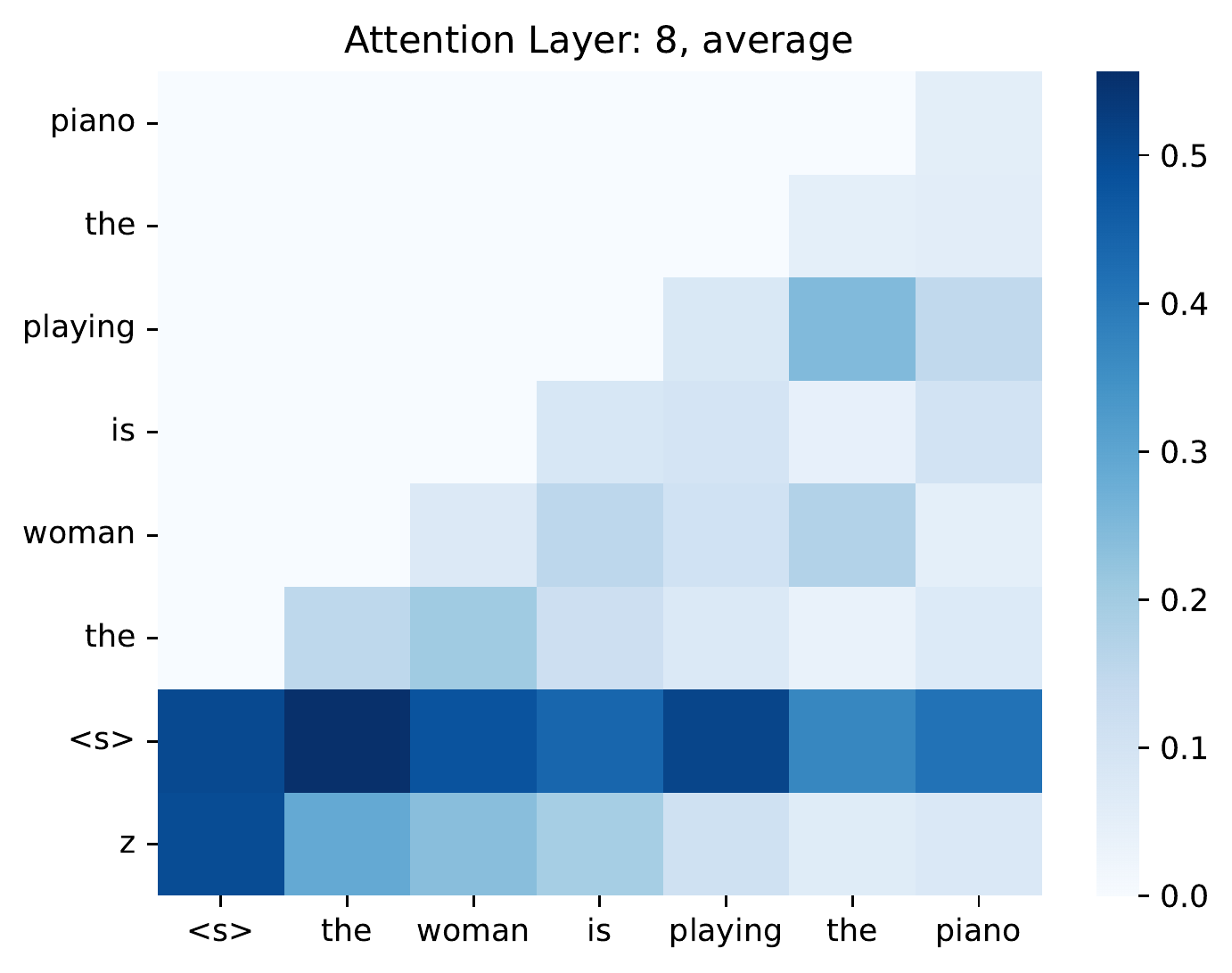} \
\includegraphics[scale=0.35]{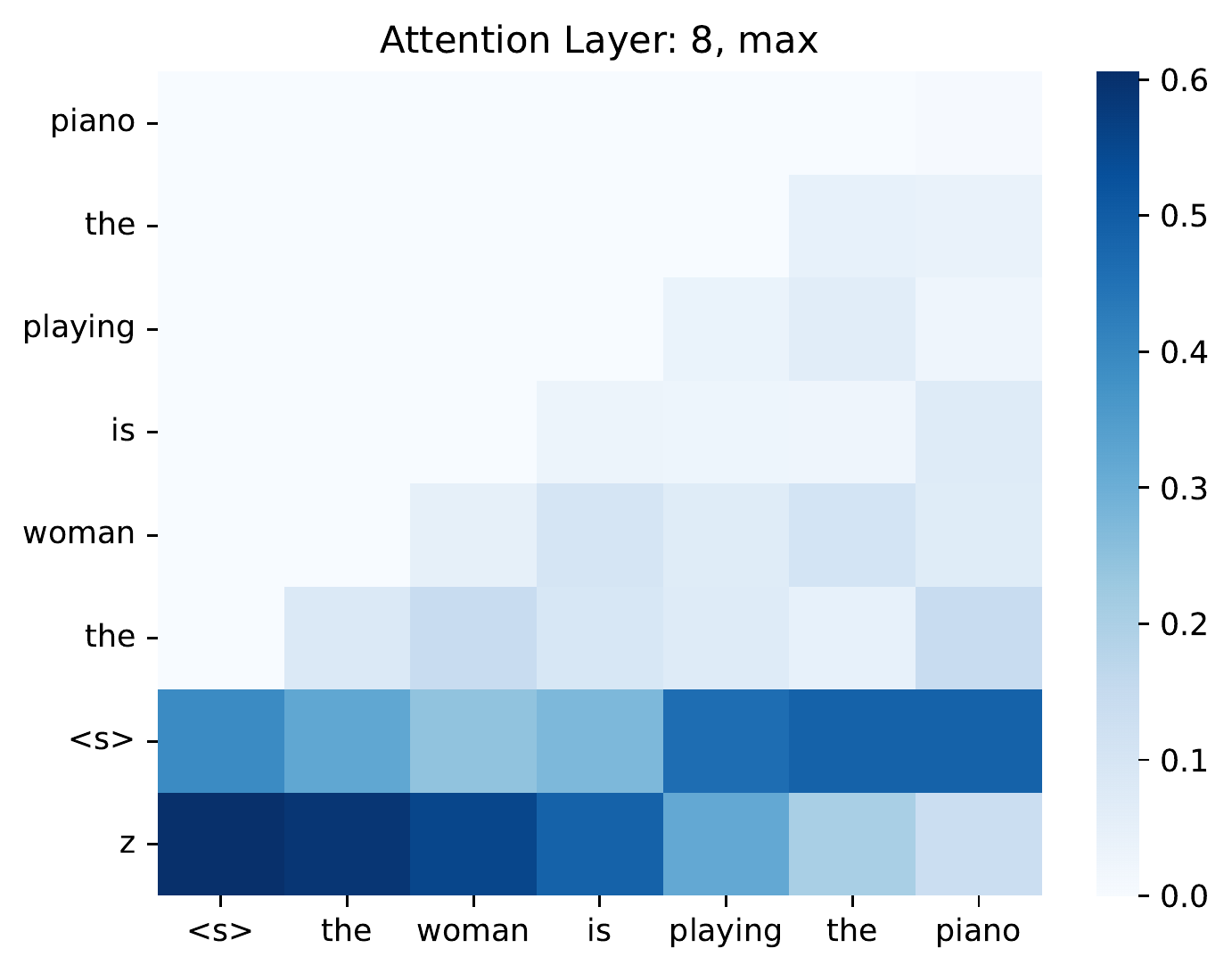} \
\includegraphics[scale=0.35]{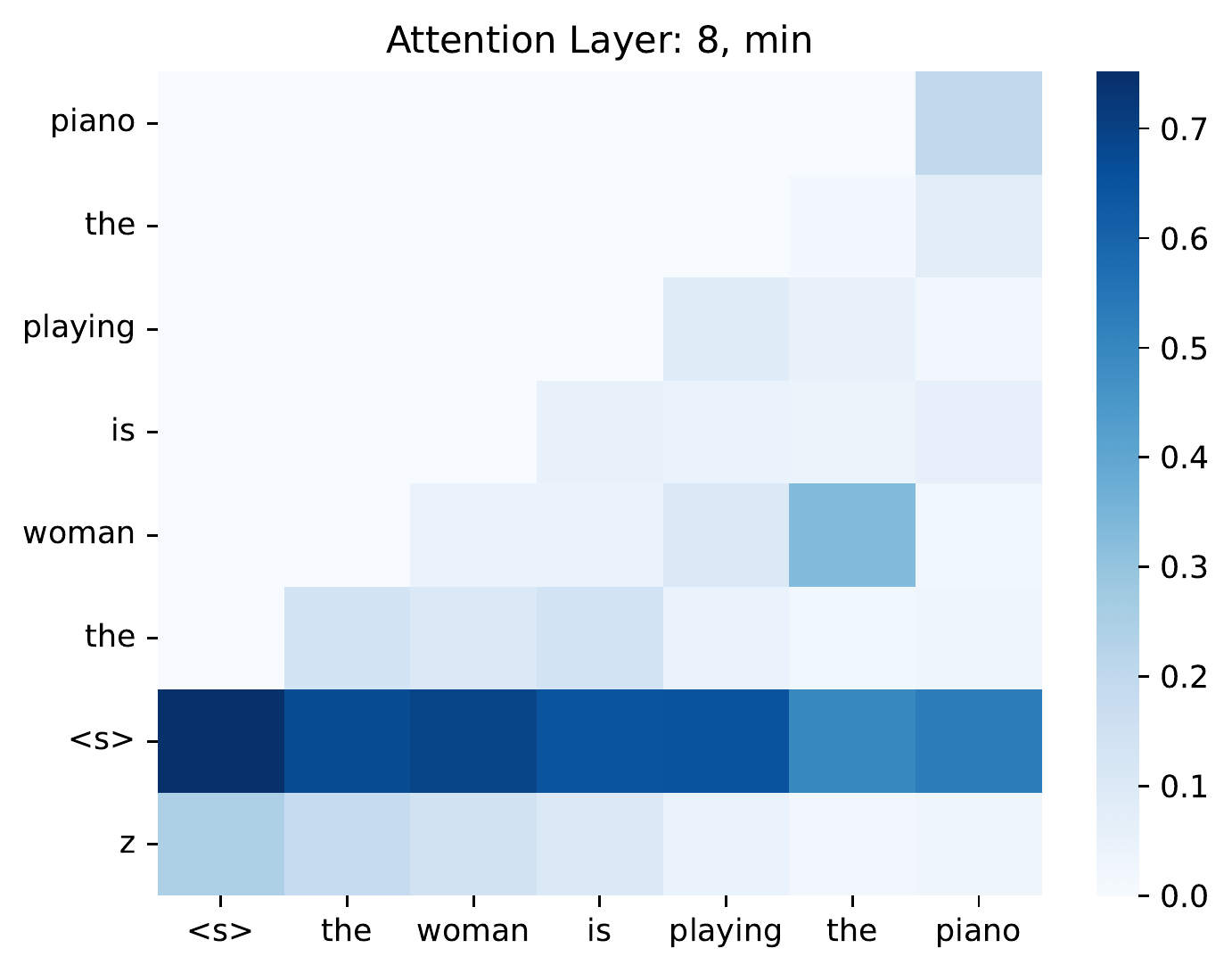} \
\includegraphics[scale=0.35]{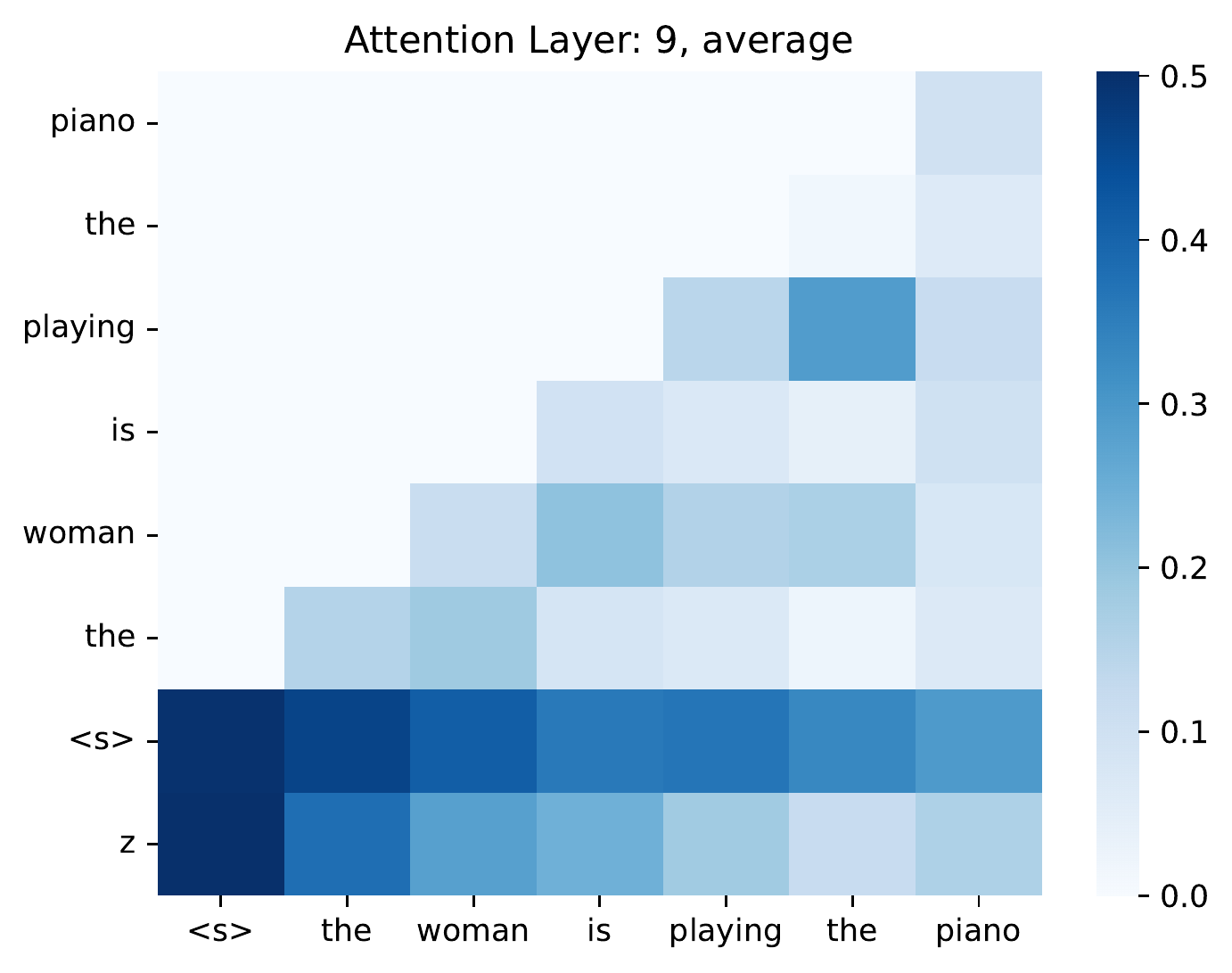} \
\includegraphics[scale=0.35]{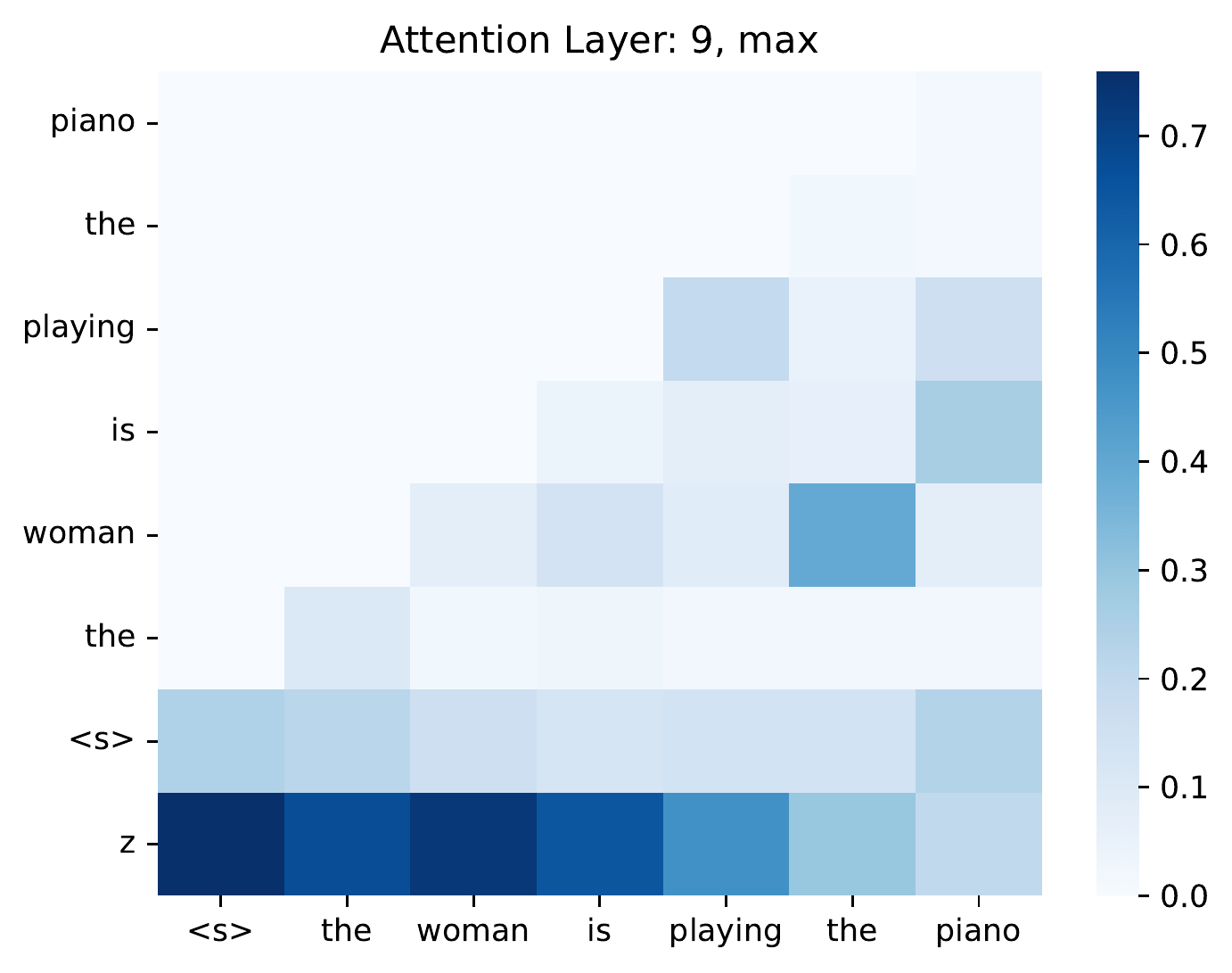} \
\includegraphics[scale=0.35]{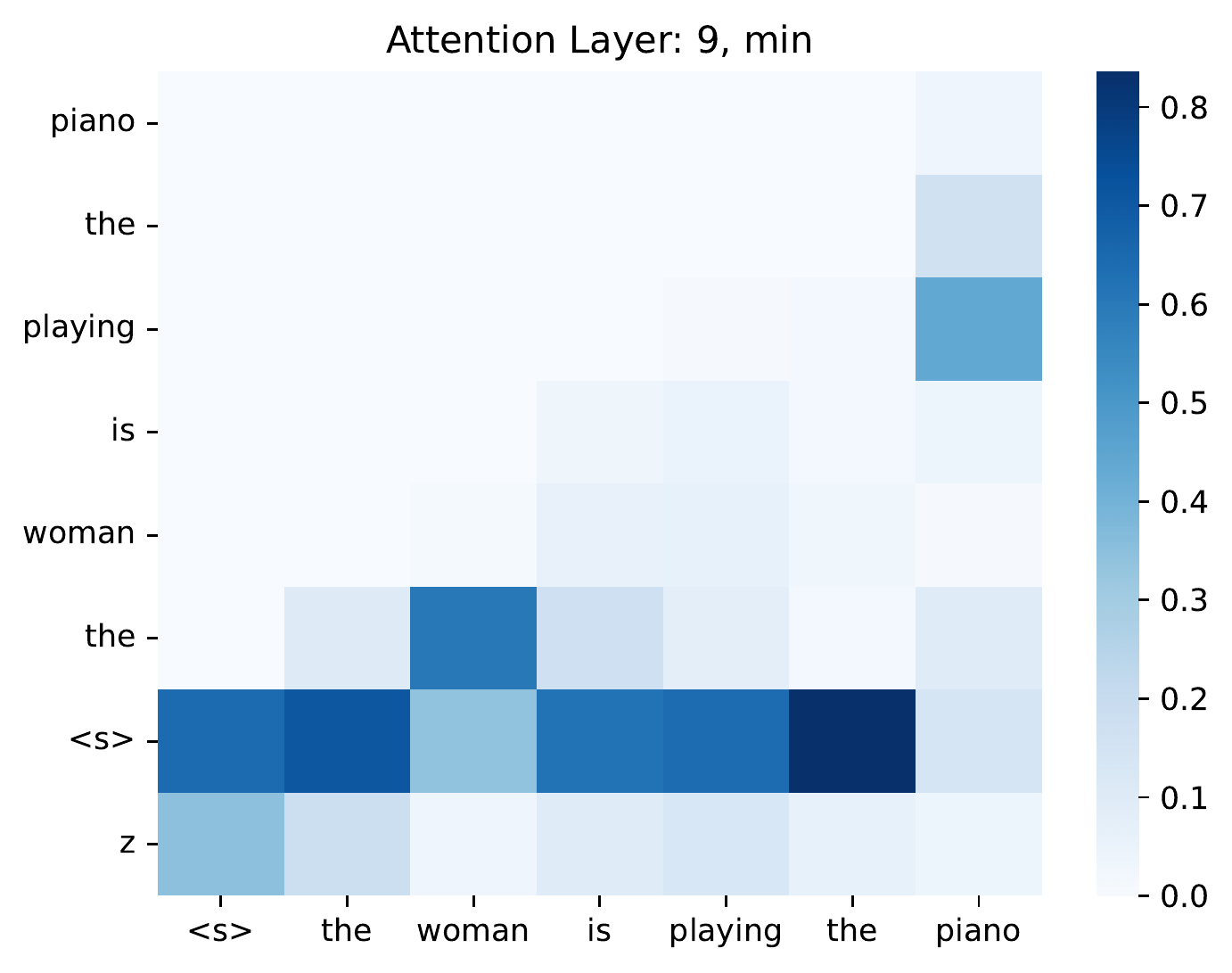} \
\includegraphics[scale=0.35]{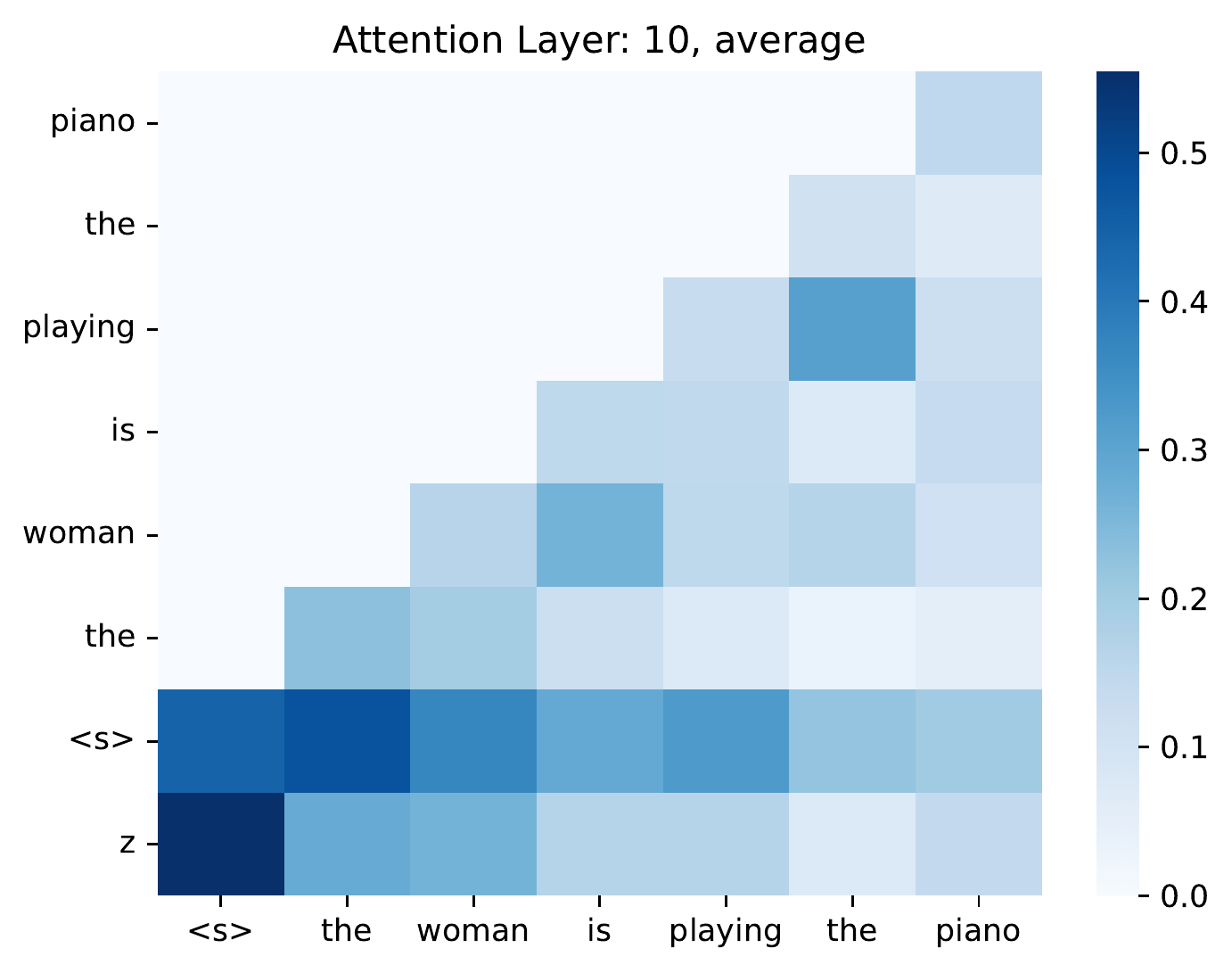} \
\includegraphics[scale=0.35]{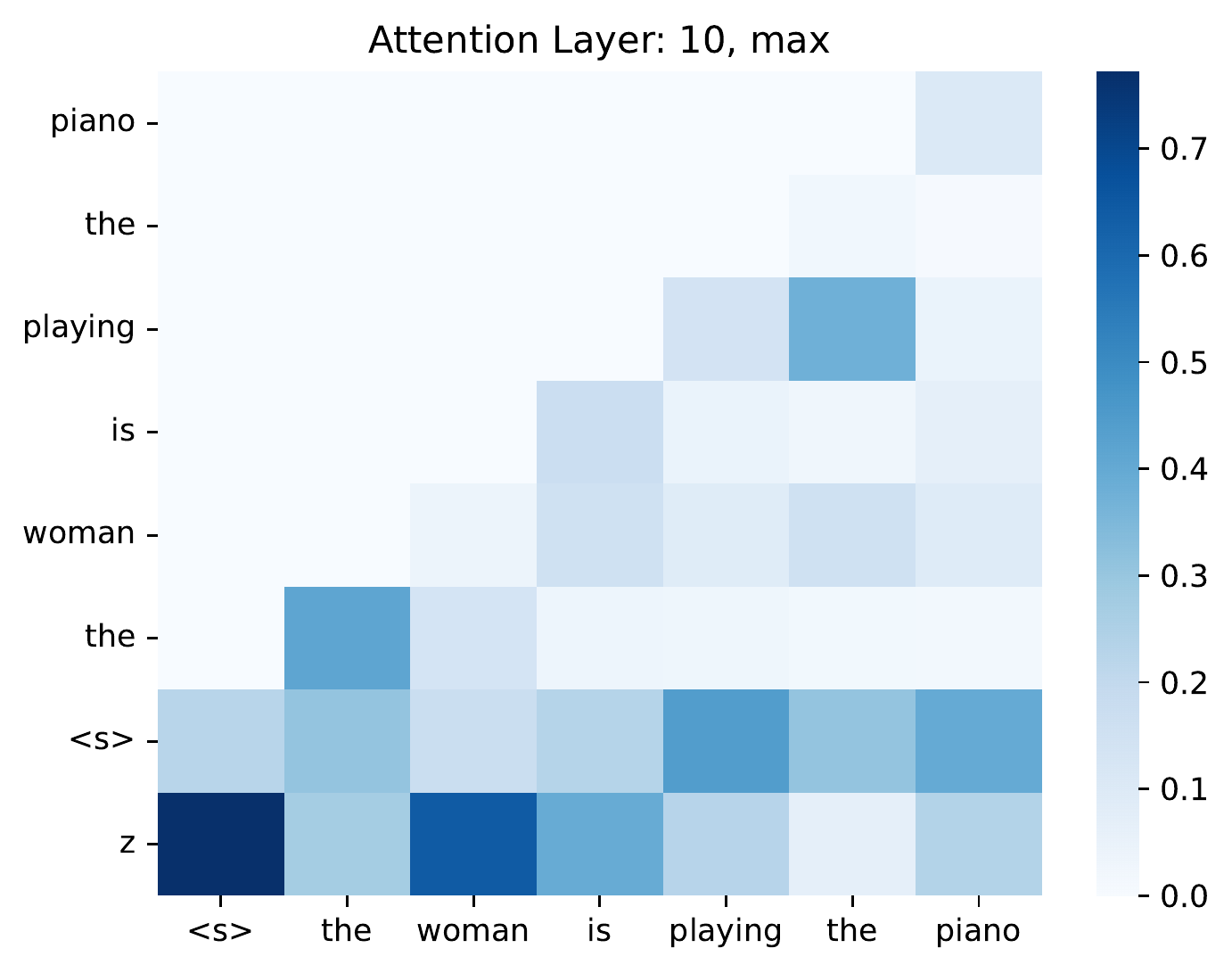} \
\includegraphics[scale=0.35]{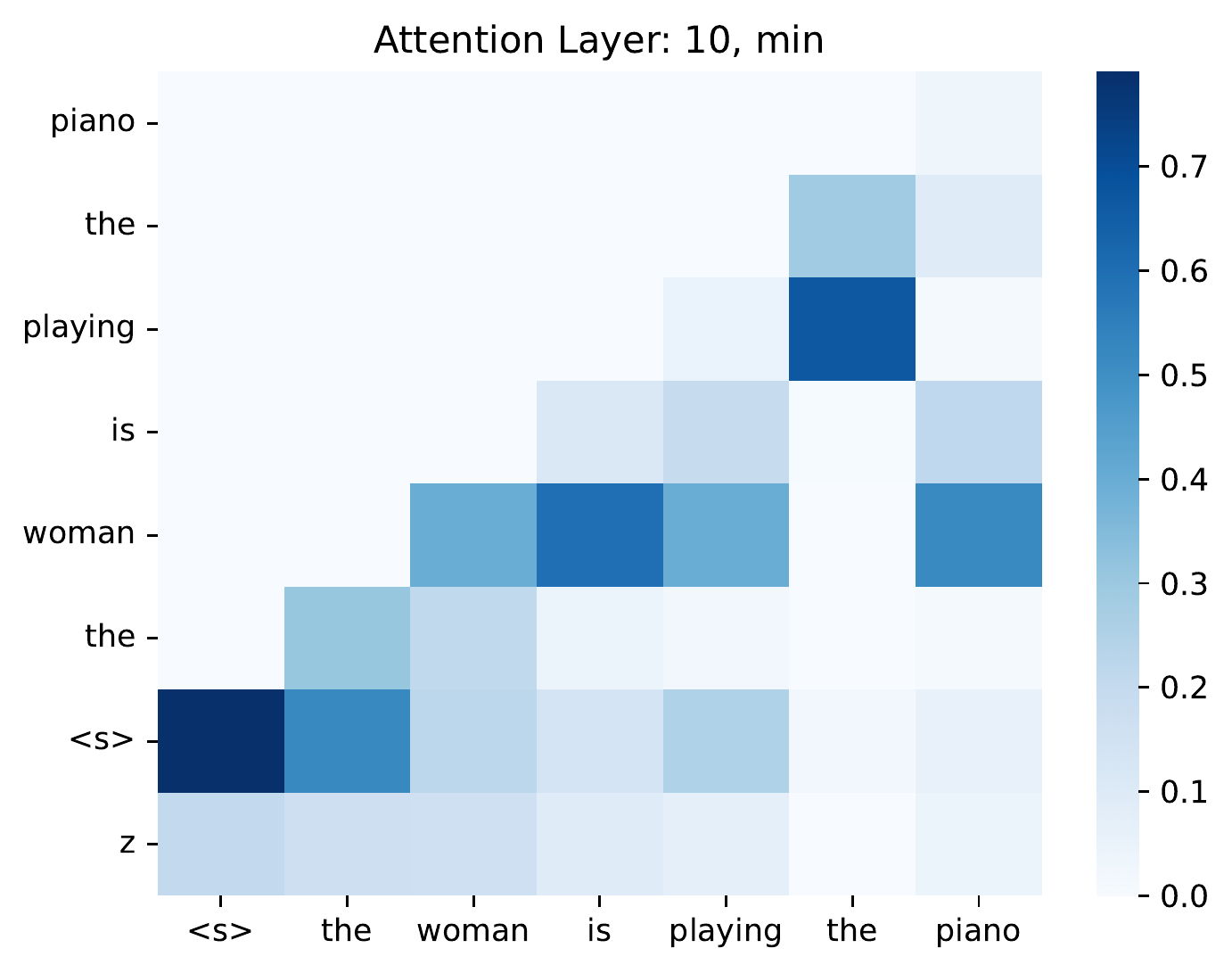} \
\includegraphics[scale=0.35]{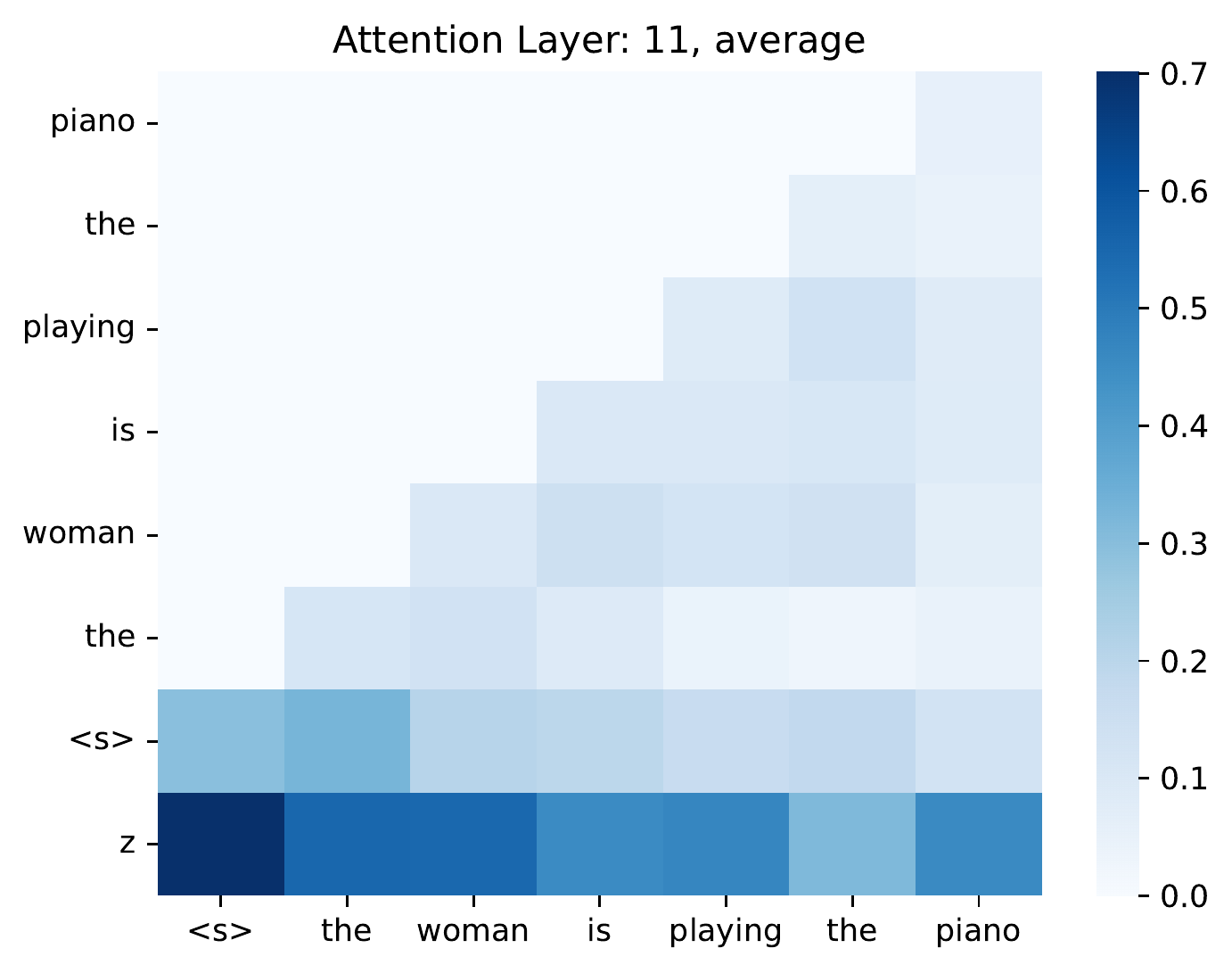} \
\includegraphics[scale=0.35]{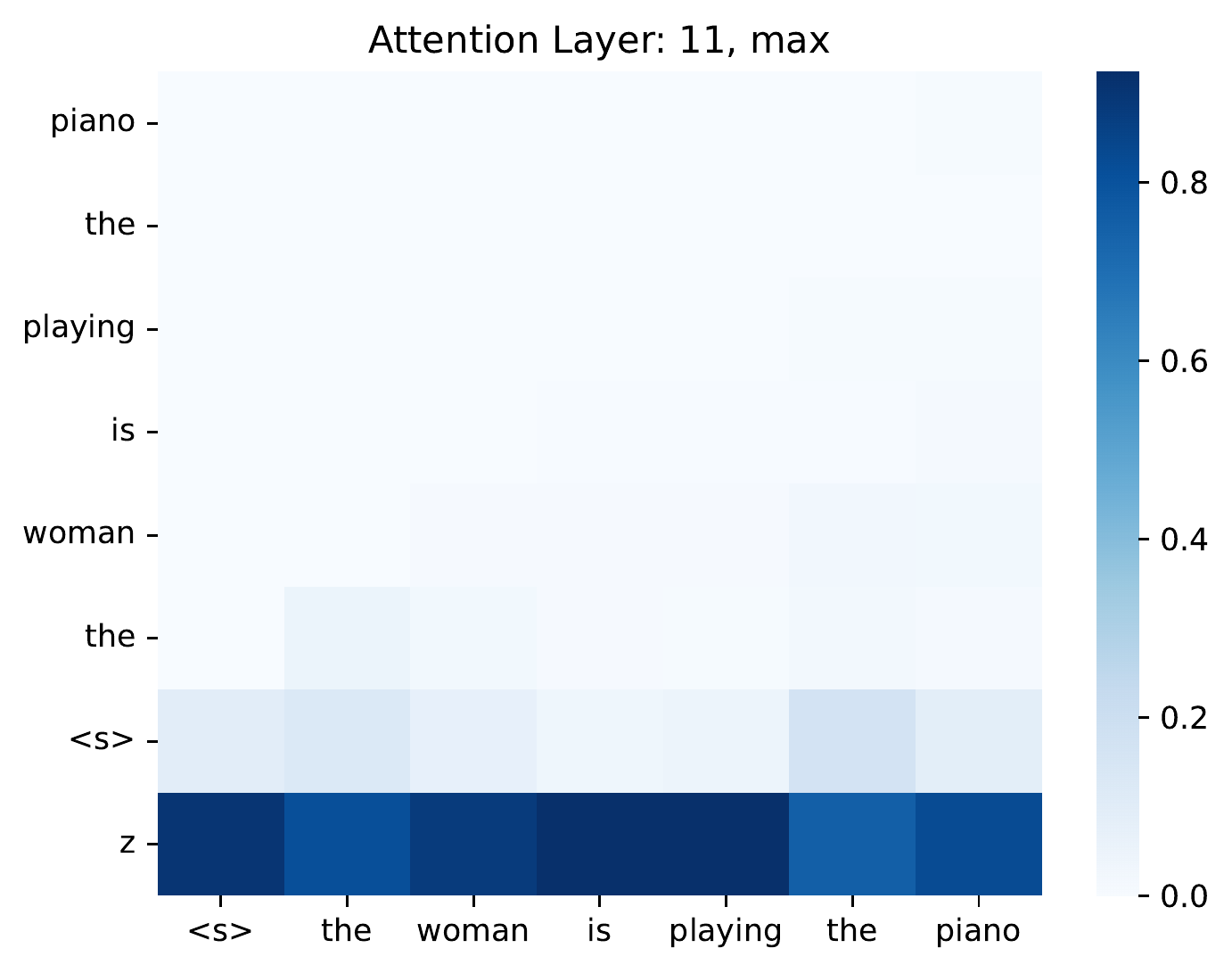} \
\includegraphics[scale=0.35]{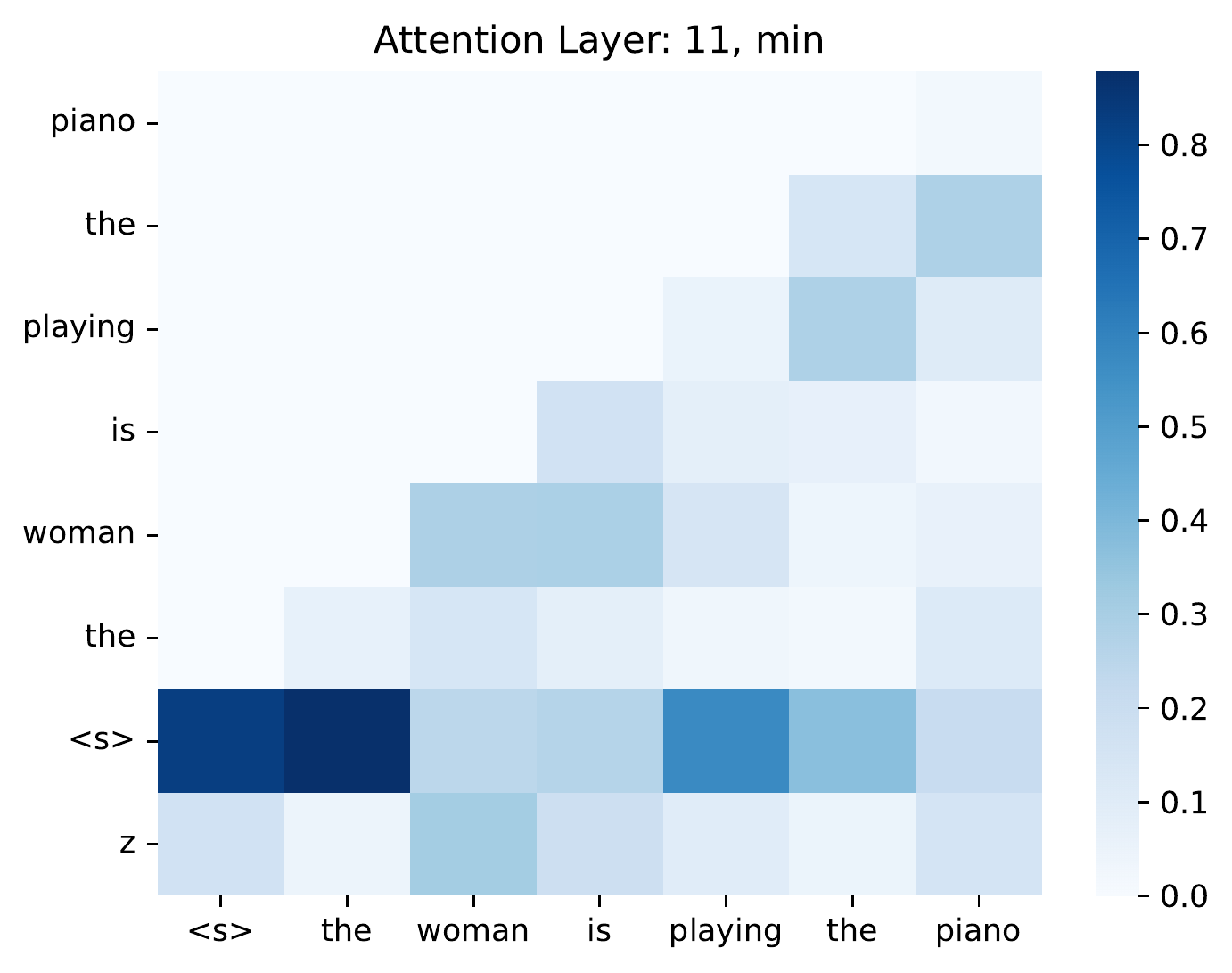} \
\caption{Attention weight of Memory paradigm for layer 6 to layer 11. }
\label{fig:attn_weight2}
\end{figure*}

\end{document}